\documentclass{ieeeaccess}
\usepackage{cite}
\usepackage{amsmath,amssymb,amsfonts}
\usepackage{algorithmic}
\usepackage{graphicx}
\usepackage{textcomp}
\usepackage[linesnumbered,ruled]{algorithm2e}
\usepackage{booktabs}
\usepackage{epsfig}
\usepackage{subfigure}
\usepackage{caption,setspace}
\usepackage{xcolor}
\usepackage{changepage}

\let\vec\mathbf

\def\BibTeX{{\rm B\kern-.05em{\sc i\kern-.025em b}\kern-.08em
    T\kern-.1667em\lower.7ex\hbox{E}\kern-.125emX}}
\begin{document}
\history{Date of publication xxxx 00, 0000, date of current version xxxx 00, 0000.}
\doi{10.1109/ACCESS.2017.DOI}

\title{Advanced Cauchy Mutation for Differential Evolution in Numerical Optimization}
\author{
\uppercase{Tae Jong Choi}\authorrefmark{1,2},
\uppercase{Julian Togelius}\authorrefmark{2} \IEEEmembership{Member, IEEE}, and
\uppercase{Yun-Gyung Cheong}\authorrefmark{3} \IEEEmembership{Member, IEEE}
}
\address[1]{Department of Electrical and Computer Engineering, Sungkyunkwan University, Suwon-si, Gyeonggi-do 16419, Republic of Korea}
\address[2]{Department of Computer Science and Engineering, Tandon School of Engineering, New York University, Brooklyn, NY 11201, USA}
\address[3]{College of Software, Sungkyunkwan University, Suwon-si, Gyeonggi-do 16419, Republic of Korea}
\tfootnote{This work was supported by the National Research Foundation of Korea(NRF) grant funded by the Korea government(MSIT) (No. NRF-2017R1C1B2012752) and the Ministry of Education (No. NRF-2019R1A2C1006316).}

\markboth
{T. J. Choi \headeretal: Advanced Cauchy Mutation for Differential Evolution in Numerical Optimization}
{T. J. Choi \headeretal: Advanced Cauchy Mutation for Differential Evolution in Numerical Optimization}

\corresp{Corresponding author: Yun-Gyung Cheong (e-mail: aimecca@skku.edu).}

\begin{abstract}
Among many evolutionary algorithms, differential evolution (DE) has received much attention over the last two decades. DE is a simple yet powerful evolutionary algorithm that has been used successfully to optimize various real-world problems. Since it was introduced, many researchers have developed new methods for DE, and one of them makes use of a mutation based on the Cauchy distribution to increase the convergence speed of DE. The method monitors the results of each individual in the selection operator and performs the Cauchy mutation on consecutively failed individuals, which generates mutant vectors by perturbing the best individual with the Cauchy distribution. Therefore, the method can locate the consecutively failed individuals to new positions close to the best individual. Although this approach is interesting, it fails to take into account establishing a balance between exploration and exploitation. In this paper, we propose a sigmoid based parameter control that alters the failure threshold for performing the Cauchy mutation in a time-varying schedule, which can establish a good ratio between exploration and exploitation. Experiments and comparisons have been done with six conventional and six advanced DE variants on a set of 30 benchmark problems, which indicate that the DE variants assisted by the proposed algorithm are highly competitive, especially for multimodal functions.
\end{abstract}


\titlepgskip=-15pt

\maketitle

\section{Introduction}
\label{sec:Introduction}
\PARstart{A}{n} evolutionary algorithm (EA) is a population-based metaheuristic inspired by biological evolution and genetic variations observed in nature. An EA takes a population of individuals and searches for a global optimum by artificially designed evolutionary operators. An EA does not make any assumption about a given problem, such as continuity and differentiability. Thus, it can be applied to any problem.

Among many EAs, differential evolution (DE) \cite{storn1997differential, price2006differential} is widely considered to be one of the most effective EAs to optimize multidimensional real-valued functions. DE takes a population of individuals and iterates the mutation, crossover, and selection operators to search for a global optimum. Since it was introduced, DE has received much attention because it is simple and straightforward to implement. The effectiveness of DE has been demonstrated successfully in various real-world problems \cite{das2011differential, das2016recent}.

To increase the convergence speed of DE, Ali and Pant \cite{ali2011improving} proposed a variant of DE called modified DE (MDE), which makes use of a mutation based on the Cauchy distribution as an additional operator. MDE monitors the results of each individual in the selection operator. When an individual consecutively failed to find a better position than its current position for a predefined number of generations, MDE performs the Cauchy mutation on the individual, which generates a mutant vector by perturbing the best individual with the Cauchy distribution. Therefore, MDE can locate the consecutively failed individual to a new position close to the best individual. However, MDE uses the same failure threshold for performing the Cauchy mutation throughout the whole search process, which causes a serious limitation of establishing a balance between exploration and exploitation. Moreover, MDE uses the best individual based Cauchy mutation, which may reduce the diversity of individuals drastically. Here, exploration is the property of searching for entirely new areas of a search space, while exploitation is the property of searching for the areas close to previously searched \cite{eiben1998evolutionary, vcrepinvsek2013exploration}.

EAs need to establish a balance between exploration and exploitation to be successful \cite{eiben1998evolutionary, vcrepinvsek2013exploration}. In this paper, we propose a variant of MDE called advanced Cauchy mutation DE (ACM-DE), which alters the failure threshold for performing the Cauchy mutation in a time-varying schedule. The earlier work of ACM-DE can be found in \cite{choi2018accelerating}. ACM-DE uses a sigmoid based parameter control, which assigns a high failure threshold at the beginning of the search process and gradually reduces it over generations. Therefore, ACM-DE performs the Cauchy mutation with a low probability at the early stage of the search process to preserve the diversity of individuals and a high probability at the late stage of the search process to increase the convergence speed. ACM-DE also uses the $p$-best individual \cite{zhang2009jade} based Cauchy mutation to prevent premature convergence. Therefore, ACM-DE can establish a good ratio between exploration and exploitation compared to MDE.

Experiments and comparisons were carried out with six conventional and six advanced DE variants on the Congress on Evolutionary Computation (CEC) 2017 benchmark problems \cite{awad2016problem}, which contains a set of 30 benchmark problems. The experimental results indicate that the DE variants assisted by the proposed Cauchy mutation significantly outperform the DE variants assisted by the previous Cauchy mutation as well as the original DE variants. As a result, the main contributions of this paper are as follows.

\begin{enumerate}
\item ACM is simple and easy to embed into any DE variant as an additional operator. \label{item:1}
\item ACM can establish a good ratio between exploration and exploitation. \label{item:2}
\item ACM is a clear advance on the previous Cauchy mutation, and the effectiveness of ACM is demonstrated successfully by experiments and comparisons. \label{item:3}
\end{enumerate}

The rest of this paper is organized as follows: we introduce DE and the Cauchy distribution in Section \ref{sec:Background}. In Section \ref{sec:RelatedWork}, we review six advanced DE variants used for experiments and comparisons in this paper. Section \ref{sec:ProposedAlgorithm} presents the technical details of the proposed algorithm. Section \ref{sec:ExperimentalSetup} presents the experimental setups. In Sections \ref{sec:ExperimentalResults} and \ref{sec:Analysis}, we discuss the experimental results of the proposed algorithm. Finally, we provide the conclusion and the future work of this paper in Section \ref{sec:Conclusion}.


\section{Background}
\label{sec:Background}

\subsection{Differential Evolution}
DE is a population-based metaheuristic that takes a population of $NP$ target vectors. Each target vector is a $D$-dimensional vector, denoted by $\vec{x_{i,g}} = (x_{i,g}^{1}, x_{i,g}^{2}, \cdots, x_{i,g}^{D})$, where $g = 1, 2, \cdots, G_{max}$. Here, $G_{max}$ denotes the maximum number of generations. DE consists of four operators: initialization, mutation, crossover, and selection. At the beginning of the search process, the initialization operator uniformly distributes the population over a search space. Then, the mutation and crossover operators generate a population of $NP$ trial vectors, and the selection operator makes up a new population for the next generation by comparing each target vector with its corresponding trial vector.

\subsubsection{Initialization}
First, let us define the lower and upper bounds as follows: $\vec{x_{min}} = (x_{min}^{1}, x_{min}^{2}, \cdots, x_{min}^{D})$ and $\vec{x_{max}} = (x_{max}^{1}, x_{max}^{2}, \cdots, x_{max}^{D})$, where $\vec{x_{min}}$ and $\vec{x_{max}}$ denote the lower and upper bounds, respectively. Each component ($j$) of each target vector ($i$) is initialized as follows:

\begin{equation}
x_{i,0}^{j} = x_{min}^{j} + rand_{i}^{j} \cdot (x_{max}^{j} - x_{min}^{j})
\end{equation}

\noindent
where $rand_{i}^{j}$ denotes a uniformly distributed random value in the interval $[0, 1]$.

\subsubsection{Mutation}
The role of the mutation operator is to generate a set of $NP$ mutant vectors. Each mutant vector is denoted by $\vec{v_{i,g}}$. The following list shows the six most commonly used conventional mutation strategies.

DE/rand/1:
\begin{equation}
\vec{v_{i,g}} = \vec{x_{r_{1},g}} + F \cdot (\vec{x_{r_{2},g}} - \vec{x_{r_{3},g}})
\label{mutation1}
\end{equation}

DE/best/1:
\begin{equation}
\vec{v_{i,g}} = \vec{x_{best,g}} + F \cdot (\vec{x_{r_{1},g}} - \vec{x_{r_{2},g}})
\label{mutation2}
\end{equation}

DE/current-to-best/1:
\begin{equation}
\vec{v_{i,g}} = \vec{x_{i,g}} + F \cdot (\vec{x_{best,g}} - \vec{x_{i,g}}) + F \cdot (\vec{x_{r_{1},g}} - \vec{x_{r_{2},g}})
\label{mutation3}
\end{equation}

DE/current-to-rand/1:
\begin{equation}
\vec{v_{i,g}} = \vec{x_{i,g}} + K \cdot (\vec{x_{r_{1},g}} - \vec{x_{i,g}}) + F \cdot (\vec{x_{r_{2},g}} - \vec{x_{r_{3},g}})
\label{mutation4}
\end{equation}

DE/rand/2:
\begin{equation}
\vec{v_{i,g}} = \vec{x_{r_{1},g}} + F \cdot (\vec{x_{r_{2},g}} - \vec{x_{r_{3},g}}) + F \cdot (\vec{x_{r_{4},g}} - \vec{x_{r_{5},g}})
\label{mutation5}
\end{equation}

DE/current-to-best/2:
\begin{equation}
\vec{v_{i,g}} = \vec{x_{i,g}} + F \cdot (\vec{x_{best,g}} - \vec{x_{i,g}}) + F \cdot (\vec{x_{r_{1},g}} - \vec{x_{r_{2},g}}) + F \cdot (\vec{x_{r_{3},g}} - \vec{x_{r_{4},g}})
\label{mutation6}
\end{equation}

\noindent
where $\vec{x_{r_{1},g}}$, $\vec{x_{r_{2},g}}$, $\vec{x_{r_{3},g}}$, $\vec{x_{r_{4},g}}$, and $\vec{x_{r_{5},g}}$ denote randomly selected donor vectors, which are mutually different and not equal to their corresponding target vector $\vec{x_{i,g}}$. Additionally, $\vec{x_{best,g}}$ denotes the best individual. Finally, $F$ and $K$ denote the scaling factor and a uniformly distributed random value in the interval $[0, 1]$, respectively.

\subsubsection{Crossover}
The role of the crossover operator is to generate a set of $NP$ trial vectors. Each trial vector is denoted by $\vec{u_{i,g}}$. There are two classical crossover operators, binomial and exponential. In the binomial crossover, an integer value $j_{rand}$ is randomly selected within $\{1, 2, \cdots, D\}$. The binomial crossover generates each component ($j$) of each trial vector ($i$) as follows.

\begin{equation}
u_{i,g}^{j} = \left\{ \begin{array}{ll}
v_{i,g}^{j} & \textrm{if $rand_{i}^{j} < CR$ or $j = j_{rand}$} \\
x_{i,g}^{j} & \textrm{otherwise}
\end{array} \right.
\end{equation}

\noindent
where $CR$ denotes the crossover rate.

In the exponential crossover, two integer values $n$ and $L$ are initialized first. The starting component $n$ is randomly selected within $\{1, 2, \cdots, D\}$. The number of components $L$ is obtained as follows.

$L = 0$

DO \{ $L = L + 1$ \}

WHILE ($(rand_{i}^{j} < CR)$ AND $(L < D)$)

\noindent
With the two integer values, the exponential crossover generates each component ($j$) of each trial vector ($i$) as follows.

\begin{equation}
u_{i,g}^{j} = \left\{ \begin{array}{ll}
v_{i,g}^{j} & \textrm{if $j = \langle n \rangle_{D}, \langle n+1 \rangle_{D}, \cdots, \langle n+L-1 \rangle_{D}$} \\
x_{i,g}^{j} & \textrm{otherwise} 
\end{array} \right.
\end{equation}

\noindent
where $\langle \cdot \rangle_{D}$ denotes the modulo operator with the divisor $D$.

\subsubsection{Selection}
The selection operator compares each target vector with its corresponding trial vector and picks the better one. In other words, if the fitness value of a trial vector is better than or equal to that of its corresponding target vector, the trial vector is selected as a member of the population for the next generation. Otherwise, the trial vector is discarded, and the target vector is selected. The selection operator makes up a new population for the next generation as follows.

\begin{equation}
\vec{x_{i,g+1}} = \left\{ \begin{array}{ll}
\vec{u_{i,g}} & \textrm{if $f(\vec{u_{i,g}}) \leq f(\vec{x_{i,g}})$} \\
\vec{x_{i,g}} & \textrm{otherwise.}
\end{array} \right.
\end{equation}

\noindent
where $f(\vec{x})$ denotes an objective function.

DE iterates the mutation, crossover, and selection operators until one of the termination criteria is satisfied. The most commonly used termination criterion is to reach the maximum number of generations $G_{max}$ or the maximum number of function evaluations $NFE_{max}$.

\subsection{Analysis of Cauchy Distribution}
The Cauchy distribution is a continuous probability distribution that has two parameters, $x_{0}$ and $\gamma$. $x_{0}$ is the location parameter, and $\gamma$ is the scale parameter that determines the shape of the Cauchy distribution. For example, if a higher value is assigned to $\gamma$, the height of the peak of the probability density function (PDF) will be shorter, and its width will be wider. On the other hand, if a lower value is assigned to $\gamma$, the height of the peak of the PDF will be taller, and its width will be narrower. The PDF of the Cauchy distribution can be defined as follows.

\begin{equation}
f(x; x_{0}, \gamma) = \frac{1}{\pi \gamma[1 + (\frac{x - x_{0}}{\gamma})^{2}]} = \frac{1}{\pi} \biggl[ \frac{\gamma}{(x - x_{0})^{2} + \gamma^{2}} \biggl]
\end{equation}

\noindent
Additionally, the cumulative distribution function of the Cauchy distribution can be defined as follows.

\begin{equation}
F(x; x_{0}, \gamma) = \frac{1}{\pi} arctan \biggl( \frac{x - x_{0}}{\gamma} \biggl) + \frac{1}{2}
\end{equation}

\begin{figure}[!t]
\centering
\includegraphics[width=\linewidth]{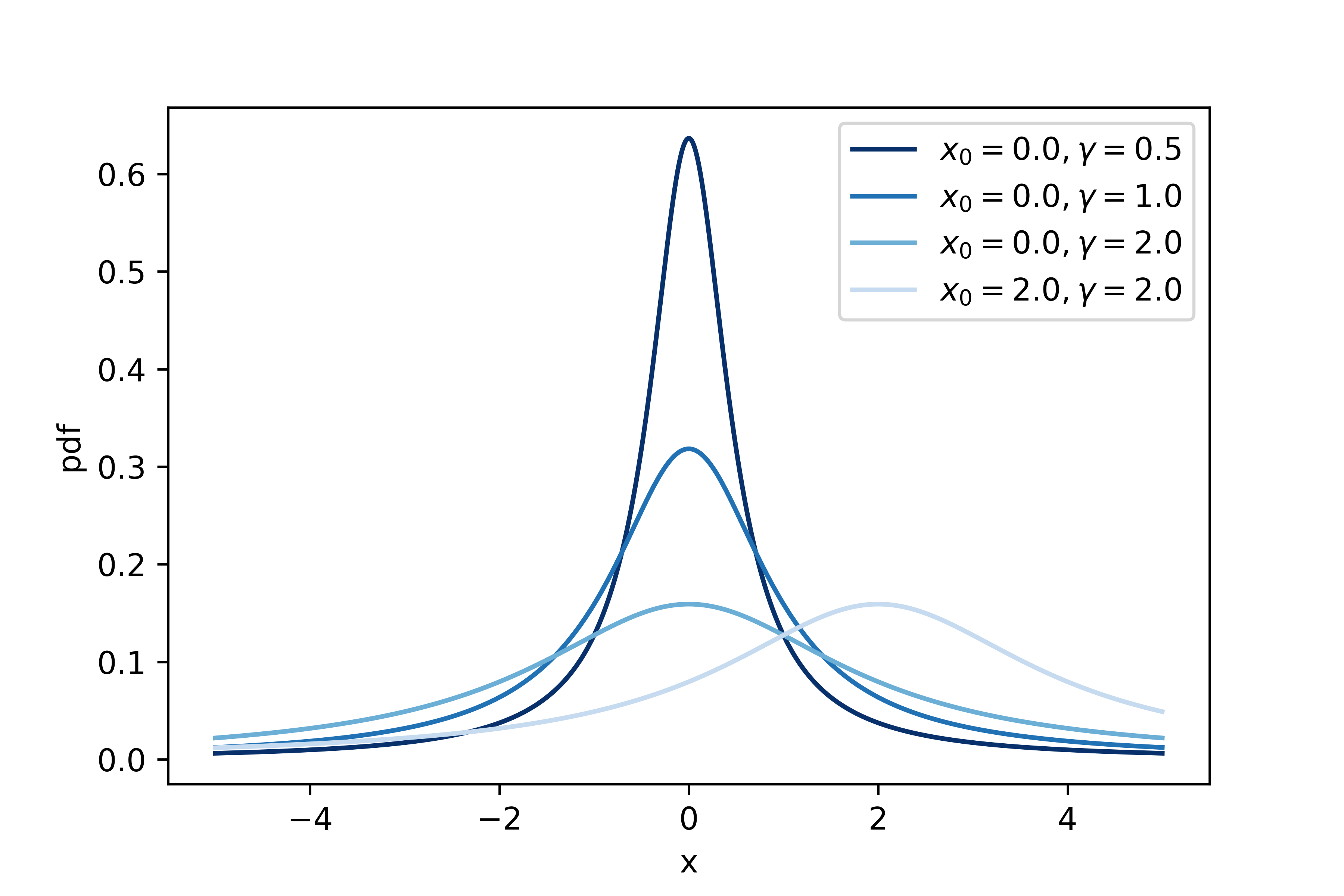}
\caption{Various probability density functions of Cauchy distribution}
\label{fig_pdfs_cauchy}
\end{figure}

Fig. \ref{fig_pdfs_cauchy} presents the various PDFs of the Cauchy distribution.

\section{Related Work}
\label{sec:RelatedWork}
Since it was introduced, many researchers have developed new methods for DE \cite{qin2008differential, mallipeddi2011differential, wang2011differential, choi2019adaptive, choi2018performance, choi2017adaptive1, choi2017improved, zhang2009jade, tanabe2013success, tanabe2014improving, wu2016differential, wu2018ensemble, choi2015adaptive1, choi2014adaptive1, stanovov2018lshade, awad2017ensemble, brest2017single, bujok2017enhanced, mohamed2017lshade, jagodzinski2017differential, choi2017adaptive2, choi2015adaptive2, choi2014adaptive2, choi2013adaptive, choi2018asynchronous1, choi2018asynchronous2, moraglio2009geometric, moraglio2013geometric, yang2014differential, yang2016differential}. For more detailed information, please refer to the following papers \cite{das2011differential, das2016recent, al2018algorithmic, opara2019differential}. In this section, we review six advanced DE variants used for experiments and comparisons in this paper.

\subsection{S\lowercase{a}DE}
Qin et al. proposed a self-adaptive DE variant called SaDE \cite{qin2008differential}, which automatically adjusts mutation strategies and control parameters during the search process. SaDE uses four mutation strategies, "DE/rand/1", "DE/current-to-best/1", "DE/rand/2", and "DE/current-to-rand/1." At each generation, SaDE assigns each individual one of the mutation strategies according to strategy probabilities $p_{k}$, $k = \{ 1, 2, 3, 4 \}$. The strategy probabilities are initialized to $\frac{1}{K}$, $K = 4$ at the beginning of the search process. To update the strategy probabilities, SaDE monitors the success and failure results of each individual in the selection operator. At each generation, after a predefined number of generations $LP$, SaDE updates the strategy probabilities as follows.

\begin{equation}
p_{k,g} = \frac{S_{k,g}}{\sum_{k=1}^{K} S_{k,g}}
\end{equation}

\noindent
where

\begin{equation}
S_{k,g} = \frac{\sum_{t=g-LP}^{g-1} ns_{k,t}}{\sum_{t=g-LP}^{g-1} ns_{k,t} + \sum_{t=g-LP}^{g-1} nf_{k,t}} + \epsilon
\end{equation}

\noindent
where $ns_{k,g}$ and $nf_{k,g}$ denote the success and failure memories with the $k$th mutation strategy, respectively. Additionally, $\epsilon = 0.001$ is used to avoid division by zero.

At each generation, SaDE assigns each individual a scaling factor $F_{i,g}$ with the Gaussian distribution as follows.

\begin{equation}
F_{i,g} = rndn_{i}(0.5, 0.3)
\end{equation}

\noindent
Similarly, at each generation, SaDE assigns each individual a crossover rate $CR_{i,g}$ with the Gaussian distribution as follows.

\begin{equation}
CR_{i,G} = rndn_{i}(CRm_{k}, 0.1)
\end{equation}

\noindent
where $CRm_{k}$ is the median of the crossover rates used by successfully evolved individuals with the $k$th mutation strategy. After that, the crossover rate is truncated to $[0, 1]$.

\subsection{EPSDE}
Mallipeddi et al. proposed a self-adaptive DE variant called EPSDE \cite{mallipeddi2011differential}, which uses a pool of mutation strategies and two pools of control parameters. The pool of mutation strategies contains "DE/best/2", "DE/rand/1", and "DE/current-to-rand/1." Additionally, the pool of scaling factors contains the values in the range of $0.4 - 0.9$ in steps of 0.1, and the pool of crossover rates contains the values in the range of $0.1 - 0.9$ in the steps of 0.1. At each generation, EPSDE assigns each individual a combination of mutation strategy and control parameters taken from the respective pools. If an individual successfully evolved with the combination, the individual uses the same combination for the next generation. Otherwise, EPSDE assigns the individual a new combination or one of the combinations used by successfully evolved individuals with equal probability for the next generation.

\subsection{C\lowercase{o}DE}
Wang et al. proposed an advanced DE variant called CoDE \cite{wang2011differential}, which generates three candidate trial vectors for each individual. CoDE uses three mutation strategies, which are "DE/rand/1", "DE/rand/2", and "DE/current-to-rand/1." Additionally, CoDE uses three pairs of control parameters, which are "$F=1.0, CR=0.1$", "$F=1.0, CR=0.9$", and "$F=0.8, CR=0.2$." At each generation, CoDE generates three candidate trial vectors for each individual by using different mutation strategies and three randomly selected pairs of control parameters. Therefore, the three candidate trial vectors have different characteristics. Among the three candidate trial vectors, the best one is selected for the selection operator.

\subsection{SHADE}
Tanabe and Fukunaga proposed an advanced DE variant called SHADE \cite{tanabe2013success}, which is an extension of an advanced DE variant called JADE \cite{zhang2009jade}. We first introduce JADE and then SHADE. JADE uses an advanced mutation strategy called DE/current-to-$p$best, which can be defined as follows.

DE/current-to-$p$best:
\begin{equation}
\vec{v_{i,g}} = \vec{x_{i,g}} + F \cdot (\vec{x_{pbest,g}} - \vec{x_{i,g}}) + F \cdot (\vec{x_{r_{1},g}} - \vec{\tilde{x}_{r_{2},g}})
\end{equation}

\noindent
where $\vec{x_{pbest,g}}$ and $\vec{\tilde{x}_{r_{2},g}}$ denote one of the top $100p\%$ individuals with $p \in (0, 1]$ and a randomly selected donor vector from a population or an archive of recently discarded target vectors, respectively.

At each generation, JADE assigns each individual a pair of control parameters as follows.

\begin{equation}
F_{i,g} = rndc_{i}(\mu_{F}, 0.1)
\end{equation}

\begin{equation}
CR_{i,g} = rndn_{i}(\mu_{CR}, 0.1)
\end{equation}

\noindent
where $rndc_{i}$ and $rndn_{i}$ denote the Cauchy distribution and the Gaussian distribution, respectively. After that, the scaling factor is regenerated if $F_{i,g} \leq 0$ or truncated to 1 if $F_{i,g} > 1$, and the crossover rate is truncated to $[0, 1]$. The value of $\mu_{F}$ and the value of $\mu_{CR}$ are updated as follows.

\begin{equation}
\mu_{F} = (1 - c) \cdot \mu_{F} + c \cdot mean_{L}(S_{F})
\end{equation}

\begin{equation}
\mu_{CR} = (1 - c) \cdot \mu_{CR} + c \cdot mean_{A}(S_{CR})
\end{equation}

\noindent
where $c$ is a constant in the interval $[0, 1]$. Additionally, $mean_{L}$ and $mean_{A}$ denote the Lehmer mean and the arithmetic mean, respectively. Finally, $S_{F}$ and $S_{CR}$ denote the scaling factors and the crossover rates used by successfully evolved individuals, respectively.

SHADE uses an improved parameter control of JADE, which contains historical memories with $H$ entries for the values of $\mu_{F}$ and the values of $\mu_{CR}$, denoted by $M_{F,k}$ and $M_{CR,k}$. The historical memories are initialized to 0.5 at the beginning of the search process. At each generation, SHADE assigns each individual a pair of control parameters as follows.

\begin{equation}
F_{i,g} = rndc_{i}(M_{F,r_{i}}, 0.1)
\end{equation}

\begin{equation}
CR_{i,g} = rndn_{i}(M_{CR,r_{i}}, 0.1)
\end{equation}

\noindent
where $r_{i}$ is randomly selected within $\{1, 2, \cdots, H\}$. After the selection operator, SHADE updates the historical memories as follows.

\begin{equation}
M_{F,\langle g \rangle_{H}} = \left\{ \begin{array}{ll}
meanw_{L}(S_{F}) & \textrm{if $S_{F} \neq \emptyset$} \\
M_{F,\langle g \rangle_{H}} & \textrm{otherwise.}
\end{array} \right.
\end{equation}

\begin{equation}
M_{CR,\langle g \rangle_{H}} = \left\{ \begin{array}{ll}
meanw_{A}(S_{CR}) & \textrm{if $S_{CR} \neq \emptyset$} \\
M_{CR,\langle g \rangle_{H}} & \textrm{otherwise.}
\end{array} \right.
\end{equation}

\noindent
where $meanw_{L}$ and $meanw_{A}$ denote the weighted Lehmer mean and the weighted arithmetic mean, respectively.

\subsection{MPEDE}
Wu et al. proposed a multi-population-based DE variant called MPEDE \cite{wu2016differential}, which uses three mutation strategies, i.e., "DE/current-to-pbest/1", "DE/current-to-rand/1", and "DE/rand/1." MPEDE splits a population into four subpopulations, which are three equally sized smaller subpopulations and one larger reward subpopulation. Each subpopulation uses a distinct mutation strategy to generate mutant vectors. After every predefined number of generations $LP$, the best mutation strategy is determined by comparing the average fitness improvement. The reward subpopulation then uses the best mutation strategy to generate mutant vectors for the next $LP$ generations. Additionally, MPEDE uses the parameter control of JADE \cite{zhang2009jade} to automatically adjust control parameters during the search process. As a result, MPEDE can find the best mutation strategy and assign it to the reward subpopulation during the search process.

\subsection{EDEV}
Wu et al. proposed a multi-population-based DE variant called EDEV \cite{wu2018ensemble}, which uses three DE variants, i.e., JADE \cite{zhang2009jade}, CoDE \cite{wang2011differential}, and EPSDE \cite{mallipeddi2011differential}. Similar to MPEDE, EDEV splits a population into four subpopulations, which are three equally sized smaller subpopulations and one larger reward subpopulation. Each subpopulation uses a distinct DE variant to generate mutant vectors. After every predefined number of generations $LP$, the best DE variant is determined by comparing the average fitness improvement. The reward subpopulation then uses the best DE variant to generate mutant vectors for the next $LP$ generations. As a result, EDEV can find the best DE variant and assign it to the reward subpopulation during the search process.


\section{Proposed Algorithm}
\label{sec:ProposedAlgorithm}
This section outlines the proposed Cauchy mutation and discusses the details of each algorithmic component.

\begin{figure*}[t]
 \centering
 \subfigure[DE/rand/1]{
  \includegraphics[scale=0.25]{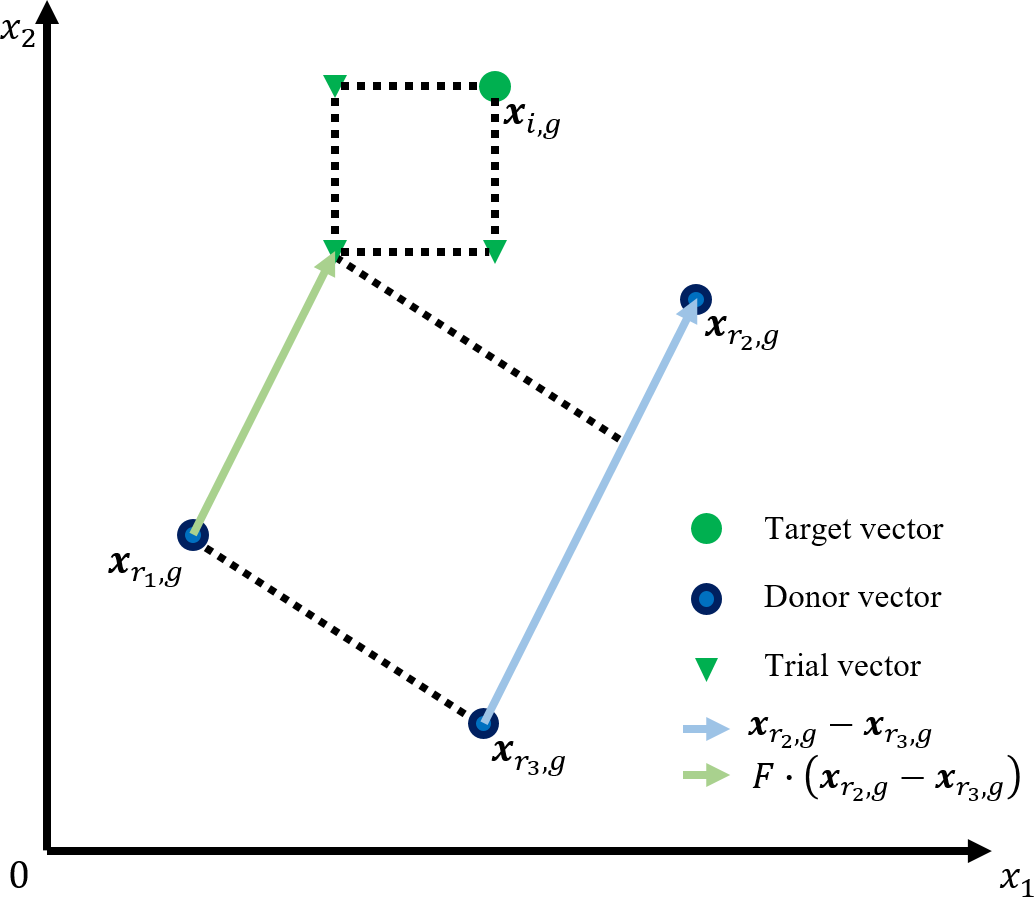}
   }
 \subfigure[Cauchy mutation of MDE]{
  \includegraphics[scale=0.25]{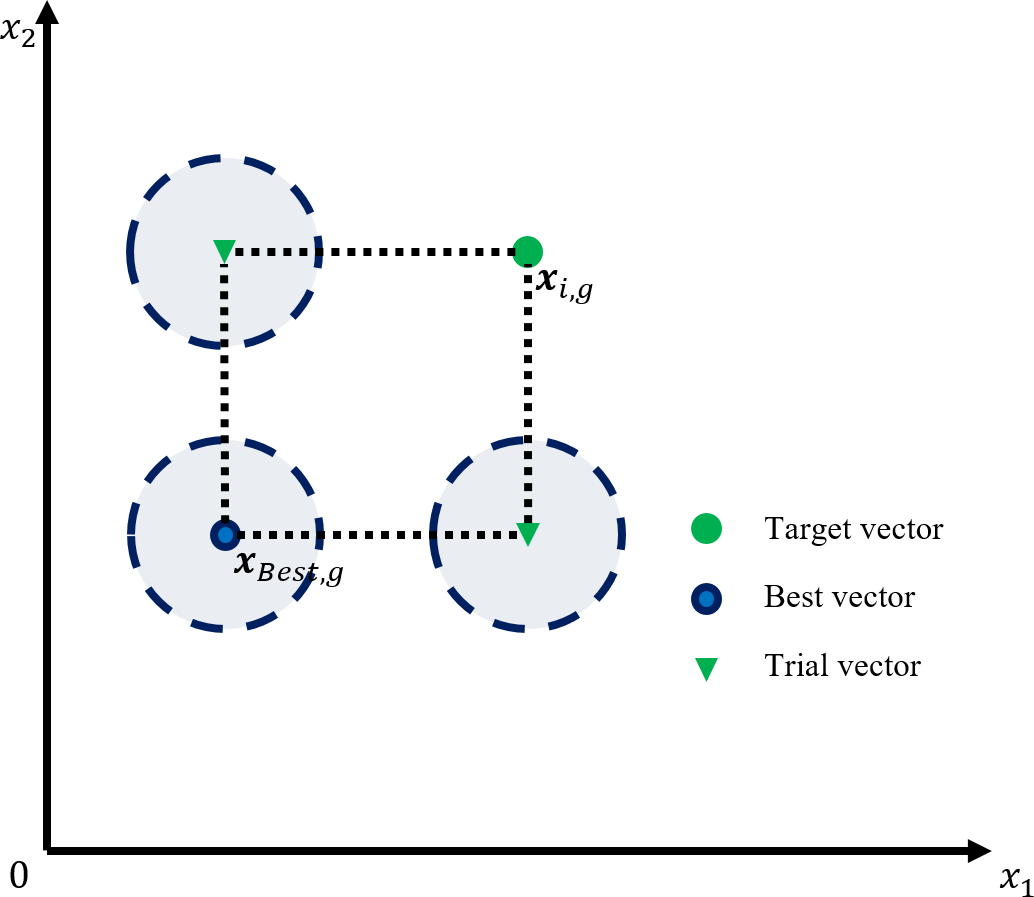}
   }
 \subfigure[Cauchy mutation of ACM-DE]{
  \includegraphics[scale=0.25]{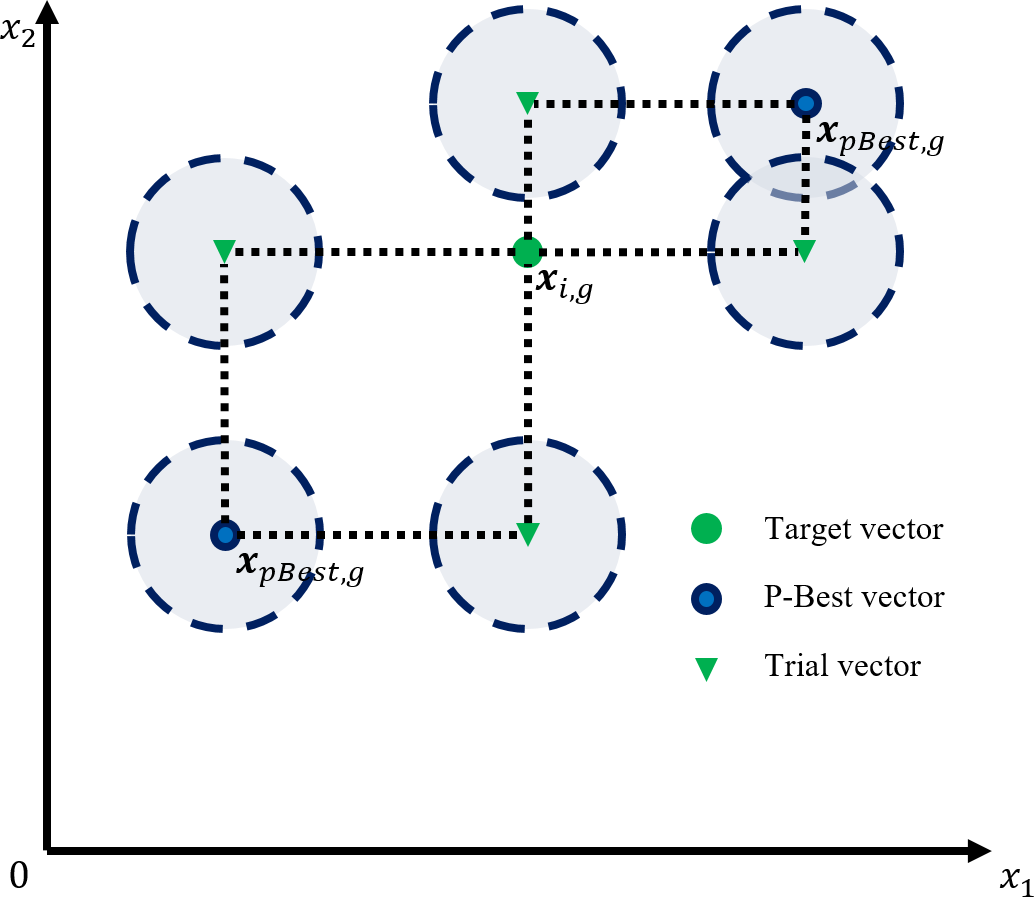}
   }
 \caption[]{Cauchy Mutation of ACM-DE}
 \label{fig:CauchyMutation}
\end{figure*}

\begin{figure}[!t]
\centering
\includegraphics[width=\linewidth]{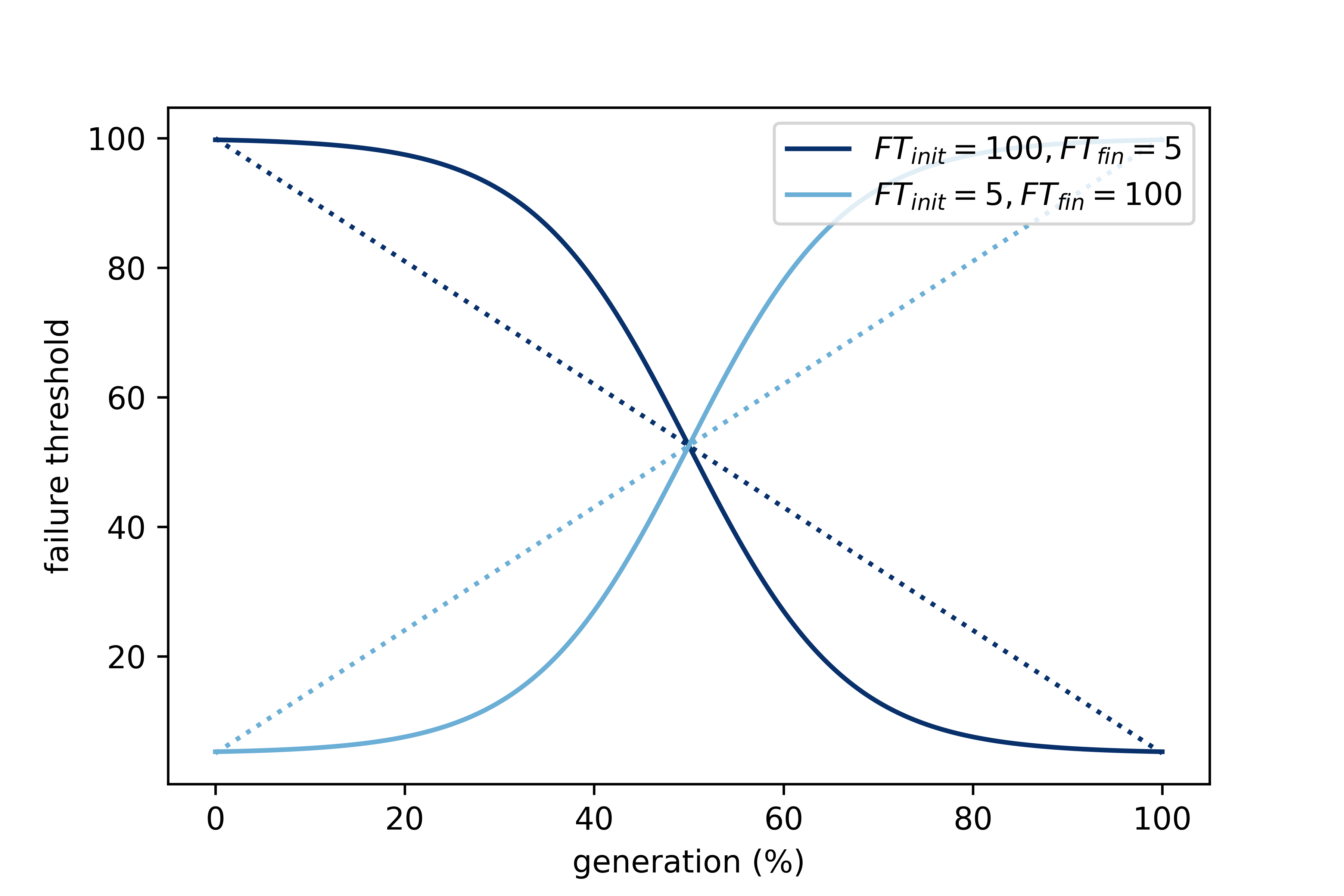}
\caption{Sigmoid Based Parameter Control of ACM-DE}
\label{fig:SigmoidBasedParameterControl}
\end{figure}

\subsection{Review of MDE}
\label{sec:MDE}
MDE \cite{ali2011improving} is the basis of the proposed algorithm, which keeps track of the results of each individual in the selection operator. When an individual consecutively failed to find a better position than its current position for a predefined number of generations, MDE assumes that the individual gets trapped in a local optimum. To help the individual escape from the local optimum, MDE performs the Cauchy mutation on the individual, which generates a mutant vector by perturbing the best individual with the Cauchy distribution. The Cauchy mutation of MDE generates a trial vector as follows.

\begin{equation}
u_{i,g}^{j} = \left\{ \begin{array}{ll}
rndc_{i}^{j}(x_{best,g}^{j}, 0.1) & \textrm{if $rand_{i}^{j} < 0.5$ or $j = j_{rand}$} \\
x_{i,g}^{j} & \textrm{otherwise}
\end{array} \right.
\end{equation}

\noindent
where $rndc_{i}^{j}$ and $rand_{i}^{j}$ denote the Cauchy distribution and a uniformly distributed random value in the interval $[0, 1]$, respectively. Therefore, MDE can locate the consecutively failed individual to a new position close to the best individual.

Compared to other DE variants that make use of a mutation based on the Cauchy distribution, MDE has two main advantages: First, MDE carries out the Cauchy mutation on an individual basis, while other DE variants\cite{qin2010multi, zhang2017adaptive} do so on a population basis. In other words, the DE variants \cite{qin2010multi, zhang2017adaptive} calculate the running metric of convergence \cite{deb2002running} in every predefined number of generations and carry out the Cauchy mutation if the calculated convergence is lower than a predefined threshold. This approach is effective when all individuals get stuck in the same local optimum but ineffective when different individuals get stuck in different local optima. Secondly, there is no increment in computational cost. The Cauchy mutation of MDE requires a computational cost similar to that of the mutation and crossover operators. On the other hand, the Cauchy mutation of the DE variants \cite{qin2010multi, zhang2017adaptive} requires a considerable computational cost because they need to calculate the running metric of convergence and the diversity of each component in every predefined number of generations.

\subsection{Advanced Cauchy Mutation}
\label{sec:AdvancedCauchyMutation}
Although the effectiveness of MDE has been demonstrated successfully on the classical benchmark problems, MDE is not well suited to optimize complex problems. One of the major drawbacks of MDE is that it uses the same failure threshold for performing the Cauchy mutation throughout the whole search process, which causes a serious limitation of establishing a balance between exploration and exploitation. Moreover, MDE uses the best individual based Cauchy mutation, which may reduce the diversity of individuals drastically. The aim of this paper is thus to propose an improved approach, which removes all of the difficulties that MDE faces.

ACM-DE consists of two algorithmic components: a sigmoid based parameter control and the $p$-best individual \cite{zhang2009jade} based Cauchy mutation. The formal can alters the failure threshold for performing the Cauchy mutation in a time-varying schedule. The latter can prevent premature convergence by preventing consecutively failed individuals from converging into one region. The following sections provide a detailed description of these algorithmic components.

\subsubsection{Sigmoid Based Parameter Control for Failure Threshold}
MDE uses a fixed failure threshold for performing the Cauchy mutation, which causes one of the following problems.

\begin{itemize}
\item Premature convergence: MDE with a low failure threshold may execute the Cauchy mutation too frequently at the early stage of the search process, which may hinder from discovering promising regions. \label{item:1}
\item Ineffectiveness: MDE with a high failure threshold may execute the Cauchy mutation too rarely at the end of the search process, which may not take advantage of the Cauchy mutation to increase the convergence speed. \label{item:2}
\end{itemize}

To address these problems, we propose a sigmoid based parameter control, which gradually decreases the failure threshold as a sigmoid function of the number of generations. First, let us define two failure threshold parameters, one for the initial failure threshold $FT_{init}$ and the other for the final failure threshold $FT_{fin}$. At each generation, a new failure threshold $FT_{g}$ is obtained as follows.

\begin{equation}
FT_{g} = FT_{init} + S \Big(\frac{g}{G_{max}}\Big) \cdot (FT_{fin} - FT_{init})
\end{equation}

\noindent
where

\begin{equation}
S(x) = \frac{1}{1 + e^{-(lb + x \cdot (ub - lb))}}
\end{equation}

\noindent
where $lb$ and $ub$ are two constants, -6 and 6, respectively.

Therefore, ACM-DE assigns a high failure threshold at the beginning of the search process and gradually reduces it over generations. In doing so, ACM-DE performs the Cauchy mutation on consecutively failed individuals with a low probability at the early stage of the search process to preserve the diversity of individuals and a high probability at the late stage of the search process to increase the convergence speed.

\subsubsection{$p$-Best Individual Based Cauchy Mutation}
MDE uses the components of the best individual in the phase of the Cauchy mutation. Although consecutively failed individuals can quickly move toward the region in which the best individual is located, the population diversity of MDE may be reduced drastically. If the population diversity of MDE becomes too low, premature convergence may occur.

To address this problem, we replaced the best individual with the $p$-best individual \cite{zhang2009jade}, which uses the components of any of the top $p\%$ individuals in the phase of the Cauchy mutation. The Cauchy mutation of ACM-DE generates a trial vector as follows.

\begin{equation}
u_{i,g}^{j} = \left\{ \begin{array}{ll}
rndc_{i}^{j}(x_{pbest,g}^{j}, 0.1) & \textrm{if $rand_{i}^{j} < CR_{i,g}^{c}$ or $j = j_{rand}$} \\
x_{i,g}^{j} & \textrm{otherwise}
\end{array} \right.
\end{equation}

\noindent
where $CR_{i,g}^{c}$ denotes the crossover rate for the Cauchy mutation, which is either 0.1 or 0.9 with equal probability.

Therefore, the Cauchy mutation of ACM-DE uses the components of any of the top $p\%$ individuals to move consecutively failed individuals to the regions in which the top $p\%$ individuals are located. In doing so, the Cauchy mutation of ACM-DE can prevent the consecutively failed individuals from converging into one region, which can prevent premature convergence.

\begin{algorithm}
    \SetKwInOut{Input}{Input}
    \SetKwInOut{Output}{Output}

    \Input{Objective function $f(\vec{x})$, lower bound $\vec{x_{min}}$, upper bound $\vec{x_{max}}$, maximum number of generations $G_{max}$, initial failure threshold $FT_{init}$, final failure threshold $FT_{fin}$, scale factor $F$, crossover rate $CR$, and population size $NP$}
    \Output{Final best objective value $f(\vec{x_{best,G_{max}}})$}
    \tcc{Initialization phase}
    \For{$i = 0;\ i < NP;\ i = i + 1$}{
        \For{$j = 0;\ j < D;\ j = j + 1$}{
            $x_{i,0}^{j} = x_{min}^{j} + rand_{i}^{j} \cdot (x_{max}^{j} - x_{min}^{j})$\;
        }
        $FC_{i} = 0$\;
    }
    $g = 1$\;
    \tcc{Iteration phase}
    \While{None of termination criteria is satisfied}{
        \tcc{Recombination operator}
        $FT_{g} = FT_{init} + S \Big(\frac{g}{G_{max}}\Big) \cdot (FT_{fin} - FT_{init})$\;
        \For{$i = 0;\ i < NP;\ i = i + 1$}{
            \eIf{$FC_{i} < FT_{g}$ or $FC_{i} \bmod{FT_{g}} \ne 0$}
            {
                \tcc{DE/rand/1/bin phase}
                Carry out Algorithm \ref{alg:StandardMutation}\;
            }
            {
                \tcc{ACM/bin phase}
                Carry out Algorithm \ref{alg:AdaptiveCauchyMutation}\;
            }
        }
        \tcc{Selection operator}
        \For{$i = 0;\ i < NP;\ i = i + 1$}{
            \eIf{$f(\vec{u_{i,g}}) \leq f(\vec{x_{i,g}})$}
            {
                $\vec{x_{i,g+1}} = \vec{u_{i,g}}$\;
                $FC_{i} = 0$\;
            }
            {
                $\vec{x_{i,g+1}} = \vec{x_{i,g}}$\;
                $FC_{i} = FC_{i} + 1$\;
            }
        }
        $g = g + 1$\;
    }
    \caption{ACM-DE (DE/rand/1/bin)}
    \label{alg:ACM-DE}
\end{algorithm}

\begin{algorithm}
    Select three random donor vectors $\vec{x_{r_{1},g}}$, $\vec{x_{r_{2},g}}$, $\vec{x_{r_{3},g}}$ where $r_{1} \neq r_{2} \neq r_{3} \neq i$\;
    Select random integer $j_{rand}$ within $[1,D]$\;
    \For{$j = 0;\ j < D;\ j = j + 1$}{
        \eIf{$rand_{i}^{j} \leq CR$ or $j = j_{rand}$}
        {
            $u_{i,g}^{j} = x_{r_{1},g}^{j} + F \cdot (x_{r_{2},g}^{j} - x_{r_{3},g}^{j})$\;
        }
        {
            $u_{i,g}^{j} = x_{i,g}^{j}$\;
        }
    }
    \caption{DE/rand/1/bin}
    \label{alg:StandardMutation}
\end{algorithm}

\begin{algorithm}
    Select any of top $p\%$ individuals $\vec{x_{pBest,g}}$\;
    Select random integer $j_{rand}$ within $[1,D]$\;
    Select crossover rate $CR_{i,g}^{c}$ from either 0.1 or 0.9\;
    \For{$j = 0;\ j < D;\ j = j + 1$}{
        \eIf{$rand_{i}^{j} \leq CR_{i,g}^{c}$ or $j = j_{rand}$}
        {
            $u_{i,g}^{j} = rndc_{i}^{j}(x_{pbest,g}^{j}, 0.1)$\;
        }
        {
            $u_{i,g}^{j} = x_{i,g}^{j}$\;
        }
    }
    \caption{ACM/bin}
    \label{alg:AdaptiveCauchyMutation}
\end{algorithm}

\subsection{Combination}
\label{sec:pro_com}

ACM-DE is the combination of DE with the proposed Cauchy mutation. During the iteration phase, the standard mutation or the proposed Cauchy mutation is used to search for a global optimum. The switch between the two mutations occurs according to the failure counter of each individual $FC_{i}$. If the failure counter of an individual is equal to the failure threshold $FT_{g}$, ACM-DE performs the proposed Cauchy mutation on the individual. Otherwise, ACM-DE performs the standard mutation on the individual. The offspring generated by either the standard mutation or the proposed Cauchy mutation inherits the information from its corresponding target vector through the binomial crossover. The pseudo-code of ACM-DE is presented in Algorithm \ref{alg:ACM-DE}. Additionally, Figs. \ref{fig:CauchyMutation} and \ref{fig:SigmoidBasedParameterControl} show an illustration of the Cauchy mutation of ACM-DE compared to that of MDE and an illustration of the failure threshold decrease and increase with the sigmoid based parameter control, respectively.


\section{Experimental Setup}
\label{sec:ExperimentalSetup}

\subsection{Test Functions}
To evaluate the performance of test algorithms, we employed 30 test functions from CEC 2017 benchmark problems \cite{awad2016problem}. The CEC 2017 benchmark problems consist of three unimodal functions ($F_{1}$-$F_{3}$), seven simple multimodal functions ($F_{4}$-$F_{10}$), ten expanded multimodal functions ($F_{11}$-$F_{20}$), and ten hybrid composition functions ($F_{21}$-$F_{30}$). For more detailed information, please refer to the following paper \cite{awad2016problem}.

All experimental settings such as the number of function evaluations, the lower and upper bounds, and the termination criteria for the test functions are initialized as same as in \cite{awad2016problem}.

\subsection{Performance Metrics}
To evaluate the performance of test algorithms, we employed the following performance criteria.

\subsubsection{Function Error Value}
We employed the function error value (FEV) to evaluate the accuracy of a test algorithm. The FEV metric corresponds to the absolute difference between the global optimum of an objective function and the final best objective value of a test algorithm. The FEV metric can be defined as follows.

\begin{equation}
\textrm{FEV} = |f(\vec{x_{*}}) - f(\vec{x_{best,G_{max}}})|
\end{equation}

\noindent
where $f(\vec{x})$ denotes an objective function. Additionally, $\vec{x_{*}}$ and $\vec{x_{best,G_{max}}}$ denote the global optimum of the objective function and the final best objective value of a test algorithm, respectively. As the value of the FEV metric decreases, the accuracy of a test algorithm increases.

\subsubsection{Statistical Test}
We employed the Wilcoxon signed-rank test at a significance level of 0.05 to determine whether the performance differences between two algorithms are statistically significant \cite{demvsar2006statistical}. Hereafter, the symbols used in the tables indicate the following.

\begin{enumerate}
\item +: The proposed algorithm is significantly superior than the comparison algorithm according to the Wilcoxon signed-rank test. \label{item:1}
\item =: The difference between the proposed algorithm and the comparison algorithm is not significant according to the Wilcoxon signed-rank test. \label{item:2}
\item -: The proposed algorithm is significantly inferior than the comparison algorithm according to the Wilcoxon signed-rank test. \label{item:3}
\end{enumerate}


\section{Experimental Results}
\label{sec:ExperimentalResults}

\begin{table*}[htbp]
  \tiny
  \centering
  \caption{Averages and standard deviations of function error values with conventional DE variants ($D = 30$)}
    \begin{adjustwidth}{-0.75cm}{}
    \begin{tabular}{c|ccc||ccc||ccc}
    \toprule
          & \multicolumn{3}{c}{DE/rand/1/bin} & \multicolumn{3}{c}{DE/best/1/bin} & \multicolumn{3}{c}{DE/current-to-best/1/bin} \\
          & ACM   & CM    & Original & ACM   & CM    & Original & ACM   & CM    & Original \\
          & MEAN (STD DEV) & MEAN (STD DEV) & MEAN (STD DEV) & MEAN (STD DEV) & MEAN (STD DEV) & MEAN (STD DEV) & MEAN (STD DEV) & MEAN (STD DEV) & MEAN (STD DEV) \\
    \midrule
    F1    & 2.83E+02 (1.30E+02) & \textbf{1.75E+01 (1.23E+01) -} & 1.97E+02 (9.78E+01) - & 1.64E-14 (5.94E-15) & 1.73E-14 (5.90E-15) = & \textbf{1.61E-14 (6.97E-15) =} & 1.27E+03 (2.45E+03) & 5.26E+02 (1.87E+03) = & \textbf{2.37E+02 (7.30E+02) =} \\
    F2    & 5.07E+06 (1.63E+07) & \textbf{1.64E+05 (9.23E+05) -} & 6.11E+26 (1.18E+27) + & \textbf{1.74E+02 (8.09E+02)} & 1.29E+06 (8.09E+06) + & 2.69E+11 (6.94E+11) + & 2.12E+13 (1.19E+14) & \textbf{5.10E+11 (1.90E+12) -} & 6.83E+15 (4.86E+16) = \\
    F3    & \textbf{4.36E+03 (1.56E+03)} & 4.39E+03 (2.23E+03) = & 4.29E+04 (5.35E+03) + & 2.44E+01 (1.68E+01) & \textbf{6.88E+00 (1.03E+01) -} & 5.91E+02 (3.34E+02) + & 1.26E+01 (1.35E+01) & \textbf{6.17E+00 (1.04E+01) -} & 1.95E+02 (1.40E+02) + \\
    F4    & \textbf{8.52E+01 (2.88E-01)} & 8.86E+01 (1.55E+01) = & \textbf{8.52E+01 (2.54E-01) =} & 9.15E+01 (2.64E+01) & \textbf{8.86E+01 (2.74E+01) =} & 9.71E+01 (2.36E+01) = & 1.04E+02 (2.30E+01) & \textbf{9.86E+01 (2.11E+01) =} & 1.05E+02 (2.29E+01) = \\
    F5    & \textbf{4.06E+01 (1.15E+01)} & 5.33E+01 (1.60E+01) + & 1.67E+02 (1.08E+01) + & \textbf{5.18E+01 (1.73E+01)} & 6.68E+01 (1.92E+01) + & 7.43E+01 (5.07E+01) = & \textbf{3.93E+01 (1.13E+01)} & 5.80E+01 (2.00E+01) + & 1.41E+02 (9.94E+00) + \\
    F6    & 4.62E-12 (4.19E-12) & 5.11E-12 (6.22E-12) = & \textbf{1.11E-12 (7.26E-13) -} & \textbf{2.51E-02 (6.44E-02)} & 1.69E-01 (7.96E-01) + & 5.07E-02 (1.42E-01) = & 5.79E-03 (2.37E-02) & 2.42E-02 (1.11E-01) = & \textbf{3.24E-03 (1.61E-02) =} \\
    F7    & \textbf{6.65E+01 (8.99E+00)} & 8.74E+01 (1.53E+01) + & 2.11E+02 (8.60E+00) + & \textbf{7.75E+01 (1.57E+01)} & 9.90E+01 (1.97E+01) + & 1.72E+02 (3.42E+01) + & \textbf{6.46E+01 (1.03E+01)} & 8.09E+01 (1.50E+01) + & 1.78E+02 (9.09E+00) + \\
    F8    & \textbf{3.88E+01 (1.28E+01)} & 5.96E+01 (1.83E+01) + & 1.72E+02 (8.60E+00) + & \textbf{4.81E+01 (1.32E+01)} & 6.70E+01 (1.75E+01) + & 7.33E+01 (5.06E+01) = & \textbf{4.25E+01 (9.16E+00)} & 5.82E+01 (1.70E+01) + & 1.40E+02 (1.03E+01) + \\
    F9    & \textbf{8.94E-15 (3.10E-14)} & 1.16E-01 (5.62E-01) = & \textbf{8.94E-15 (3.10E-14) =} & \textbf{9.28E+00 (1.80E+01)} & 8.38E+01 (1.41E+02) + & 9.32E+00 (1.14E+01) = & \textbf{2.48E-01 (3.67E-01)} & 3.11E+00 (1.42E+01) + & 3.27E-01 (5.45E-01) = \\
    F10   & \textbf{2.31E+03 (5.94E+02)} & 2.97E+03 (5.70E+02) + & 6.62E+03 (2.67E+02) + & \textbf{2.83E+03 (5.93E+02)} & 2.95E+03 (6.14E+02) = & 6.08E+03 (5.42E+02) + & \textbf{2.44E+03 (6.35E+02)} & 2.91E+03 (5.76E+02) + & 6.30E+03 (3.56E+02) + \\
    F11   & \textbf{2.56E+01 (2.39E+01)} & 4.56E+01 (2.98E+01) + & 1.05E+02 (2.05E+01) + & 7.49E+01 (4.48E+01) & 7.32E+01 (3.93E+01) = & \textbf{6.96E+01 (3.57E+01) =} & \textbf{5.61E+01 (3.02E+01)} & 7.96E+01 (3.44E+01) + & 8.28E+01 (3.08E+01) + \\
    F12   & 1.30E+05 (7.87E+04) & \textbf{6.42E+04 (2.81E+04) -} & 5.69E+06 (1.75E+06) + & \textbf{2.20E+04 (1.31E+04)} & 2.41E+04 (1.43E+04) = & \textbf{2.20E+04 (1.24E+04) =} & 2.90E+04 (2.62E+04) & \textbf{2.21E+04 (1.55E+04) =} & 3.04E+04 (2.53E+04) = \\
    F13   & 1.26E+04 (5.07E+03) & \textbf{1.07E+04 (1.02E+04) =} & 2.51E+04 (1.03E+04) + & \textbf{5.58E+03 (8.30E+03)} & 8.94E+03 (1.30E+04) = & 7.16E+03 (1.23E+04) = & \textbf{4.78E+03 (2.27E+03)} & 8.02E+03 (4.68E+03) + & 6.33E+03 (3.30E+03) + \\
    F14   & \textbf{9.97E+01 (9.62E+00)} & 1.31E+02 (4.30E+01) + & 1.19E+02 (8.27E+00) + & 8.89E+01 (2.67E+01) & \textbf{8.52E+01 (5.23E+01) =} & 9.86E+01 (2.56E+01) = & \textbf{1.06E+02 (2.62E+01)} & 1.12E+02 (3.29E+01) = & 1.18E+02 (2.79E+01) + \\
    F15   & \textbf{2.00E+02 (3.39E+01)} & 3.39E+03 (5.94E+03) + & 3.45E+02 (5.36E+01) + & \textbf{1.13E+02 (6.03E+01)} & 1.46E+02 (1.67E+02) = & 1.44E+02 (9.64E+01) = & \textbf{1.85E+02 (6.16E+01)} & 2.63E+02 (1.09E+02) + & 2.44E+02 (9.22E+01) + \\
    F16   & \textbf{6.34E+02 (2.61E+02)} & 7.78E+02 (2.66E+02) + & 1.06E+03 (1.38E+02) + & 5.60E+02 (2.33E+02) & 7.49E+02 (2.49E+02) + & \textbf{4.38E+02 (2.89E+02) -} & \textbf{4.45E+02 (1.96E+02)} & 6.88E+02 (2.81E+02) + & 7.65E+02 (2.56E+02) + \\
    F17   & \textbf{7.54E+01 (6.63E+01)} & 2.51E+02 (1.54E+02) + & 2.00E+02 (3.14E+01) + & \textbf{1.89E+02 (1.04E+02)} & 3.07E+02 (1.54E+02) + & 1.95E+02 (1.02E+02) = & \textbf{1.18E+02 (6.11E+01)} & 2.07E+02 (1.12E+02) + & 2.01E+02 (6.84E+01) + \\
    F18   & \textbf{1.09E+05 (7.75E+04)} & 1.52E+05 (1.40E+05) = & 5.76E+05 (1.78E+05) + & \textbf{6.08E+04 (6.57E+04)} & 6.11E+04 (5.24E+04) = & 6.99E+04 (4.27E+04) + & \textbf{4.03E+04 (2.36E+04)} & 7.41E+04 (4.75E+04) + & 6.11E+04 (2.71E+04) + \\
    F19   & \textbf{9.19E+01 (1.83E+01)} & 4.78E+03 (7.62E+03) + & 1.22E+02 (2.95E+01) + & \textbf{5.86E+01 (5.19E+01)} & 7.87E+01 (5.57E+01) + & 8.05E+01 (6.16E+01) + & \textbf{8.49E+01 (3.33E+01)} & 1.06E+02 (4.25E+01) + & 9.33E+01 (3.68E+01) = \\
    F20   & \textbf{7.37E+01 (7.57E+01)} & 3.17E+02 (1.48E+02) + & 2.15E+02 (4.60E+01) + & \textbf{1.59E+02 (1.03E+02)} & 3.64E+02 (1.66E+02) + & 1.75E+02 (9.57E+01) = & \textbf{1.36E+02 (6.25E+01)} & 2.72E+02 (1.31E+02) + & 2.17E+02 (7.26E+01) + \\
    F21   & \textbf{2.41E+02 (1.16E+01)} & 2.57E+02 (1.74E+01) + & 3.66E+02 (1.15E+01) + & \textbf{2.51E+02 (1.36E+01)} & 2.68E+02 (1.79E+01) + & 2.67E+02 (4.37E+01) = & \textbf{2.38E+02 (1.11E+01)} & 2.57E+02 (1.66E+01) + & 3.35E+02 (1.27E+01) + \\
    F22   & \textbf{1.00E+02 (0.00E+00)} & \textbf{1.00E+02 (0.00E+00) =} & \textbf{1.00E+02 (0.00E+00) =} & 1.66E+03 (1.63E+03) & \textbf{1.39E+03 (1.60E+03) =} & 2.68E+03 (2.74E+03) = & \textbf{1.00E+02 (9.48E-01)} & 4.04E+02 (9.33E+02) = & 3.55E+02 (1.27E+03) = \\
    F23   & \textbf{3.83E+02 (1.09E+01)} & 4.05E+02 (1.51E+01) + & 5.16E+02 (8.74E+00) + & 4.03E+02 (2.01E+01) & 4.23E+02 (2.15E+01) + & \textbf{4.00E+02 (1.74E+01) =} & \textbf{3.90E+02 (1.43E+01)} & 4.15E+02 (2.47E+01) + & 4.67E+02 (2.68E+01) + \\
    F24   & \textbf{4.62E+02 (1.28E+01)} & 4.75E+02 (1.45E+01) + & 5.87E+02 (9.52E+00) + & \textbf{4.74E+02 (1.64E+01)} & 4.91E+02 (2.33E+01) + & 4.76E+02 (2.03E+01) = & \textbf{4.66E+02 (1.64E+01)} & 4.83E+02 (2.19E+01) + & 5.48E+02 (1.60E+01) + \\
    F25   & \textbf{3.87E+02 (0.00E+00)} & \textbf{3.87E+02 (7.84E-01) =} & \textbf{3.87E+02 (0.00E+00) =} & 3.94E+02 (1.34E+01) & 3.95E+02 (1.53E+01) = & \textbf{3.92E+02 (1.04E+01) =} & 4.03E+02 (1.61E+01) & \textbf{3.95E+02 (1.25E+01) -} & 4.05E+02 (1.61E+01) = \\
    F26   & \textbf{1.35E+03 (1.32E+02)} & 1.57E+03 (2.15E+02) + & 2.56E+03 (1.06E+02) + & 1.56E+03 (4.63E+02) & 1.67E+03 (5.12E+02) + & \textbf{1.44E+03 (4.59E+02) =} & \textbf{1.30E+03 (3.72E+02)} & 1.44E+03 (5.57E+02) + & 1.77E+03 (5.70E+02) + \\
    F27   & \textbf{5.01E+02 (5.28E+00)} & 5.12E+02 (7.02E+00) + & 5.07E+02 (9.98E+00) + & \textbf{5.19E+02 (1.36E+01)} & 5.20E+02 (1.32E+01) = & 5.20E+02 (1.44E+01) = & 5.17E+02 (1.20E+01) & 5.20E+02 (1.13E+01) = & \textbf{5.16E+02 (1.20E+01) =} \\
    F28   & 3.64E+02 (4.85E+01) & \textbf{3.58E+02 (6.14E+01) =} & 4.03E+02 (2.17E+01) + & 4.30E+02 (2.72E+01) & \textbf{4.21E+02 (3.55E+01) =} & 4.36E+02 (2.65E+01) = & 4.68E+02 (2.93E+01) & \textbf{4.43E+02 (2.32E+01) -} & 4.64E+02 (3.61E+01) = \\
    F29   & \textbf{4.69E+02 (8.13E+01)} & 6.68E+02 (1.39E+02) + & 8.91E+02 (7.45E+01) + & 6.07E+02 (1.32E+02) & 6.72E+02 (1.71E+02) + & \textbf{5.91E+02 (9.45E+01) =} & \textbf{5.25E+02 (6.62E+01)} & 5.95E+02 (1.08E+02) + & 7.12E+02 (8.46E+01) + \\
    F30   & 1.05E+04 (2.22E+03) & \textbf{7.12E+03 (3.18E+03) -} & 4.15E+04 (1.09E+04) + & \textbf{6.10E+03 (3.02E+03)} & 6.42E+03 (3.46E+03) = & 6.91E+03 (3.61E+03) = & \textbf{8.07E+03 (4.11E+03)} & 9.80E+03 (7.68E+03) = & 1.10E+04 (8.03E+03) + \\
    \midrule
    +/=/- &       & 17/9/4 & 24/4/2 &       & 15/14/1 & 6/23/1 &       & 18/8/4 & 19/11/0 \\
    \midrule
    \midrule
          & \multicolumn{3}{c}{DE/current-to-rand/1} & \multicolumn{3}{c}{DE/rand/2/bin} & \multicolumn{3}{c}{DE/current-to-best/2/bin} \\
          & ACM   & CM    & Original & ACM   & CM    & Original & ACM   & CM    & Original \\
          & MEAN (STD DEV) & MEAN (STD DEV) & MEAN (STD DEV) & MEAN (STD DEV) & MEAN (STD DEV) & MEAN (STD DEV) & MEAN (STD DEV) & MEAN (STD DEV) & MEAN (STD DEV) \\
    \midrule
    F1    & 2.83E+02 (1.30E+02) & \textbf{1.75E+01 (1.23E+01) -} & 1.97E+02 (9.78E+01) - & 1.64E-14 (5.94E-15) & 1.73E-14 (5.90E-15) = & \textbf{1.61E-14 (6.97E-15) =} & 1.27E+03 (2.45E+03) & 5.26E+02 (1.87E+03) = & \textbf{2.37E+02 (7.30E+02) =} \\
    F2    & 5.07E+06 (1.63E+07) & \textbf{1.64E+05 (9.23E+05) -} & 6.11E+26 (1.18E+27) + & \textbf{1.74E+02 (8.09E+02)} & 1.29E+06 (8.09E+06) + & 2.69E+11 (6.94E+11) + & 2.12E+13 (1.19E+14) & \textbf{5.10E+11 (1.90E+12) -} & 6.83E+15 (4.86E+16) = \\
    F3    & \textbf{4.36E+03 (1.56E+03)} & 4.39E+03 (2.23E+03) = & 4.29E+04 (5.35E+03) + & 2.44E+01 (1.68E+01) & \textbf{6.88E+00 (1.03E+01) -} & 5.91E+02 (3.34E+02) + & 1.26E+01 (1.35E+01) & \textbf{6.17E+00 (1.04E+01) -} & 1.95E+02 (1.40E+02) + \\
    F4    & \textbf{8.52E+01 (2.88E-01)} & 8.86E+01 (1.55E+01) = & \textbf{8.52E+01 (2.54E-01) =} & 9.15E+01 (2.64E+01) & \textbf{8.86E+01 (2.74E+01) =} & 9.71E+01 (2.36E+01) = & 1.04E+02 (2.30E+01) & \textbf{9.86E+01 (2.11E+01) =} & 1.05E+02 (2.29E+01) = \\
    F5    & \textbf{4.06E+01 (1.15E+01)} & 5.33E+01 (1.60E+01) + & 1.67E+02 (1.08E+01) + & \textbf{5.18E+01 (1.73E+01)} & 6.68E+01 (1.92E+01) + & 7.43E+01 (5.07E+01) = & \textbf{3.93E+01 (1.13E+01)} & 5.80E+01 (2.00E+01) + & 1.41E+02 (9.94E+00) + \\
    F6    & 4.62E-12 (4.19E-12) & 5.11E-12 (6.22E-12) = & \textbf{1.11E-12 (7.26E-13) -} & \textbf{2.51E-02 (6.44E-02)} & 1.69E-01 (7.96E-01) + & 5.07E-02 (1.42E-01) = & 5.79E-03 (2.37E-02) & 2.42E-02 (1.11E-01) = & \textbf{3.24E-03 (1.61E-02) =} \\
    F7    & \textbf{6.65E+01 (8.99E+00)} & 8.74E+01 (1.53E+01) + & 2.11E+02 (8.60E+00) + & \textbf{7.75E+01 (1.57E+01)} & 9.90E+01 (1.97E+01) + & 1.72E+02 (3.42E+01) + & \textbf{6.46E+01 (1.03E+01)} & 8.09E+01 (1.50E+01) + & 1.78E+02 (9.09E+00) + \\
    F8    & \textbf{3.88E+01 (1.28E+01)} & 5.96E+01 (1.83E+01) + & 1.72E+02 (8.60E+00) + & \textbf{4.81E+01 (1.32E+01)} & 6.70E+01 (1.75E+01) + & 7.33E+01 (5.06E+01) = & \textbf{4.25E+01 (9.16E+00)} & 5.82E+01 (1.70E+01) + & 1.40E+02 (1.03E+01) + \\
    F9    & \textbf{8.94E-15 (3.10E-14)} & 1.16E-01 (5.62E-01) = & \textbf{8.94E-15 (3.10E-14) =} & \textbf{9.28E+00 (1.80E+01)} & 8.38E+01 (1.41E+02) + & 9.32E+00 (1.14E+01) = & \textbf{2.48E-01 (3.67E-01)} & 3.11E+00 (1.42E+01) + & 3.27E-01 (5.45E-01) = \\
    F10   & \textbf{2.31E+03 (5.94E+02)} & 2.97E+03 (5.70E+02) + & 6.62E+03 (2.67E+02) + & \textbf{2.83E+03 (5.93E+02)} & 2.95E+03 (6.14E+02) = & 6.08E+03 (5.42E+02) + & \textbf{2.44E+03 (6.35E+02)} & 2.91E+03 (5.76E+02) + & 6.30E+03 (3.56E+02) + \\
    F11   & \textbf{2.56E+01 (2.39E+01)} & 4.56E+01 (2.98E+01) + & 1.05E+02 (2.05E+01) + & 7.49E+01 (4.48E+01) & 7.32E+01 (3.93E+01) = & \textbf{6.96E+01 (3.57E+01) =} & \textbf{5.61E+01 (3.02E+01)} & 7.96E+01 (3.44E+01) + & 8.28E+01 (3.08E+01) + \\
    F12   & 1.30E+05 (7.87E+04) & \textbf{6.42E+04 (2.81E+04) -} & 5.69E+06 (1.75E+06) + & \textbf{2.20E+04 (1.31E+04)} & 2.41E+04 (1.43E+04) = & \textbf{2.20E+04 (1.24E+04) =} & 2.90E+04 (2.62E+04) & \textbf{2.21E+04 (1.55E+04) =} & 3.04E+04 (2.53E+04) = \\
    F13   & 1.26E+04 (5.07E+03) & \textbf{1.07E+04 (1.02E+04) =} & 2.51E+04 (1.03E+04) + & \textbf{5.58E+03 (8.30E+03)} & 8.94E+03 (1.30E+04) = & 7.16E+03 (1.23E+04) = & \textbf{4.78E+03 (2.27E+03)} & 8.02E+03 (4.68E+03) + & 6.33E+03 (3.30E+03) + \\
    F14   & \textbf{9.97E+01 (9.62E+00)} & 1.31E+02 (4.30E+01) + & 1.19E+02 (8.27E+00) + & 8.89E+01 (2.67E+01) & \textbf{8.52E+01 (5.23E+01) =} & 9.86E+01 (2.56E+01) = & \textbf{1.06E+02 (2.62E+01)} & 1.12E+02 (3.29E+01) = & 1.18E+02 (2.79E+01) + \\
    F15   & \textbf{2.00E+02 (3.39E+01)} & 3.39E+03 (5.94E+03) + & 3.45E+02 (5.36E+01) + & \textbf{1.13E+02 (6.03E+01)} & 1.46E+02 (1.67E+02) = & 1.44E+02 (9.64E+01) = & \textbf{1.85E+02 (6.16E+01)} & 2.63E+02 (1.09E+02) + & 2.44E+02 (9.22E+01) + \\
    F16   & \textbf{6.34E+02 (2.61E+02)} & 7.78E+02 (2.66E+02) + & 1.06E+03 (1.38E+02) + & 5.60E+02 (2.33E+02) & 7.49E+02 (2.49E+02) + & \textbf{4.38E+02 (2.89E+02) -} & \textbf{4.45E+02 (1.96E+02)} & 6.88E+02 (2.81E+02) + & 7.65E+02 (2.56E+02) + \\
    F17   & \textbf{7.54E+01 (6.63E+01)} & 2.51E+02 (1.54E+02) + & 2.00E+02 (3.14E+01) + & \textbf{1.89E+02 (1.04E+02)} & 3.07E+02 (1.54E+02) + & 1.95E+02 (1.02E+02) = & \textbf{1.18E+02 (6.11E+01)} & 2.07E+02 (1.12E+02) + & 2.01E+02 (6.84E+01) + \\
    F18   & \textbf{1.09E+05 (7.75E+04)} & 1.52E+05 (1.40E+05) = & 5.76E+05 (1.78E+05) + & \textbf{6.08E+04 (6.57E+04)} & 6.11E+04 (5.24E+04) = & 6.99E+04 (4.27E+04) + & \textbf{4.03E+04 (2.36E+04)} & 7.41E+04 (4.75E+04) + & 6.11E+04 (2.71E+04) + \\
    F19   & \textbf{9.19E+01 (1.83E+01)} & 4.78E+03 (7.62E+03) + & 1.22E+02 (2.95E+01) + & \textbf{5.86E+01 (5.19E+01)} & 7.87E+01 (5.57E+01) + & 8.05E+01 (6.16E+01) + & \textbf{8.49E+01 (3.33E+01)} & 1.06E+02 (4.25E+01) + & 9.33E+01 (3.68E+01) = \\
    F20   & \textbf{7.37E+01 (7.57E+01)} & 3.17E+02 (1.48E+02) + & 2.15E+02 (4.60E+01) + & \textbf{1.59E+02 (1.03E+02)} & 3.64E+02 (1.66E+02) + & 1.75E+02 (9.57E+01) = & \textbf{1.36E+02 (6.25E+01)} & 2.72E+02 (1.31E+02) + & 2.17E+02 (7.26E+01) + \\
    F21   & \textbf{2.41E+02 (1.16E+01)} & 2.57E+02 (1.74E+01) + & 3.66E+02 (1.15E+01) + & \textbf{2.51E+02 (1.36E+01)} & 2.68E+02 (1.79E+01) + & 2.67E+02 (4.37E+01) = & \textbf{2.38E+02 (1.11E+01)} & 2.57E+02 (1.66E+01) + & 3.35E+02 (1.27E+01) + \\
    F22   & \textbf{1.00E+02 (0.00E+00)} & \textbf{1.00E+02 (0.00E+00) =} & \textbf{1.00E+02 (0.00E+00) =} & 1.66E+03 (1.63E+03) & \textbf{1.39E+03 (1.60E+03) =} & 2.68E+03 (2.74E+03) = & \textbf{1.00E+02 (9.48E-01)} & 4.04E+02 (9.33E+02) = & 3.55E+02 (1.27E+03) = \\
    F23   & \textbf{3.83E+02 (1.09E+01)} & 4.05E+02 (1.51E+01) + & 5.16E+02 (8.74E+00) + & 4.03E+02 (2.01E+01) & 4.23E+02 (2.15E+01) + & \textbf{4.00E+02 (1.74E+01) =} & \textbf{3.90E+02 (1.43E+01)} & 4.15E+02 (2.47E+01) + & 4.67E+02 (2.68E+01) + \\
    F24   & \textbf{4.62E+02 (1.28E+01)} & 4.75E+02 (1.45E+01) + & 5.87E+02 (9.52E+00) + & \textbf{4.74E+02 (1.64E+01)} & 4.91E+02 (2.33E+01) + & 4.76E+02 (2.03E+01) = & \textbf{4.66E+02 (1.64E+01)} & 4.83E+02 (2.19E+01) + & 5.48E+02 (1.60E+01) + \\
    F25   & \textbf{3.87E+02 (0.00E+00)} & \textbf{3.87E+02 (7.84E-01) =} & \textbf{3.87E+02 (0.00E+00) =} & 3.94E+02 (1.34E+01) & 3.95E+02 (1.53E+01) = & \textbf{3.92E+02 (1.04E+01) =} & 4.03E+02 (1.61E+01) & \textbf{3.95E+02 (1.25E+01) -} & 4.05E+02 (1.61E+01) = \\
    F26   & \textbf{1.35E+03 (1.32E+02)} & 1.57E+03 (2.15E+02) + & 2.56E+03 (1.06E+02) + & 1.56E+03 (4.63E+02) & 1.67E+03 (5.12E+02) + & \textbf{1.44E+03 (4.59E+02) =} & \textbf{1.30E+03 (3.72E+02)} & 1.44E+03 (5.57E+02) + & 1.77E+03 (5.70E+02) + \\
    F27   & \textbf{5.01E+02 (5.28E+00)} & 5.12E+02 (7.02E+00) + & 5.07E+02 (9.98E+00) + & \textbf{5.19E+02 (1.36E+01)} & 5.20E+02 (1.32E+01) = & 5.20E+02 (1.44E+01) = & 5.17E+02 (1.20E+01) & 5.20E+02 (1.13E+01) = & \textbf{5.16E+02 (1.20E+01) =} \\
    F28   & 3.64E+02 (4.85E+01) & \textbf{3.58E+02 (6.14E+01) =} & 4.03E+02 (2.17E+01) + & 4.30E+02 (2.72E+01) & \textbf{4.21E+02 (3.55E+01) =} & 4.36E+02 (2.65E+01) = & 4.68E+02 (2.93E+01) & \textbf{4.43E+02 (2.32E+01) -} & 4.64E+02 (3.61E+01) = \\
    F29   & \textbf{4.69E+02 (8.13E+01)} & 6.68E+02 (1.39E+02) + & 8.91E+02 (7.45E+01) + & 6.07E+02 (1.32E+02) & 6.72E+02 (1.71E+02) + & \textbf{5.91E+02 (9.45E+01) =} & \textbf{5.25E+02 (6.62E+01)} & 5.95E+02 (1.08E+02) + & 7.12E+02 (8.46E+01) + \\
    F30   & 1.05E+04 (2.22E+03) & \textbf{7.12E+03 (3.18E+03) -} & 4.15E+04 (1.09E+04) + & \textbf{6.10E+03 (3.02E+03)} & 6.42E+03 (3.46E+03) = & 6.91E+03 (3.61E+03) = & \textbf{8.07E+03 (4.11E+03)} & 9.80E+03 (7.68E+03) = & 1.10E+04 (8.03E+03) + \\
    \midrule
    +/=/- &       & 21/6/3 & 22/8/0 &       & 14/8/8 & 28/2/0 &       & 20/6/4 & 25/3/2 \\
    \bottomrule
    \end{tabular}%
    \end{adjustwidth}
  \label{tab:conventional_d30}%
  The symbols "+/=/-" show the statistical results of the Wilcoxon signed-rank test with $\alpha = 0.05$ significance level. "+" represents that ACM variant is significantly superior than the corresponding variant. "=" represents that the performance difference between ACM variant and the corresponding variant is not statistically significant. And, "-" represents that ACM variant is significantly inferior than the corresponding variant.
\end{table*}%

\begin{figure*}[htbp]
 \centering
 \subfigure[$F_{10}$]{
  \includegraphics[scale=0.25]{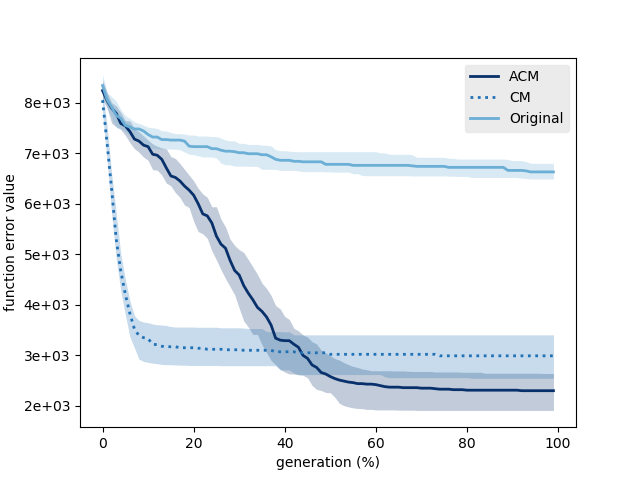}
   }
 \subfigure[$F_{16}$]{
  \includegraphics[scale=0.25]{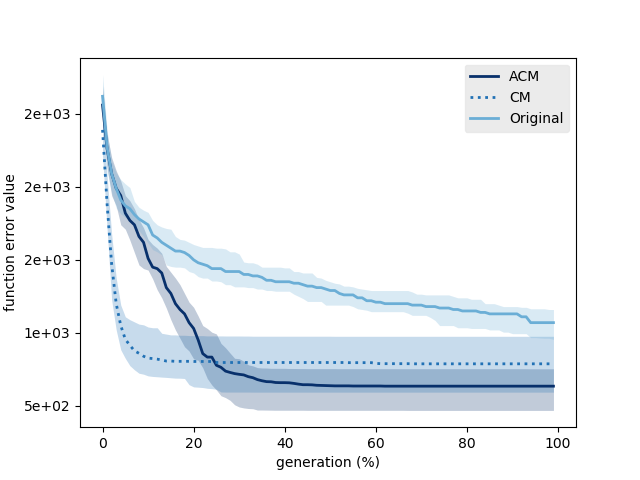}
   }
 \subfigure[$F_{20}$]{
  \includegraphics[scale=0.25]{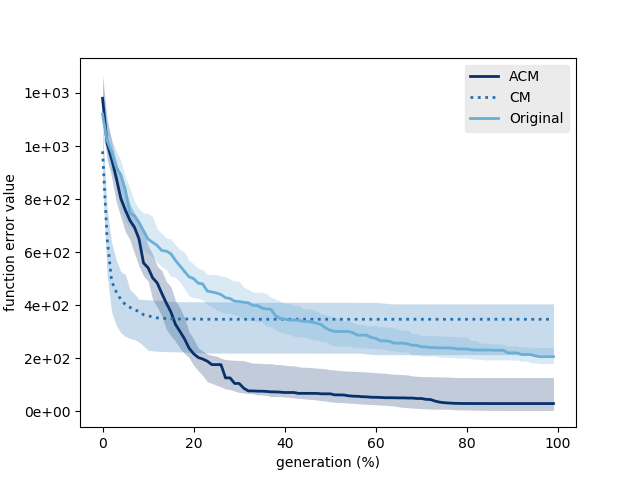}
   }
 \subfigure[$F_{26}$]{
  \includegraphics[scale=0.25]{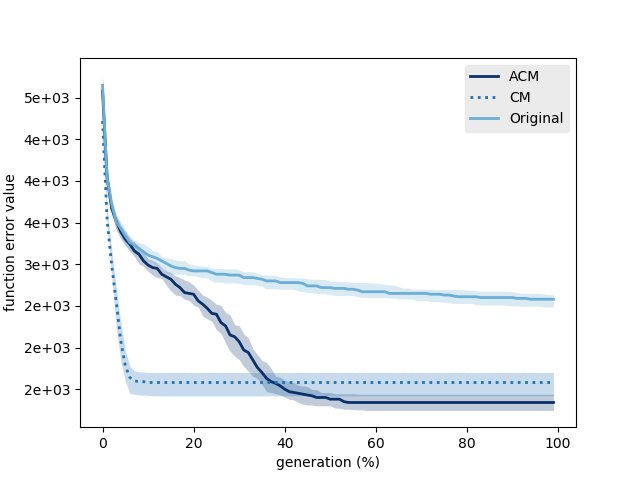}
   }
 \caption[]{Median and interquartile ranges (25th and 75th) of function error values with DE/rand/1/bin variants ($D = 30$)}
 \label{fig:convergenceGraphs_rand1bin_d30}
\end{figure*}

\begin{table*}[htbp]
  \tiny
  \centering
  \caption{Averages and standard deviations of function error values with conventional DE variants ($D = 50$)}
    \begin{adjustwidth}{-0.75cm}{}
    \begin{tabular}{c|ccc||ccc||ccc}
    \toprule
          & \multicolumn{3}{c}{DE/rand/1/bin} & \multicolumn{3}{c}{DE/best/1/bin} & \multicolumn{3}{c}{DE/current-to-best/1/bin} \\
          & ACM   & CM    & Original & ACM   & CM    & Original & ACM   & CM    & Original \\
          & MEAN (STD DEV) & MEAN (STD DEV) & MEAN (STD DEV) & MEAN (STD DEV) & MEAN (STD DEV) & MEAN (STD DEV) & MEAN (STD DEV) & MEAN (STD DEV) & MEAN (STD DEV) \\
    \midrule
    F1    & 3.44E+03 (3.53E+03) & 5.20E+03 (6.39E+03) = & \textbf{2.54E+03 (2.86E+03) =} & 5.60E+03 (7.28E+03) & 5.94E+03 (6.99E+03) = & \textbf{5.15E+03 (5.92E+03) =} & 1.01E+07 (1.83E+07) & \textbf{1.03E+06 (5.37E+05) -} & 2.92E+07 (3.66E+07) + \\
    F2    & 2.19E+16 (9.76E+16) & \textbf{2.25E+13 (1.27E+14) -} & 4.74E+57 (9.34E+57) + & \textbf{8.58E+11 (3.93E+12)} & 1.27E+15 (7.11E+15) + & 1.40E+28 (6.87E+28) + & 8.01E+31 (2.34E+32) & \textbf{1.40E+29 (8.57E+29) -} & 9.11E+37 (6.34E+38) + \\
    F3    & 5.20E+04 (1.08E+04) & \textbf{3.86E+04 (9.28E+03) -} & 1.17E+05 (1.44E+04) + & 6.45E+03 (2.40E+03) & \textbf{4.92E+03 (3.05E+03) -} & 1.55E+04 (4.13E+03) + & 4.22E+03 (1.98E+03) & \textbf{3.11E+03 (1.58E+03) -} & 8.67E+03 (2.63E+03) + \\
    F4    & \textbf{7.89E+01 (4.96E+01)} & 1.08E+02 (5.60E+01) + & 1.08E+02 (3.81E+01) + & 1.79E+02 (3.96E+01) & \textbf{1.50E+02 (4.96E+01) -} & 1.99E+02 (5.47E+01) = & 2.53E+02 (4.62E+01) & \textbf{1.91E+02 (4.13E+01) -} & 2.65E+02 (4.86E+01) = \\
    F5    & \textbf{7.91E+01 (1.92E+01)} & 1.15E+02 (2.89E+01) + & 3.65E+02 (1.45E+01) + & \textbf{1.09E+02 (2.47E+01)} & 1.46E+02 (3.59E+01) + & 1.46E+02 (9.60E+01) = & \textbf{8.90E+01 (1.62E+01)} & 1.23E+02 (2.92E+01) + & 3.19E+02 (1.65E+01) + \\
    F6    & 1.75E-05 (1.48E-06) & \textbf{1.33E-05 (2.43E-06) -} & 1.34E-05 (2.82E-06) - & \textbf{6.28E-01 (8.91E-01)} & 1.74E+00 (2.58E+00) + & 6.77E-01 (8.50E-01) = & 5.59E-02 (1.22E-01) & 6.80E-01 (1.50E+00) + & \textbf{5.19E-02 (1.10E-01) =} \\
    F7    & \textbf{1.20E+02 (1.84E+01)} & 1.71E+02 (3.21E+01) + & 4.39E+02 (1.14E+01) + & \textbf{1.69E+02 (2.49E+01)} & 2.11E+02 (4.81E+01) + & 3.33E+02 (9.76E+01) + & \textbf{1.37E+02 (2.07E+01)} & 1.65E+02 (2.78E+01) + & 3.80E+02 (2.08E+01) + \\
    F8    & \textbf{7.35E+01 (1.88E+01)} & 1.11E+02 (2.23E+01) + & 3.67E+02 (1.84E+01) + & \textbf{1.12E+02 (2.75E+01)} & 1.43E+02 (4.07E+01) + & 1.30E+02 (8.97E+01) = & \textbf{8.70E+01 (1.65E+01)} & 1.25E+02 (2.96E+01) + & 3.18E+02 (1.87E+01) + \\
    F9    & \textbf{7.78E-13 (9.28E-13)} & 5.57E-01 (1.67E+00) = & 8.76E-13 (1.05E-12) = & \textbf{2.36E+02 (2.94E+02)} & 1.82E+03 (1.81E+03) + & 3.73E+02 (6.14E+02) = & \textbf{7.88E+00 (6.89E+00)} & 5.06E+02 (6.70E+02) + & 9.99E+00 (1.05E+01) = \\
    F10   & \textbf{4.18E+03 (7.92E+02)} & 5.37E+03 (8.12E+02) + & 1.28E+04 (3.96E+02) + & \textbf{5.09E+03 (8.77E+02)} & 5.40E+03 (8.12E+02) = & 1.21E+04 (6.44E+02) + & \textbf{4.59E+03 (9.57E+02)} & 5.53E+03 (5.90E+02) + & 1.23E+04 (4.08E+02) + \\
    F11   & \textbf{5.18E+01 (9.38E+00)} & 9.71E+01 (2.54E+01) + & 1.92E+02 (9.60E+00) + & \textbf{1.49E+02 (4.83E+01)} & 1.82E+02 (4.70E+01) + & 1.63E+02 (5.92E+01) = & \textbf{1.37E+02 (3.85E+01)} & 1.88E+02 (5.12E+01) + & 1.43E+02 (4.50E+01) = \\
    F12   & 2.08E+06 (9.71E+05) & \textbf{1.95E+06 (1.05E+06) =} & 7.04E+07 (1.98E+07) + & 5.32E+05 (5.16E+05) & 8.51E+05 (8.84E+05) = & \textbf{4.94E+05 (3.37E+05) =} & \textbf{3.52E+05 (2.84E+05)} & 1.19E+06 (2.59E+06) + & 1.51E+06 (3.70E+06) = \\
    F13   & 2.12E+04 (1.03E+04) & \textbf{4.51E+03 (5.32E+03) -} & 4.84E+04 (4.32E+04) + & 9.70E+03 (7.83E+03) & \textbf{8.16E+03 (7.30E+03) =} & 8.86E+03 (7.99E+03) = & \textbf{8.52E+03 (4.43E+03)} & 1.29E+04 (6.94E+03) + & 1.26E+04 (6.13E+03) + \\
    F14   & \textbf{1.91E+04 (1.81E+04)} & 4.10E+04 (3.61E+04) + & 1.29E+05 (3.78E+04) + & \textbf{4.50E+03 (5.95E+03)} & 1.15E+04 (1.73E+04) + & 6.18E+03 (1.20E+04) = & \textbf{2.54E+03 (2.42E+03)} & 1.03E+04 (1.32E+04) + & 4.43E+03 (7.10E+03) = \\
    F15   & 3.86E+03 (1.62E+03) & \textbf{3.59E+03 (3.89E+03) =} & 9.12E+03 (2.97E+03) + & 5.56E+03 (5.49E+03) & \textbf{5.08E+03 (5.48E+03) =} & 5.31E+03 (5.54E+03) = & \textbf{1.88E+03 (9.16E+02)} & 3.62E+03 (2.17E+03) + & 3.30E+03 (1.75E+03) + \\
    F16   & \textbf{1.31E+03 (3.96E+02)} & 1.47E+03 (3.97E+02) = & 2.82E+03 (1.87E+02) + & \textbf{1.01E+03 (4.51E+02)} & 1.39E+03 (4.51E+02) + & 1.02E+03 (5.82E+02) = & \textbf{1.05E+03 (3.57E+02)} & 1.33E+03 (3.58E+02) + & 1.74E+03 (3.43E+02) + \\
    F17   & \textbf{8.18E+02 (2.41E+02)} & 1.15E+03 (2.74E+02) + & 1.54E+03 (1.43E+02) + & 8.88E+02 (2.87E+02) & 1.20E+03 (2.97E+02) + & \textbf{8.23E+02 (4.01E+02) =} & \textbf{8.21E+02 (2.34E+02)} & 1.02E+03 (2.66E+02) + & 1.14E+03 (1.84E+02) + \\
    F18   & 4.53E+05 (3.20E+05) & \textbf{3.72E+05 (2.17E+05) =} & 4.66E+06 (1.27E+06) + & \textbf{2.40E+05 (1.87E+05)} & 4.03E+05 (3.07E+05) + & 6.14E+05 (3.45E+05) + & \textbf{1.48E+05 (7.92E+04)} & 3.21E+05 (1.93E+05) + & 4.33E+05 (2.15E+05) + \\
    F19   & \textbf{7.67E+03 (3.32E+03)} & 1.47E+04 (9.40E+03) + & 9.44E+03 (3.95E+03) + & \textbf{2.20E+03 (4.31E+03)} & 4.94E+03 (8.67E+03) + & 6.63E+03 (9.94E+03) + & 1.47E+03 (2.58E+03) & 3.21E+03 (2.56E+03) + & \textbf{9.13E+02 (6.44E+02) =} \\
    F20   & \textbf{6.29E+02 (2.34E+02)} & 9.46E+02 (3.04E+02) + & 1.34E+03 (1.14E+02) + & \textbf{6.72E+02 (2.65E+02)} & 9.27E+02 (3.27E+02) + & 8.60E+02 (3.80E+02) + & \textbf{5.64E+02 (2.49E+02)} & 8.60E+02 (2.78E+02) + & 1.05E+03 (1.78E+02) + \\
    F21   & \textbf{2.83E+02 (2.14E+01)} & 3.17E+02 (2.63E+01) + & 5.71E+02 (1.38E+01) + & \textbf{3.16E+02 (2.01E+01)} & 3.46E+02 (2.95E+01) + & 3.40E+02 (9.37E+01) = & \textbf{2.90E+02 (2.05E+01)} & 3.17E+02 (2.96E+01) + & 5.13E+02 (1.98E+01) + \\
    F22   & \textbf{4.35E+03 (1.45E+03)} & 5.84E+03 (1.64E+03) + & 1.17E+04 (3.87E+03) + & \textbf{5.46E+03 (1.04E+03)} & 5.99E+03 (1.16E+03) + & 1.21E+04 (9.99E+02) + & \textbf{4.58E+03 (1.77E+03)} & 6.03E+03 (1.76E+03) + & 9.39E+03 (5.50E+03) + \\
    F23   & \textbf{4.96E+02 (2.13E+01)} & 5.42E+02 (3.05E+01) + & 7.85E+02 (1.21E+01) + & 5.62E+02 (3.54E+01) & 5.93E+02 (3.96E+01) + & \textbf{5.58E+02 (3.65E+01) =} & \textbf{5.35E+02 (2.84E+01)} & 5.84E+02 (3.86E+01) + & 6.88E+02 (7.84E+01) + \\
    F24   & \textbf{5.79E+02 (2.21E+01)} & 5.98E+02 (2.68E+01) + & 8.50E+02 (1.38E+01) + & \textbf{6.31E+02 (3.80E+01)} & 6.61E+02 (4.63E+01) + & 6.36E+02 (5.78E+01) = & \textbf{6.09E+02 (2.82E+01)} & 6.55E+02 (4.51E+01) + & 7.31E+02 (9.35E+01) + \\
    F25   & \textbf{4.80E+02 (0.00E+00)} & 5.13E+02 (3.30E+01) + & \textbf{4.80E+02 (0.00E+00) =} & 5.73E+02 (3.37E+01) & \textbf{5.52E+02 (2.96E+01) -} & 5.64E+02 (3.27E+01) = & 6.37E+02 (4.34E+01) & \textbf{5.90E+02 (3.35E+01) -} & 6.49E+02 (4.66E+01) = \\
    F26   & \textbf{1.86E+03 (2.43E+02)} & 2.28E+03 (2.62E+02) + & 4.66E+03 (1.18E+02) + & 2.49E+03 (3.73E+02) & 2.90E+03 (5.32E+02) + & \textbf{2.38E+03 (3.24E+02) =} & \textbf{2.18E+03 (3.72E+02)} & 2.89E+03 (3.77E+02) + & 2.57E+03 (7.95E+02) = \\
    F27   & 5.08E+02 (7.97E+00) & 6.05E+02 (4.76E+01) + & \textbf{5.06E+02 (9.52E+00) =} & \textbf{6.46E+02 (5.86E+01)} & 6.56E+02 (7.97E+01) = & 6.54E+02 (7.32E+01) = & 6.89E+02 (7.86E+01) & 6.90E+02 (6.65E+01) = & \textbf{6.75E+02 (6.75E+01) =} \\
    F28   & \textbf{4.59E+02 (0.00E+00)} & 4.93E+02 (2.11E+01) + & \textbf{4.59E+02 (0.00E+00) =} & 5.69E+02 (4.09E+01) & \textbf{5.25E+02 (2.63E+01) -} & 5.89E+02 (6.35E+01) = & 7.62E+02 (8.93E+01) & \textbf{6.10E+02 (4.92E+01) -} & 7.93E+02 (1.11E+02) = \\
    F29   & \textbf{5.82E+02 (2.11E+02)} & 9.25E+02 (2.41E+02) + & 1.63E+03 (1.55E+02) + & \textbf{8.27E+02 (2.33E+02)} & 1.17E+03 (3.07E+02) + & 8.48E+02 (1.84E+02) = & \textbf{7.06E+02 (1.52E+02)} & 1.02E+03 (2.66E+02) + & 1.05E+03 (1.87E+02) + \\
    F30   & \textbf{7.72E+05 (8.96E+04)} & 8.87E+05 (1.92E+05) + & 8.45E+06 (1.87E+06) + & \textbf{7.69E+05 (1.32E+05)} & 7.79E+05 (1.64E+05) = & 7.76E+05 (1.56E+05) = & 1.17E+06 (4.60E+05) & \textbf{1.13E+06 (3.83E+05) =} & 1.49E+06 (6.10E+05) + \\
    \midrule
    +/=/- &       & 20/6/4 & 24/5/1 &       & 19/7/4 & 8/22/0 &       & 22/2/6 & 19/11/0 \\
    \midrule
    \midrule
          & \multicolumn{3}{c}{DE/current-to-rand/1} & \multicolumn{3}{c}{DE/rand/2/bin} & \multicolumn{3}{c}{DE/current-to-best/2/bin} \\
          & ACM   & CM    & Original & ACM   & CM    & Original & ACM   & CM    & Original \\
          & MEAN (STD DEV) & MEAN (STD DEV) & MEAN (STD DEV) & MEAN (STD DEV) & MEAN (STD DEV) & MEAN (STD DEV) & MEAN (STD DEV) & MEAN (STD DEV) & MEAN (STD DEV) \\
    \midrule
    F1    & 7.02E+03 (9.59E+03) & \textbf{2.87E+03 (3.54E+03) -} & 1.18E+04 (1.98E+04) = & 5.79E+05 (1.36E+05) & \textbf{5.25E+04 (1.80E+04) -} & 1.69E+08 (9.73E+07) + & 7.89E+04 (8.59E+04) & \textbf{6.86E+03 (7.75E+03) -} & 9.00E+04 (1.72E+05) = \\
    F2    & \textbf{1.87E+15 (1.27E+16)} & 2.14E+23 (1.46E+24) - & 5.10E+44 (1.78E+45) + & 3.12E+19 (1.73E+20) & \textbf{9.42E+13 (4.30E+14) -} & 1.25E+61 (4.24E+61) + & 3.53E+15 (1.18E+16) & \textbf{1.96E+15 (9.21E+15) =} & 9.33E+51 (2.05E+52) + \\
    F3    & 1.90E+04 (4.59E+03) & \textbf{1.40E+04 (4.20E+03) -} & 4.51E+04 (6.12E+03) + & 6.56E+04 (1.34E+04) & \textbf{4.83E+04 (9.33E+03) -} & 1.34E+05 (1.39E+04) + & 3.58E+04 (8.38E+03) & \textbf{2.87E+04 (5.35E+03) -} & 8.05E+04 (1.01E+04) + \\
    F4    & 8.96E+01 (4.58E+01) & 9.28E+01 (5.72E+01) = & \textbf{8.21E+01 (4.63E+01) =} & 1.31E+02 (4.33E+01) & \textbf{1.21E+02 (5.25E+01) =} & 3.31E+02 (1.76E+01) + & \textbf{5.27E+01 (4.49E+01)} & 1.02E+02 (6.03E+01) + & 8.95E+01 (5.88E+01) + \\
    F5    & \textbf{6.51E+01 (1.67E+01)} & 1.02E+02 (2.40E+01) + & 3.35E+02 (1.26E+01) + & \textbf{1.04E+02 (2.77E+01)} & 1.40E+02 (3.53E+01) + & 4.13E+02 (1.26E+01) + & \textbf{8.31E+01 (2.24E+01)} & 1.31E+02 (3.32E+01) + & 3.71E+02 (1.45E+01) + \\
    F6    & \textbf{1.14E-13 (7.65E-29)} & 3.43E-03 (1.59E-02) = & \textbf{1.14E-13 (7.65E-29) =} & 2.32E-01 (1.61E-02) & \textbf{1.10E-01 (1.41E-02) -} & 3.87E+00 (3.74E-01) + & 4.43E-04 (1.13E-04) & 1.43E-02 (5.02E-02) - & \textbf{2.65E-04 (7.82E-05) -} \\
    F7    & \textbf{9.84E+01 (1.53E+01)} & 1.42E+02 (2.45E+01) + & 3.82E+02 (1.34E+01) + & \textbf{1.49E+02 (2.74E+01)} & 2.08E+02 (3.88E+01) + & 5.33E+02 (1.70E+01) + & \textbf{1.31E+02 (2.08E+01)} & 1.87E+02 (3.45E+01) + & 4.42E+02 (1.42E+01) + \\
    F8    & \textbf{6.48E+01 (1.89E+01)} & 1.02E+02 (2.66E+01) + & 3.39E+02 (9.39E+00) + & \textbf{9.78E+01 (2.39E+01)} & 1.39E+02 (3.93E+01) + & 4.10E+02 (1.97E+01) + & \textbf{8.50E+01 (2.34E+01)} & 1.25E+02 (2.61E+01) + & 3.75E+02 (1.32E+01) + \\
    F9    & 1.96E-02 (8.95E-02) & 7.67E-02 (3.43E-01) = & \textbf{1.07E-02 (6.45E-02) =} & \textbf{9.37E-01 (1.10E-01)} & 4.38E+01 (8.52E+01) + & 1.29E+03 (1.64E+02) + & 1.25E-13 (3.39E-14) & 1.65E+01 (4.16E+01) + & \textbf{1.23E-13 (3.07E-14) =} \\
    F10   & \textbf{4.52E+03 (9.21E+02)} & 5.78E+03 (9.10E+02) + & 1.29E+04 (3.45E+02) + & \textbf{4.25E+03 (6.77E+02)} & 5.59E+03 (8.06E+02) + & 1.29E+04 (3.57E+02) + & \textbf{4.28E+03 (7.43E+02)} & 5.51E+03 (7.51E+02) + & 1.28E+04 (3.77E+02) + \\
    F11   & \textbf{3.91E+01 (9.00E+00)} & 7.23E+01 (2.12E+01) + & 1.45E+02 (9.61E+00) + & \textbf{6.17E+01 (1.15E+01)} & 1.34E+02 (5.06E+01) + & 2.99E+02 (1.47E+01) + & \textbf{5.07E+01 (1.05E+01)} & 1.23E+02 (4.20E+01) + & 1.85E+02 (9.08E+00) + \\
    F12   & \textbf{2.83E+06 (9.86E+05)} & 3.57E+06 (1.41E+06) + & 2.21E+07 (4.51E+06) + & 3.55E+06 (1.84E+06) & \textbf{2.45E+06 (1.53E+06) -} & 3.43E+08 (6.51E+07) + & 2.04E+06 (1.03E+06) & \textbf{1.82E+06 (1.04E+06) =} & 5.27E+07 (1.40E+07) + \\
    F13   & \textbf{7.74E+04 (2.03E+04)} & 9.04E+04 (2.71E+04) + & 1.35E+05 (2.56E+04) + & 3.80E+04 (1.96E+04) & \textbf{9.20E+03 (7.31E+03) -} & 7.89E+05 (3.23E+05) + & 4.15E+04 (1.48E+04) & \textbf{1.25E+04 (1.00E+04) -} & 7.75E+04 (3.27E+04) + \\
    F14   & \textbf{1.27E+03 (6.83E+02)} & 6.88E+03 (6.43E+03) + & 9.94E+04 (2.78E+04) + & \textbf{2.57E+04 (2.45E+04)} & 4.00E+04 (3.33E+04) + & 3.30E+05 (1.18E+05) + & \textbf{5.19E+03 (5.99E+03)} & 1.19E+04 (1.86E+04) + & 6.57E+04 (1.75E+04) + \\
    F15   & \textbf{1.26E+04 (3.28E+03)} & 2.01E+04 (5.02E+03) + & 2.86E+04 (5.38E+03) + & 1.19E+04 (4.39E+03) & \textbf{4.95E+03 (5.99E+03) -} & 5.98E+04 (2.71E+04) + & 6.37E+03 (2.63E+03) & \textbf{4.02E+03 (2.59E+03) -} & 1.26E+04 (4.70E+03) + \\
    F16   & \textbf{1.12E+03 (4.11E+02)} & 1.21E+03 (4.33E+02) = & 2.48E+03 (2.03E+02) + & \textbf{1.38E+03 (4.02E+02)} & 1.60E+03 (3.84E+02) + & 3.05E+03 (1.66E+02) + & \textbf{1.38E+03 (3.43E+02)} & 1.51E+03 (3.45E+02) + & 2.76E+03 (2.08E+02) + \\
    F17   & \textbf{7.38E+02 (2.57E+02)} & 1.11E+03 (2.90E+02) + & 1.52E+03 (1.40E+02) + & \textbf{8.03E+02 (2.68E+02)} & 1.06E+03 (2.93E+02) + & 1.72E+03 (1.30E+02) + & \textbf{7.58E+02 (2.30E+02)} & 1.06E+03 (2.65E+02) + & 1.56E+03 (1.46E+02) + \\
    F18   & \textbf{7.85E+04 (3.60E+04)} & 2.04E+05 (1.21E+05) + & 1.87E+06 (5.04E+05) + & \textbf{4.31E+05 (2.45E+05)} & 5.87E+05 (4.25E+05) + & 7.38E+06 (2.01E+06) + & \textbf{2.22E+05 (1.41E+05)} & 4.74E+05 (3.14E+05) + & 2.51E+06 (7.91E+05) + \\
    F19   & \textbf{5.32E+03 (1.62E+03)} & 9.54E+03 (3.19E+03) + & 1.33E+04 (3.83E+03) + & \textbf{9.52E+03 (5.14E+03)} & 1.41E+04 (1.15E+04) + & 4.30E+04 (1.78E+04) + & 2.68E+03 (1.19E+03) & \textbf{1.48E+03 (9.17E+02) -} & 5.44E+03 (1.83E+03) + \\
    F20   & \textbf{5.94E+02 (2.67E+02)} & 8.33E+02 (2.78E+02) + & 1.43E+03 (1.44E+02) + & \textbf{6.55E+02 (2.60E+02)} & 9.63E+02 (2.94E+02) + & 1.47E+03 (1.60E+02) + & \textbf{6.47E+02 (2.09E+02)} & 8.55E+02 (2.88E+02) + & 1.33E+03 (1.49E+02) + \\
    F21   & \textbf{2.64E+02 (1.44E+01)} & 2.99E+02 (2.35E+01) + & 5.34E+02 (1.42E+01) + & \textbf{2.97E+02 (1.79E+01)} & 3.38E+02 (3.48E+01) + & 6.07E+02 (1.53E+01) + & \textbf{2.86E+02 (1.90E+01)} & 3.30E+02 (2.88E+01) + & 5.75E+02 (1.60E+01) + \\
    F22   & \textbf{3.36E+03 (2.69E+03)} & 5.12E+03 (2.60E+03) + & 6.19E+03 (6.53E+03) = & \textbf{4.48E+03 (1.68E+03)} & 6.37E+03 (1.30E+03) + & 1.25E+04 (2.57E+03) + & \textbf{4.64E+03 (1.48E+03)} & 6.53E+03 (7.49E+02) + & 1.16E+04 (4.02E+03) + \\
    F23   & \textbf{4.83E+02 (1.88E+01)} & 5.32E+02 (2.27E+01) + & 7.56E+02 (1.56E+01) + & \textbf{5.27E+02 (2.65E+01)} & 5.61E+02 (2.96E+01) + & 8.29E+02 (1.50E+01) + & \textbf{5.02E+02 (2.41E+01)} & 5.51E+02 (3.35E+01) + & 7.94E+02 (1.45E+01) + \\
    F24   & \textbf{5.51E+02 (1.61E+01)} & 5.84E+02 (2.64E+01) + & 8.20E+02 (1.30E+01) + & \textbf{5.98E+02 (2.16E+01)} & 6.16E+02 (2.57E+01) + & 8.73E+02 (1.64E+01) + & \textbf{5.89E+02 (1.95E+01)} & 6.27E+02 (2.98E+01) + & 8.60E+02 (1.31E+01) + \\
    F25   & \textbf{5.13E+02 (3.12E+01)} & 5.22E+02 (3.40E+01) = & 5.19E+02 (3.10E+01) = & \textbf{4.81E+02 (1.55E+00)} & 5.22E+02 (3.87E+01) + & 5.91E+02 (1.01E+01) + & 4.86E+02 (2.03E+01) & 5.08E+02 (3.48E+01) = & \textbf{4.84E+02 (1.64E+01) =} \\
    F26   & \textbf{1.68E+03 (1.99E+02)} & 2.30E+03 (3.29E+02) + & 4.25E+03 (1.23E+02) + & \textbf{2.07E+03 (2.41E+02)} & 2.59E+03 (3.63E+02) + & 5.15E+03 (1.15E+02) + & \textbf{1.90E+03 (2.23E+02)} & 2.51E+03 (3.14E+02) + & 4.67E+03 (1.78E+02) + \\
    F27   & 5.51E+02 (6.13E+01) & 5.96E+02 (6.57E+01) + & \textbf{5.48E+02 (6.64E+01) =} & \textbf{5.11E+02 (7.79E+00)} & 6.22E+02 (5.17E+01) + & 5.91E+02 (1.58E+01) + & 5.14E+02 (1.18E+01) & 5.85E+02 (4.19E+01) + & \textbf{5.10E+02 (1.52E+01) =} \\
    F28   & 4.75E+02 (2.29E+01) & 4.90E+02 (2.59E+01) = & \textbf{4.73E+02 (2.12E+01) =} & \textbf{4.59E+02 (0.00E+00)} & 4.96E+02 (2.02E+01) + & 5.11E+02 (9.91E+00) + & 4.61E+02 (9.61E+00) & 4.84E+02 (2.61E+01) + & \textbf{4.60E+02 (6.86E+00) =} \\
    F29   & \textbf{4.67E+02 (1.60E+02)} & 8.60E+02 (2.68E+02) + & 1.38E+03 (2.25E+02) + & \textbf{6.83E+02 (2.42E+02)} & 1.06E+03 (2.79E+02) + & 2.17E+03 (1.67E+02) + & \textbf{5.90E+02 (2.05E+02)} & 9.17E+02 (2.81E+02) + & 1.71E+03 (1.91E+02) + \\
    F30   & \textbf{1.06E+06 (1.32E+05)} & 1.61E+06 (3.25E+05) + & 1.47E+07 (2.87E+06) + & \textbf{9.15E+05 (1.81E+05)} & 1.02E+06 (2.19E+05) + & 3.03E+07 (5.73E+06) + & \textbf{8.44E+05 (1.47E+05)} & 8.85E+05 (1.89E+05) = & 5.43E+06 (1.37E+06) + \\
    \midrule
    +/=/- &       & 21/6/3 & 22/8/0 &       & 22/1/7 & 30/0/0 &       & 20/4/6 & 24/5/1 \\
    \bottomrule
    \end{tabular}%
    \end{adjustwidth}
  \label{tab:conventional_d50}%
  The symbols "+/=/-" show the statistical results of the Wilcoxon signed-rank test with $\alpha = 0.05$ significance level. "+" represents that ACM variant is significantly superior than the corresponding variant. "=" represents that the performance difference between ACM variant and the corresponding variant is not statistically significant. And, "-" represents that ACM variant is significantly inferior than the corresponding variant.
\end{table*}%

\begin{figure*}[htbp]
 \centering
 \subfigure[$F_{10}$]{
  \includegraphics[scale=0.25]{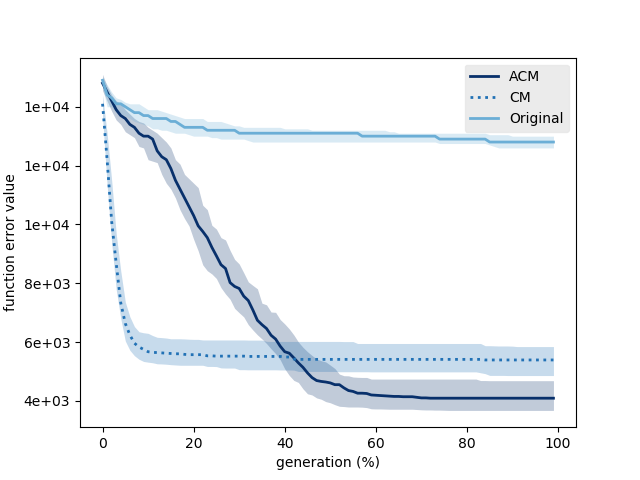}
   }
 \subfigure[$F_{16}$]{
  \includegraphics[scale=0.25]{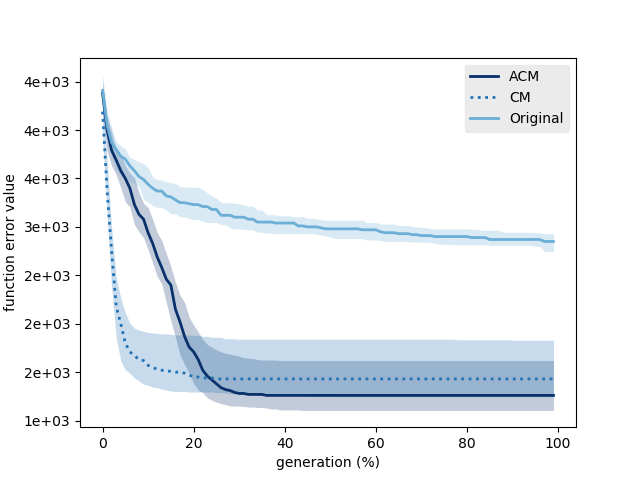}
   }
 \subfigure[$F_{20}$]{
  \includegraphics[scale=0.25]{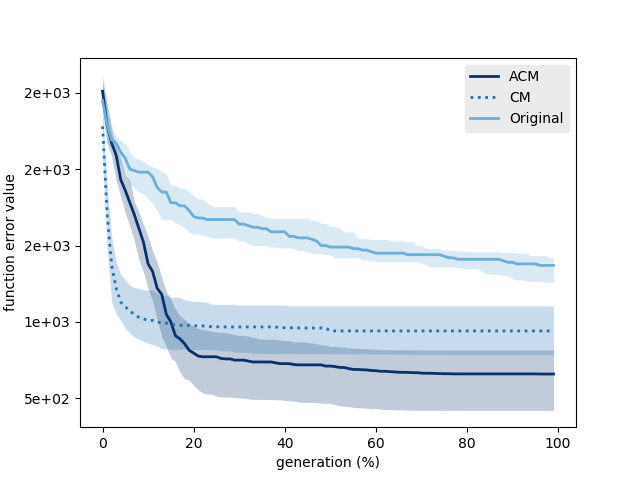}
   }
 \subfigure[$F_{26}$]{
  \includegraphics[scale=0.25]{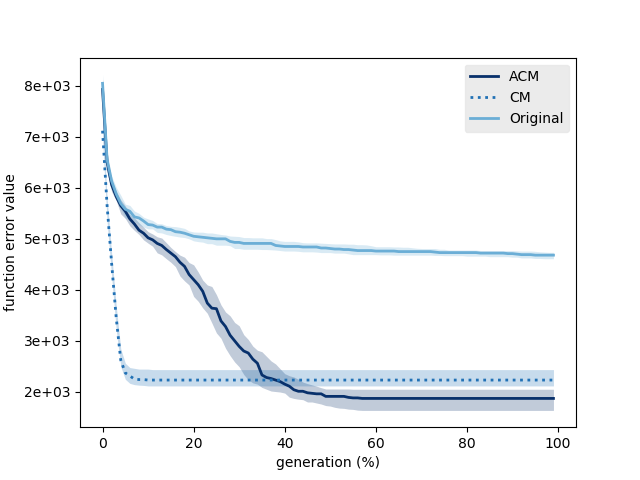}
   }
 \caption[]{Median and interquartile ranges (25th and 75th) of function error values with DE/rand/1/bin variants ($D = 50$)}
 \label{fig:convergenceGraphs_rand1bin_d50}
\end{figure*}

\begin{table*}[htbp]
  \tiny
  \centering
  \caption{Averages and standard deviations of function error values with advanced DE variants ($D = 30$)}
    \begin{adjustwidth}{-0.75cm}{}
    \begin{tabular}{c|ccc||ccc||ccc}
    \toprule
          & \multicolumn{3}{c}{SaDE} & \multicolumn{3}{c}{EPSDE} & \multicolumn{3}{c}{CoDE} \\
          & ACM   & CM    & Original & ACM   & CM    & Original & ACM   & CM    & Original \\
          & MEAN (STD DEV) & MEAN (STD DEV) & MEAN (STD DEV) & MEAN (STD DEV) & MEAN (STD DEV) & MEAN (STD DEV) & MEAN (STD DEV) & MEAN (STD DEV) & MEAN (STD DEV) \\
    \midrule
    F1    & 9.67E+02 (1.20E+03) & 7.51E+01 (2.05E+02) - & \textbf{8.56E+00 (2.52E+01) -} & \textbf{5.01E-15 (7.42E-15)} & 5.29E-15 (7.49E-15) = & 6.13E-15 (7.10E-15) = & 5.48E+00 (2.19E+00) & \textbf{1.45E-01 (9.03E-02) -} & 5.20E+00 (2.07E+00) = \\
    F2    & 2.57E+05 (1.54E+06) & \textbf{1.26E+05 (6.75E+05) =} & 1.71E+12 (6.79E+12) = & 1.52E-05 (4.59E-05) & \textbf{5.50E-07 (7.49E-07) -} & 7.12E+12 (3.69E+13) = & \textbf{5.46E+08 (3.07E+09)} & 1.13E+18 (8.05E+18) = & 8.21E+21 (1.82E+22) + \\
    F3    & 5.53E+02 (1.77E+03) & \textbf{4.55E+02 (9.88E+02) =} & 7.85E+03 (1.71E+04) = & 1.75E-03 (1.20E-02) & \textbf{2.67E-14 (3.08E-14) -} & 4.44E+03 (8.42E+03) + & \textbf{1.48E+03 (5.95E+02)} & 3.05E+03 (1.77E+03) + & 3.28E+04 (5.57E+03) + \\
    F4    & 3.38E+01 (3.27E+01) & \textbf{1.83E+01 (2.83E+01) =} & 9.40E+01 (1.60E+01) + & 3.66E+01 (2.95E+01) & \textbf{1.56E+01 (2.57E+01) -} & 4.02E+01 (2.86E+01) = & 5.92E+01 (7.24E+00) & \textbf{4.43E+01 (2.22E+01) -} & 8.31E+01 (5.34E+00) + \\
    F5    & \textbf{2.65E+01 (5.98E+00)} & 4.56E+01 (1.43E+01) + & 3.81E+01 (7.38E+00) + & \textbf{3.52E+01 (8.14E+00)} & 5.77E+01 (1.56E+01) + & 6.10E+01 (8.44E+00) + & \textbf{3.95E+01 (1.11E+01)} & 6.29E+01 (1.72E+01) + & 1.21E+02 (9.48E+00) + \\
    F6    & \textbf{1.14E-13 (7.65E-29)} & 4.64E-11 (5.81E-11) + & 1.21E-13 (2.69E-14) = & \textbf{1.14E-13 (7.65E-29)} & 8.94E-04 (6.39E-03) = & \textbf{1.14E-13 (7.65E-29) =} & 1.10E-04 (4.50E-05) & \textbf{5.52E-06 (3.33E-06) -} & 9.64E-05 (3.14E-05) = \\
    F7    & \textbf{5.73E+01 (7.28E+00)} & 7.40E+01 (1.37E+01) + & 7.06E+01 (6.59E+00) + & \textbf{5.47E+01 (5.68E+00)} & 7.90E+01 (1.48E+01) + & 9.49E+01 (8.42E+00) + & \textbf{7.43E+01 (9.07E+00)} & 9.34E+01 (2.04E+01) + & 1.76E+02 (1.34E+01) + \\
    F8    & \textbf{2.54E+01 (6.28E+00)} & 4.99E+01 (1.33E+01) + & 3.69E+01 (6.87E+00) + & \textbf{3.15E+01 (8.34E+00)} & 5.85E+01 (1.54E+01) + & 6.56E+01 (7.66E+00) + & \textbf{4.46E+01 (1.23E+01)} & 6.44E+01 (1.99E+01) + & 1.28E+02 (1.08E+01) + \\
    F9    & \textbf{3.04E-02 (9.73E-02)} & 1.19E-01 (4.05E-01) = & 4.64E-01 (1.04E+00) = & \textbf{8.90E-02 (2.74E-01)} & 3.94E-01 (1.21E+00) = & 2.55E-01 (9.81E-01) = & \textbf{2.97E-03 (6.64E-03)} & 4.16E-01 (1.95E+00) - & 2.18E+02 (8.40E+01) + \\
    F10   & \textbf{1.62E+03 (4.45E+02)} & 2.74E+03 (6.00E+02) + & 2.38E+03 (3.67E+02) + & \textbf{1.75E+03 (5.49E+02)} & 2.85E+03 (5.43E+02) + & 3.59E+03 (2.88E+02) + & \textbf{1.79E+03 (4.26E+02)} & 2.94E+03 (5.87E+02) + & 4.65E+03 (3.08E+02) + \\
    F11   & \textbf{1.57E+01 (1.30E+01)} & 4.45E+01 (2.60E+01) + & 3.08E+01 (2.30E+01) + & \textbf{3.70E+01 (3.34E+01)} & 5.19E+01 (3.81E+01) + & 4.68E+01 (4.16E+01) = & \textbf{1.99E+01 (1.38E+01)} & 2.71E+01 (1.91E+01) + & 1.10E+02 (1.82E+01) + \\
    F12   & 1.07E+05 (1.77E+05) & 1.00E+05 (2.01E+05) = & \textbf{2.17E+04 (5.87E+04) -} & \textbf{6.31E+03 (6.52E+03)} & 6.90E+03 (7.28E+03) = & 7.43E+03 (8.03E+03) = & 1.26E+05 (7.82E+04) & \textbf{5.89E+04 (3.77E+04) -} & 3.50E+06 (9.63E+05) + \\
    F13   & 3.50E+03 (5.02E+03) & \textbf{1.60E+03 (4.16E+03) -} & 2.54E+03 (7.73E+03) - & 3.46E+02 (4.51E+02) & 2.72E+02 (3.62E+02) = & \textbf{2.50E+02 (2.66E+02) =} & 7.29E+02 (4.37E+02) & \textbf{3.73E+02 (9.85E+01) -} & 8.40E+02 (3.85E+02) = \\
    F14   & 2.16E+03 (3.54E+03) & \textbf{1.62E+03 (4.62E+03) -} & 3.41E+03 (7.71E+03) = & 7.76E+01 (5.11E+01) & 6.58E+01 (4.82E+01) = & \textbf{6.48E+01 (4.24E+01) =} & \textbf{6.50E+01 (8.46E+00)} & 6.61E+01 (1.22E+01) = & 7.29E+01 (7.99E+00) + \\
    F15   & \textbf{4.17E+02 (1.00E+03)} & 9.99E+02 (2.25E+03) + & 8.64E+03 (1.24E+04) + & 1.31E+02 (1.03E+02) & \textbf{1.03E+02 (9.67E+01) =} & 1.68E+02 (1.44E+02) = & \textbf{7.92E+01 (1.91E+01)} & 8.30E+01 (1.53E+01) = & 9.47E+01 (2.24E+01) + \\
    F16   & \textbf{3.81E+02 (1.68E+02)} & 6.26E+02 (2.71E+02) + & 4.47E+02 (1.50E+02) = & \textbf{3.64E+02 (1.81E+02)} & 7.59E+02 (2.67E+02) + & 3.74E+02 (1.28E+02) = & \textbf{3.91E+02 (1.65E+02)} & 6.56E+02 (2.62E+02) + & 5.45E+02 (1.46E+02) + \\
    F17   & \textbf{2.72E+01 (2.03E+01)} & 2.12E+02 (1.38E+02) + & 6.94E+01 (2.24E+01) + & \textbf{4.83E+01 (5.27E+01)} & 2.40E+02 (1.68E+02) + & 9.79E+01 (4.53E+01) + & \textbf{7.10E+01 (6.40E+01)} & 1.86E+02 (1.22E+02) + & 1.01E+02 (4.74E+01) + \\
    F18   & \textbf{1.08E+04 (2.52E+04)} & 3.77E+04 (5.26E+04) = & 2.87E+04 (7.44E+04) = & \textbf{4.71E+02 (5.16E+02)} & 8.61E+02 (1.44E+03) = & 7.67E+02 (1.22E+03) = & 5.38E+03 (6.15E+03) & \textbf{3.05E+03 (4.52E+03) -} & 1.29E+04 (1.18E+04) + \\
    F19   & \textbf{3.37E+02 (1.16E+03)} & 3.79E+03 (6.18E+03) + & 2.16E+03 (4.70E+03) = & 6.98E+01 (5.70E+01) & \textbf{5.40E+01 (4.02E+01) =} & 7.70E+01 (5.46E+01) = & 4.06E+01 (7.74E+00) & \textbf{3.85E+01 (6.32E+00) =} & 4.08E+01 (8.04E+00) = \\
    F20   & \textbf{5.32E+01 (4.97E+01)} & 2.04E+02 (1.25E+02) + & 1.28E+02 (5.52E+01) + & \textbf{9.40E+01 (7.38E+01)} & 3.30E+02 (1.71E+02) + & 1.11E+02 (6.14E+01) = & 1.01E+02 (8.45E+01) & 2.03E+02 (1.09E+02) + & \textbf{8.92E+01 (6.20E+01) =} \\
    F21   & \textbf{2.31E+02 (4.58E+00)} & 2.44E+02 (1.30E+01) + & 2.36E+02 (5.30E+00) + & \textbf{2.35E+02 (7.80E+00)} & 2.59E+02 (1.59E+01) + & 2.63E+02 (8.18E+00) + & \textbf{2.45E+02 (9.78E+00)} & 2.66E+02 (1.42E+01) + & 3.24E+02 (1.08E+01) + \\
    F22   & \textbf{1.00E+02 (0.00E+00)} & \textbf{1.00E+02 (0.00E+00) =} & \textbf{1.00E+02 (0.00E+00) =} & \textbf{1.00E+02 (2.80E-01)} & 2.09E+02 (5.64E+02) = & \textbf{1.00E+02 (0.00E+00) =} & \textbf{1.00E+02 (0.00E+00)} & 1.56E+02 (3.99E+02) = & \textbf{1.00E+02 (0.00E+00) =} \\
    F23   & \textbf{3.72E+02 (5.50E+00)} & 3.97E+02 (1.74E+01) + & 3.80E+02 (6.61E+00) + & \textbf{3.79E+02 (1.08E+01)} & 4.08E+02 (1.56E+01) + & 4.06E+02 (8.88E+00) + & \textbf{3.91E+02 (1.03E+01)} & 4.11E+02 (2.25E+01) + & 4.62E+02 (9.64E+00) + \\
    F24   & \textbf{4.51E+02 (5.87E+00)} & 4.57E+02 (1.68E+01) = & 4.52E+02 (9.21E+00) = & \textbf{4.56E+02 (1.44E+01)} & 4.70E+02 (1.84E+01) + & 4.76E+02 (1.03E+01) + & \textbf{4.71E+02 (1.17E+01)} & 4.79E+02 (1.56E+01) + & 5.56E+02 (1.03E+01) + \\
    F25   & \textbf{3.87E+02 (2.38E-01)} & \textbf{3.87E+02 (5.80E-01) =} & \textbf{3.87E+02 (4.20E-01) =} & \textbf{3.87E+02 (6.88E-01)} & \textbf{3.87E+02 (1.43E+00) =} & \textbf{3.87E+02 (8.48E-01) =} & \textbf{3.87E+02 (0.00E+00)} & \textbf{3.87E+02 (7.84E-01) =} & \textbf{3.87E+02 (0.00E+00) =} \\
    F26   & \textbf{1.24E+03 (1.04E+02)} & 1.47E+03 (3.04E+02) + & 1.27E+03 (1.49E+02) = & \textbf{1.24E+03 (2.25E+02)} & 1.63E+03 (2.78E+02) + & 1.44E+03 (9.93E+01) + & \textbf{1.38E+03 (3.00E+02)} & 1.62E+03 (3.71E+02) + & 2.23E+03 (1.03E+02) + \\
    F27   & 5.03E+02 (6.00E+00) & 5.10E+02 (8.92E+00) + & \textbf{5.01E+02 (7.91E+00) =} & 5.06E+02 (1.09E+01) & 5.15E+02 (9.77E+00) + & \textbf{5.03E+02 (9.06E+00) =} & \textbf{4.93E+02 (8.78E+00)} & 5.07E+02 (7.04E+00) + & 5.11E+02 (4.57E+00) + \\
    F28   & 3.59E+02 (5.27E+01) & \textbf{3.14E+02 (3.61E+01) -} & 3.87E+02 (4.87E+01) + & \textbf{3.39E+02 (5.67E+01)} & 3.65E+02 (6.13E+01) = & 3.53E+02 (5.74E+01) = & \textbf{3.15E+02 (3.42E+01)} & 3.66E+02 (6.18E+01) + & 4.38E+02 (1.77E+01) + \\
    F29   & \textbf{4.36E+02 (3.71E+01)} & 5.58E+02 (1.40E+02) + & 4.75E+02 (4.34E+01) + & \textbf{4.42E+02 (4.73E+01)} & 6.56E+02 (1.60E+02) + & 5.11E+02 (4.30E+01) + & \textbf{4.50E+02 (5.95E+01)} & 5.51E+02 (1.32E+02) + & 6.63E+02 (7.03E+01) + \\
    F30   & \textbf{2.48E+03 (1.24E+03)} & 4.67E+03 (2.49E+03) + & 1.05E+04 (7.30E+03) + & 2.33E+03 (6.92E+02) & 2.34E+03 (3.89E+02) = & \textbf{2.32E+03 (3.69E+02) =} & 6.73E+03 (1.91E+03) & \textbf{5.05E+03 (2.38E+03) -} & 9.61E+03 (2.76E+03) + \\
    \midrule
    +/=/- &       & 17/9/4 & 14/13/3 &       & 14/13/3 & 11/19/0 &       & 16/6/8 & 23/7/0 \\
    \midrule
    \midrule
          & \multicolumn{3}{c}{SHADE} & \multicolumn{3}{c}{MPEDE} & \multicolumn{3}{c}{EDEV} \\
          & ACM   & CM    & Original & ACM   & CM    & Original & ACM   & CM    & Original \\
          & MEAN (STD DEV) & MEAN (STD DEV) & MEAN (STD DEV) & MEAN (STD DEV) & MEAN (STD DEV) & MEAN (STD DEV) & MEAN (STD DEV) & MEAN (STD DEV) & MEAN (STD DEV) \\
    \midrule
    F1    & 1.34E-14 (3.37E-15) & \textbf{1.23E-14 (4.94E-15) =} & 1.25E-14 (5.42E-15) = & 5.00E+00 (2.55E+01) & 9.02E+00 (3.96E+01) = & \textbf{1.56E+00 (9.94E+00) =} & 1.11E-15 (4.79E-15) & 2.78E-16 (1.99E-15) = & \textbf{0.00E+00 (0.00E+00) =} \\
    F2    & \textbf{4.25E-13 (4.67E-13)} & 6.77E-13 (1.37E-12) = & 4.75E-13 (9.39E-13) = & 1.20E+06 (8.57E+06) & \textbf{2.44E+04 (1.74E+05) =} & 1.24E+11 (8.58E+11) = & 5.43E+10 (3.88E+11) & \textbf{3.72E-05 (1.92E-04) =} & 3.90E+11 (2.79E+12) = \\
    F3    & \textbf{7.14E-14 (2.76E-14)} & 7.59E-14 (2.95E-14) = & \textbf{7.14E-14 (3.19E-14) =} & \textbf{4.68E+03 (6.61E+03)} & 5.15E+03 (7.92E+03) = & 8.77E+03 (1.49E+04) = & 6.07E+02 (2.66E+03) & \textbf{3.66E+02 (2.38E+03) =} & 4.60E+03 (1.32E+04) = \\
    F4    & 3.38E+01 (2.99E+01) & \textbf{3.33E+01 (3.03E+01) =} & 3.36E+01 (2.94E+01) = & 5.30E+01 (1.77E+01) & \textbf{4.73E+01 (2.52E+01) =} & 5.55E+01 (1.40E+01) = & 5.12E+01 (2.07E+01) & \textbf{4.28E+01 (2.73E+01) =} & 5.15E+01 (2.06E+01) = \\
    F5    & \textbf{1.74E+01 (3.05E+00)} & 4.28E+01 (1.43E+01) + & 1.92E+01 (2.95E+00) + & \textbf{2.45E+01 (5.88E+00)} & 3.55E+01 (1.11E+01) + & 5.04E+01 (5.18E+00) + & \textbf{2.51E+01 (5.69E+00)} & 4.58E+01 (1.74E+01) + & 3.26E+01 (6.68E+00) + \\
    F6    & 1.07E-08 (3.72E-08) & \textbf{6.71E-09 (2.74E-08) =} & 8.73E-09 (3.27E-08) = & \textbf{2.14E-13 (1.69E-13)} & 7.00E-13 (1.32E-12) + & 2.65E-13 (4.15E-13) = & 1.65E-11 (6.38E-11) & \textbf{7.18E-12 (1.25E-11) =} & 1.05E-11 (5.76E-11) = \\
    F7    & \textbf{4.87E+01 (2.67E+00)} & 6.67E+01 (1.20E+01) + & 4.98E+01 (3.56E+00) = & \textbf{6.64E+01 (8.06E+00)} & 6.66E+01 (8.51E+00) = & 8.19E+01 (5.41E+00) + & \textbf{5.65E+01 (5.37E+00)} & 6.61E+01 (1.18E+01) + & 6.19E+01 (4.61E+00) + \\
    F8    & \textbf{1.82E+01 (2.98E+00)} & 4.48E+01 (1.21E+01) + & 2.03E+01 (4.18E+00) + & \textbf{2.41E+01 (5.08E+00)} & 3.73E+01 (1.01E+01) + & 5.03E+01 (5.07E+00) + & \textbf{2.61E+01 (5.78E+00)} & 4.44E+01 (1.19E+01) + & 3.14E+01 (4.59E+00) + \\
    F9    & 1.96E-02 (8.95E-02) & 2.85E-02 (1.08E-01) = & \textbf{5.81E-14 (5.76E-14) =} & \textbf{0.00E+00 (0.00E+00)} & \textbf{0.00E+00 (0.00E+00) =} & \textbf{0.00E+00 (0.00E+00) =} & \textbf{3.51E-03 (1.75E-02)} & 8.77E-03 (2.69E-02) = & 1.96E-02 (8.95E-02) = \\
    F10   & \textbf{1.55E+03 (3.10E+02)} & 2.62E+03 (5.78E+02) + & 1.82E+03 (2.21E+02) + & \textbf{1.69E+03 (4.11E+02)} & 2.39E+03 (5.76E+02) + & 3.48E+03 (2.11E+02) + & \textbf{1.68E+03 (3.62E+02)} & 2.36E+03 (5.53E+02) + & 2.43E+03 (3.03E+02) + \\
    F11   & \textbf{2.59E+01 (2.57E+01)} & 4.17E+01 (2.97E+01) + & 2.85E+01 (2.58E+01) = & \textbf{1.82E+01 (1.41E+01)} & 3.51E+01 (2.71E+01) + & 3.62E+01 (2.02E+01) + & \textbf{1.83E+01 (2.03E+01)} & 2.88E+01 (2.37E+01) + & 2.50E+01 (2.33E+01) = \\
    F12   & \textbf{1.20E+03 (4.56E+02)} & 1.31E+03 (6.02E+02) = & 1.36E+03 (8.29E+02) = & \textbf{1.06E+03 (3.65E+02)} & 1.11E+03 (3.92E+02) = & 1.25E+03 (8.48E+02) = & \textbf{1.11E+03 (4.24E+02)} & 1.83E+03 (2.11E+03) + & 2.48E+03 (5.93E+03) = \\
    F13   & 5.98E+01 (6.84E+01) & 4.42E+01 (2.97E+01) = & \textbf{4.26E+01 (3.62E+01) -} & 9.70E+03 (1.56E+04) & \textbf{6.49E+03 (1.26E+04) =} & 1.44E+04 (2.75E+04) = & 6.06E+01 (5.34E+01) & 6.37E+01 (6.61E+01) = & \textbf{5.07E+01 (4.02E+01) =} \\
    F14   & 3.27E+01 (8.37E+00) & 3.48E+01 (1.26E+01) = & \textbf{3.18E+01 (7.35E+00) =} & \textbf{3.56E+02 (2.64E+02)} & 5.45E+02 (6.12E+02) = & 8.94E+03 (7.21E+03) + & 3.62E+01 (1.77E+01) & 3.78E+01 (9.69E+00) = & \textbf{3.26E+01 (1.44E+01) =} \\
    F15   & 2.17E+01 (1.69E+01) & 2.64E+01 (2.50E+01) = & \textbf{2.07E+01 (1.47E+01) =} & \textbf{3.32E+03 (2.40E+03)} & 3.81E+03 (4.39E+03) = & 1.95E+04 (1.10E+04) + & \textbf{2.60E+01 (2.75E+01)} & 2.89E+01 (2.70E+01) = & 2.69E+01 (1.77E+01) = \\
    F16   & 2.91E+02 (1.48E+02) & 5.71E+02 (2.12E+02) + & \textbf{2.54E+02 (1.36E+02) =} & \textbf{3.36E+02 (1.30E+02)} & 4.74E+02 (2.17E+02) + & 5.75E+02 (1.26E+02) + & \textbf{3.17E+02 (1.31E+02)} & 5.89E+02 (2.34E+02) + & 4.59E+02 (1.62E+02) + \\
    F17   & \textbf{3.82E+01 (2.64E+01)} & 1.69E+02 (1.12E+02) + & 5.73E+01 (3.16E+01) + & \textbf{4.45E+01 (2.16E+01)} & 1.19E+02 (1.13E+02) + & 1.45E+02 (2.82E+01) + & \textbf{4.14E+01 (2.66E+01)} & 1.67E+02 (1.25E+02) + & 1.06E+02 (2.47E+01) + \\
    F18   & 1.05E+02 (7.38E+01) & 1.04E+02 (7.86E+01) = & \textbf{8.98E+01 (5.34E+01) =} & \textbf{2.02E+04 (2.66E+04)} & 4.13E+04 (5.51E+04) + & 1.39E+05 (1.62E+05) + & 1.10E+03 (4.21E+03) & \textbf{8.36E+01 (7.68E+01) -} & 4.13E+04 (8.70E+04) = \\
    F19   & \textbf{1.07E+01 (4.98E+00)} & 1.30E+01 (8.48E+00) + & 1.29E+01 (7.23E+00) + & 3.34E+03 (2.52E+03) & \textbf{2.85E+03 (5.16E+03) =} & 1.78E+04 (1.22E+04) + & 1.59E+01 (1.03E+01) & \textbf{1.48E+01 (8.67E+00) =} & 1.82E+01 (1.75E+01) = \\
    F20   & 6.21E+01 (5.75E+01) & 2.48E+02 (1.24E+02) + & \textbf{5.93E+01 (3.64E+01) =} & \textbf{6.74E+01 (5.88E+01)} & 1.50E+02 (1.02E+02) + & 1.77E+02 (5.56E+01) + & \textbf{3.96E+01 (4.66E+01)} & 2.02E+02 (1.22E+02) + & 1.32E+02 (5.30E+01) + \\
    F21   & \textbf{2.19E+02 (3.61E+00)} & 2.40E+02 (1.25E+01) + & 2.21E+02 (3.85E+00) + & \textbf{2.25E+02 (5.50E+00)} & 2.42E+02 (1.11E+01) + & 2.48E+02 (7.20E+00) + & \textbf{2.28E+02 (4.40E+00)} & 2.47E+02 (1.69E+01) + & 2.34E+02 (5.93E+00) + \\
    F22   & \textbf{1.00E+02 (0.00E+00)} & 1.21E+02 (1.50E+02) = & \textbf{1.00E+02 (0.00E+00) =} & \textbf{1.00E+02 (0.00E+00)} & \textbf{1.00E+02 (0.00E+00) =} & \textbf{1.00E+02 (0.00E+00) =} & \textbf{1.00E+02 (0.00E+00)} & \textbf{1.00E+02 (0.00E+00) =} & \textbf{1.00E+02 (0.00E+00) =} \\
    F23   & \textbf{3.63E+02 (5.66E+00)} & 3.93E+02 (1.42E+01) + & 3.66E+02 (5.11E+00) + & \textbf{3.72E+02 (5.96E+00)} & 3.88E+02 (1.23E+01) + & 3.97E+02 (6.28E+00) + & \textbf{3.73E+02 (6.77E+00)} & 3.93E+02 (1.35E+01) + & 3.79E+02 (5.36E+00) + \\
    F24   & \textbf{4.37E+02 (4.83E+00)} & 4.42E+02 (7.40E+00) + & \textbf{4.37E+02 (4.67E+00) =} & \textbf{4.39E+02 (6.14E+00)} & 4.54E+02 (9.17E+00) + & 4.60E+02 (5.82E+00) + & \textbf{4.41E+02 (5.57E+00)} & 4.45E+02 (7.14E+00) + & 4.44E+02 (5.11E+00) + \\
    F25   & \textbf{3.87E+02 (0.00E+00)} & \textbf{3.87E+02 (0.00E+00) =} & \textbf{3.87E+02 (0.00E+00) =} & \textbf{3.87E+02 (0.00E+00)} & \textbf{3.87E+02 (0.00E+00) =} & \textbf{3.87E+02 (0.00E+00) =} & \textbf{3.87E+02 (0.00E+00)} & \textbf{3.87E+02 (0.00E+00) =} & \textbf{3.87E+02 (0.00E+00) =} \\
    F26   & 1.12E+03 (7.03E+01) & 1.49E+03 (2.45E+02) + & \textbf{1.11E+03 (1.31E+02) =} & \textbf{1.14E+03 (7.47E+01)} & 1.29E+03 (2.32E+02) + & 1.34E+03 (6.40E+01) + & \textbf{1.18E+03 (9.26E+01)} & 1.45E+03 (3.14E+02) + & 1.23E+03 (1.47E+02) + \\
    F27   & \textbf{5.03E+02 (8.78E+00)} & 5.10E+02 (9.00E+00) + & 5.05E+02 (6.25E+00) = & \textbf{5.00E+02 (5.55E+00)} & 5.05E+02 (6.37E+00) + & 5.02E+02 (5.84E+00) + & \textbf{4.99E+02 (7.23E+00)} & 5.06E+02 (7.11E+00) + & 5.01E+02 (5.89E+00) = \\
    F28   & 3.37E+02 (5.62E+01) & 3.36E+02 (5.53E+01) = & \textbf{3.21E+02 (4.27E+01) =} & 3.20E+02 (4.46E+01) & 3.22E+02 (4.64E+01) = & \textbf{3.19E+02 (4.25E+01) =} & \textbf{3.24E+02 (4.54E+01)} & 3.49E+02 (5.77E+01) = & 3.34E+02 (5.35E+01) = \\
    F29   & \textbf{4.34E+02 (3.21E+01)} & 5.30E+02 (9.22E+01) + & 4.78E+02 (2.91E+01) + & \textbf{4.35E+02 (2.99E+01)} & 4.81E+02 (6.72E+01) + & 5.29E+02 (2.59E+01) + & \textbf{4.31E+02 (3.06E+01)} & 4.80E+02 (5.83E+01) + & 5.06E+02 (3.90E+01) + \\
    F30   & \textbf{2.12E+03 (1.51E+02)} & 2.13E+03 (1.80E+02) = & 2.13E+03 (1.15E+02) = & 4.38E+03 (3.96E+03) & \textbf{4.23E+03 (4.13E+03) =} & 6.68E+03 (7.80E+03) = & 2.27E+03 (9.09E+02) & \textbf{2.15E+03 (1.86E+02) =} & \textbf{2.15E+03 (1.64E+02) =} \\
    \midrule
    +/=/- &       & 15/15/0 & 8/21/1 &       & 15/15/0 & 18/12/0 &       & 15/14/1 & 12/18/0 \\
    \bottomrule
    \end{tabular}%
  \end{adjustwidth}
  \label{tab:advanced_d30}%
  The symbols "+/=/-" show the statistical results of the Wilcoxon signed-rank test with $\alpha = 0.05$ significance level. "+" represents that ACM variant is significantly superior than the corresponding variant. "=" represents that the performance difference between ACM variant and the corresponding variant is not statistically significant. And, "-" represents that ACM variant is significantly inferior than the corresponding variant.
\end{table*}%

\begin{figure*}[htbp]
 \centering
 \subfigure[$F_{10}$]{
  \includegraphics[scale=0.25]{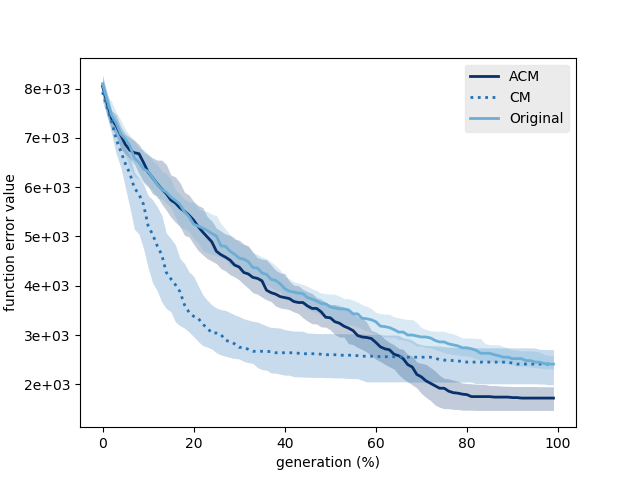}
   }
 \subfigure[$F_{16}$]{
  \includegraphics[scale=0.25]{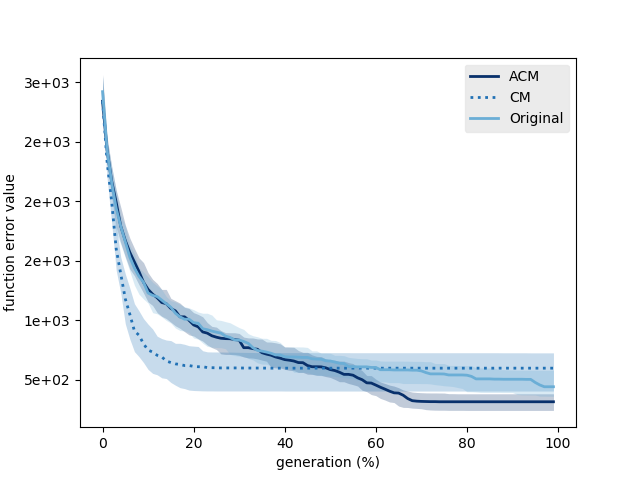}
   }
 \subfigure[$F_{20}$]{
  \includegraphics[scale=0.25]{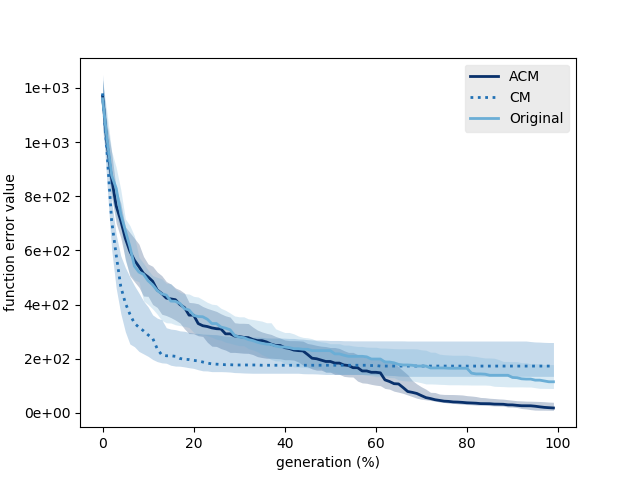}
   }
 \subfigure[$F_{26}$]{
  \includegraphics[scale=0.25]{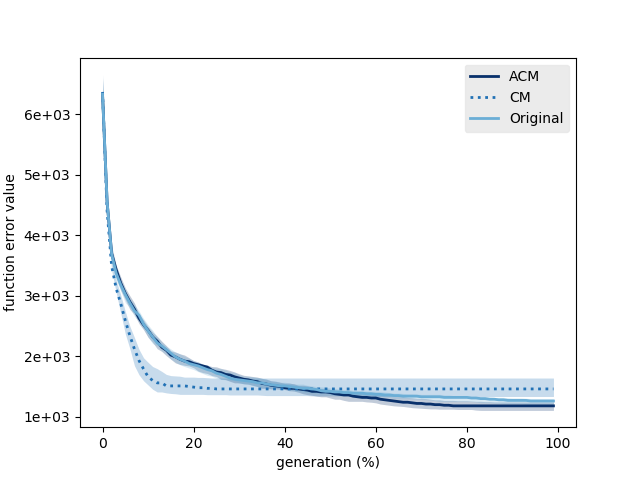}
   }
 \caption[]{Median and interquartile ranges (25th and 75th) of function error values with EDEV variants ($D = 30$)}
 \label{fig:convergenceGraphs_EDEV_d30}
\end{figure*}

\begin{table*}[htbp]
  \tiny
  \centering
  \caption{Averages and standard deviations of function error values with advanced DE variants ($D = 50$)}
    \begin{adjustwidth}{-0.75cm}{}
    \begin{tabular}{c|ccc||ccc||ccc}
    \toprule
          & \multicolumn{3}{c}{SaDE} & \multicolumn{3}{c}{EPSDE} & \multicolumn{3}{c}{CoDE} \\
          & ACM   & CM    & Original & ACM   & CM    & Original & ACM   & CM    & Original \\
          & MEAN (STD DEV) & MEAN (STD DEV) & MEAN (STD DEV) & MEAN (STD DEV) & MEAN (STD DEV) & MEAN (STD DEV) & MEAN (STD DEV) & MEAN (STD DEV) & MEAN (STD DEV) \\
    \midrule
    F1    & 2.13E+03 (1.97E+03) & \textbf{6.29E+02 (1.49E+03) -} & 9.46E+02 (2.04E+03) - & \textbf{3.31E-09 (4.72E-09)} & 1.80E-07 (1.17E-06) = & 6.73E-07 (3.74E-06) = & 3.99E+05 (1.42E+05) & \textbf{5.92E+03 (7.03E+03) -} & 1.73E+06 (6.29E+05) + \\
    F2    & \textbf{7.36E+14 (5.00E+15)} & 9.03E+14 (6.41E+15) - & 8.38E+24 (4.17E+25) + & 2.46E+08 (1.68E+09) & \textbf{4.98E+00 (2.26E+01) -} & 4.54E+32 (2.43E+33) + & \textbf{1.04E+20 (4.44E+20)} & 1.06E+40 (7.58E+40) - & 5.60E+49 (1.89E+50) + \\
    F3    & 2.86E+03 (1.09E+04) & 2.14E+04 (2.32E+04) + & \textbf{3.23E+02 (1.28E+03) =} & 1.02E+03 (2.65E+03) & \textbf{3.79E+02 (6.13E+02) =} & 5.86E+03 (1.22E+04) + & \textbf{3.99E+04 (1.06E+04)} & 4.35E+04 (1.20E+04) = & 9.84E+04 (1.06E+04) + \\
    F4    & 7.63E+01 (4.64E+01) & 5.87E+01 (5.19E+01) = & \textbf{5.16E+01 (4.12E+01) =} & 5.94E+01 (4.75E+01) & 5.68E+01 (4.72E+01) = & \textbf{5.51E+01 (4.26E+01) =} & 9.80E+01 (5.21E+01) & \textbf{8.70E+01 (5.68E+01) =} & 2.59E+02 (1.72E+01) + \\
    F5    & \textbf{6.66E+01 (1.17E+01)} & 9.79E+01 (2.20E+01) + & 8.31E+01 (1.04E+01) + & \textbf{7.03E+01 (1.35E+01)} & 1.12E+02 (2.75E+01) + & 1.91E+02 (1.75E+01) + & \textbf{9.00E+01 (1.68E+01)} & 1.30E+02 (2.91E+01) + & 3.13E+02 (1.69E+01) + \\
    F6    & 1.52E-06 (5.37E-06) & 1.91E-02 (5.19E-02) + & \textbf{1.14E-13 (7.65E-29) -} & 9.86E-08 (7.04E-07) & 1.22E-02 (3.09E-02) = & \textbf{1.14E-13 (7.65E-29) =} & 1.70E-02 (3.81E-03) & \textbf{2.21E-03 (9.16E-03) -} & 2.21E-02 (4.19E-03) + \\
    F7    & \textbf{1.28E+02 (1.37E+01)} & 1.43E+02 (2.27E+01) + & 1.48E+02 (1.65E+01) + & \textbf{1.09E+02 (1.40E+01)} & 1.55E+02 (2.54E+01) + & 2.42E+02 (1.56E+01) + & \textbf{1.46E+02 (2.12E+01)} & 1.80E+02 (2.37E+01) + & 4.15E+02 (1.93E+01) + \\
    F8    & \textbf{6.14E+01 (1.14E+01)} & 9.43E+01 (2.06E+01) + & 8.42E+01 (1.02E+01) + & \textbf{6.82E+01 (1.39E+01)} & 1.13E+02 (2.69E+01) + & 1.90E+02 (1.74E+01) + & \textbf{9.59E+01 (1.72E+01)} & 1.16E+02 (2.78E+01) + & 3.11E+02 (1.67E+01) + \\
    F9    & \textbf{1.09E+00 (2.31E+00)} & 4.47E+01 (7.11E+01) + & 3.76E+01 (6.07E+01) + & \textbf{2.56E+00 (6.41E+00)} & 1.09E+01 (2.93E+01) + & 2.63E+00 (3.60E+00) = & \textbf{5.92E-01 (1.43E-01)} & 1.46E+01 (2.37E+01) + & 2.26E+03 (4.74E+02) + \\
    F10   & \textbf{3.31E+03 (4.65E+02)} & 4.79E+03 (9.19E+02) + & 4.40E+03 (4.95E+02) + & \textbf{3.70E+03 (7.05E+02)} & 5.42E+03 (8.15E+02) + & 8.31E+03 (6.03E+02) + & \textbf{3.69E+03 (6.54E+02)} & 5.63E+03 (9.26E+02) + & 9.76E+03 (3.92E+02) + \\
    F11   & \textbf{6.29E+01 (2.10E+01)} & 8.89E+01 (3.00E+01) + & 6.33E+01 (1.59E+01) = & \textbf{9.02E+01 (5.85E+01)} & 1.37E+02 (5.24E+01) + & 1.34E+02 (8.50E+01) + & \textbf{5.78E+01 (1.08E+01)} & 6.57E+01 (2.29E+01) + & 2.11E+02 (2.24E+01) + \\
    F12   & \textbf{1.11E+05 (2.98E+05)} & 1.46E+05 (3.37E+05) = & 5.58E+05 (1.02E+06) + & 8.73E+03 (8.89E+03) & 8.37E+03 (5.54E+03) = & \textbf{7.41E+03 (4.62E+03) =} & 2.28E+06 (1.37E+06) & \textbf{1.11E+06 (7.65E+05) -} & 7.07E+07 (1.46E+07) + \\
    F13   & \textbf{1.47E+03 (1.99E+03)} & 3.21E+03 (3.57E+03) + & 3.28E+03 (1.26E+04) = & 3.39E+03 (4.87E+03) & \textbf{2.11E+03 (3.19E+03) =} & 3.91E+03 (4.53E+03) = & 2.86E+04 (1.95E+04) & \textbf{5.14E+03 (2.27E+03) -} & 3.76E+04 (2.98E+04) = \\
    F14   & \textbf{2.35E+03 (6.76E+03)} & 7.34E+03 (2.32E+04) = & 2.74E+04 (8.54E+04) = & 5.45E+02 (7.02E+02) & 6.77E+02 (1.60E+03) = & \textbf{3.72E+02 (2.70E+02) =} & 2.18E+02 (6.18E+01) & \textbf{2.12E+02 (7.74E+01) =} & 2.24E+02 (6.76E+01) = \\
    F15   & 1.24E+03 (3.67E+03) & \textbf{5.21E+02 (1.10E+03) =} & 9.62E+03 (1.15E+04) + & 3.95E+02 (2.03E+02) & 4.41E+02 (8.90E+02) - & \textbf{3.90E+02 (2.00E+02) =} & 4.33E+02 (1.51E+02) & \textbf{3.77E+02 (6.36E+01) -} & 4.96E+02 (1.28E+02) + \\
    F16   & \textbf{8.07E+02 (2.69E+02)} & 1.17E+03 (3.90E+02) + & 8.87E+02 (2.03E+02) + & \textbf{8.41E+02 (2.26E+02)} & 1.36E+03 (3.99E+02) + & 8.92E+02 (1.96E+02) = & \textbf{9.80E+02 (2.96E+02)} & 1.26E+03 (3.66E+02) + & 1.55E+03 (2.14E+02) + \\
    F17   & \textbf{5.00E+02 (1.69E+02)} & 8.55E+02 (3.18E+02) + & 6.05E+02 (1.57E+02) + & \textbf{6.61E+02 (1.85E+02)} & 1.06E+03 (2.81E+02) + & 6.84E+02 (1.94E+02) = & \textbf{6.80E+02 (2.42E+02)} & 8.82E+02 (2.35E+02) + & 9.57E+02 (1.54E+02) + \\
    F18   & \textbf{2.61E+04 (4.96E+04)} & 4.56E+05 (4.67E+05) + & 6.69E+05 (6.63E+05) + & 2.38E+03 (2.89E+03) & \textbf{1.65E+03 (1.22E+03) =} & 6.17E+04 (2.01E+05) = & 6.47E+04 (6.03E+04) & \textbf{5.38E+04 (3.78E+04) =} & 3.07E+05 (2.58E+05) + \\
    F19   & \textbf{1.41E+03 (3.47E+03)} & 1.04E+04 (8.60E+03) + & 1.73E+03 (3.82E+03) = & 1.55E+02 (5.96E+01) & 1.62E+02 (1.61E+02) = & \textbf{1.39E+02 (5.60E+01) =} & 1.66E+02 (5.10E+01) & \textbf{1.32E+02 (2.37E+01) -} & 1.68E+02 (4.32E+01) = \\
    F20   & \textbf{3.44E+02 (2.24E+02)} & 5.46E+02 (2.97E+02) + & 3.87E+02 (1.84E+02) = & \textbf{4.74E+02 (2.02E+02)} & 7.27E+02 (2.80E+02) + & 4.80E+02 (1.40E+02) = & \textbf{4.65E+02 (1.99E+02)} & 6.77E+02 (2.49E+02) + & 6.54E+02 (1.45E+02) + \\
    F21   & \textbf{2.73E+02 (1.20E+01)} & 2.97E+02 (2.46E+01) + & 2.81E+02 (1.37E+01) + & \textbf{2.74E+02 (1.60E+01)} & 3.16E+02 (2.83E+01) + & 3.98E+02 (1.82E+01) + & \textbf{2.91E+02 (1.52E+01)} & 3.28E+02 (2.71E+01) + & 5.17E+02 (1.50E+01) + \\
    F22   & \textbf{2.94E+03 (1.94E+03)} & 4.93E+03 (1.81E+03) + & 3.50E+03 (2.44E+03) = & \textbf{3.81E+03 (1.47E+03)} & 5.64E+03 (1.64E+03) + & 7.46E+03 (3.27E+03) + & \textbf{4.07E+03 (9.59E+02)} & 6.20E+03 (1.21E+03) + & 1.03E+04 (3.60E+02) + \\
    F23   & \textbf{4.84E+02 (1.27E+01)} & 5.15E+02 (2.62E+01) + & 5.04E+02 (1.44E+01) + & \textbf{4.97E+02 (2.31E+01)} & 5.56E+02 (3.06E+01) + & 6.14E+02 (1.75E+01) + & \textbf{5.15E+02 (2.29E+01)} & 5.56E+02 (3.08E+01) + & 7.34E+02 (1.57E+01) + \\
    F24   & \textbf{5.49E+02 (1.24E+01)} & 5.72E+02 (2.00E+01) + & 5.67E+02 (1.31E+01) + & \textbf{5.67E+02 (2.21E+01)} & 6.11E+02 (3.04E+01) + & 6.69E+02 (2.12E+01) + & \textbf{5.86E+02 (2.23E+01)} & 6.28E+02 (2.84E+01) + & 8.40E+02 (1.44E+01) + \\
    F25   & \textbf{5.24E+02 (2.62E+01)} & 5.43E+02 (3.47E+01) + & 5.26E+02 (3.38E+01) = & 5.32E+02 (3.74E+01) & \textbf{5.28E+02 (4.18E+01) =} & 5.31E+02 (3.83E+01) = & \textbf{4.80E+02 (2.35E+00)} & 5.14E+02 (4.12E+01) + & 5.79E+02 (2.09E+01) + \\
    F26   & \textbf{1.73E+03 (1.41E+02)} & 2.21E+03 (2.83E+02) + & 1.93E+03 (1.56E+02) + & \textbf{1.85E+03 (1.86E+02)} & 2.34E+03 (3.15E+02) + & 2.76E+03 (2.10E+02) + & \textbf{2.10E+03 (1.95E+02)} & 2.61E+03 (2.90E+02) + & 4.20E+03 (1.77E+02) + \\
    F27   & 5.39E+02 (1.50E+01) & 6.05E+02 (4.79E+01) + & \textbf{5.32E+02 (1.76E+01) -} & \textbf{5.82E+02 (5.19E+01)} & 6.04E+02 (5.08E+01) + & \textbf{5.82E+02 (5.14E+01) =} & \textbf{5.10E+02 (1.45E+01)} & 5.60E+02 (3.75E+01) + & 5.61E+02 (2.42E+01) + \\
    F28   & 4.91E+02 (2.27E+01) & 4.98E+02 (1.41E+01) = & \textbf{4.78E+02 (2.22E+01) =} & 4.94E+02 (1.96E+01) & 4.94E+02 (2.75E+01) = & \textbf{4.91E+02 (2.57E+01) =} & \textbf{4.59E+02 (0.00E+00)} & 4.82E+02 (2.32E+01) + & 4.78E+02 (1.39E+01) + \\
    F29   & \textbf{3.99E+02 (8.26E+01)} & 7.40E+02 (2.66E+02) + & 4.84E+02 (9.06E+01) + & \textbf{4.40E+02 (1.19E+02)} & 9.01E+02 (2.69E+02) + & 5.50E+02 (9.05E+01) + & \textbf{6.31E+02 (1.89E+02)} & 9.00E+02 (2.34E+02) + & 1.03E+03 (1.42E+02) + \\
    F30   & \textbf{6.10E+05 (4.80E+04)} & 7.56E+05 (1.26E+05) + & 6.58E+05 (9.74E+04) = & \textbf{6.43E+05 (5.96E+04)} & 6.49E+05 (5.03E+04) = & 6.76E+05 (1.06E+05) = & 6.38E+05 (3.86E+04) & \textbf{6.32E+05 (3.87E+04) =} & 7.23E+05 (7.89E+04) + \\
    \midrule
    +/=/- &       & 23/5/2 & 16/11/3 &       & 16/12/2 & 13/17/0 &       & 18/5/7 & 27/3/0 \\
    \midrule
    \midrule
          & \multicolumn{3}{c}{SHADE} & \multicolumn{3}{c}{MPEDE} & \multicolumn{3}{c}{EDEV} \\
          & ACM   & CM    & Original & ACM   & CM    & Original & ACM   & CM    & Original \\
          & MEAN (STD DEV) & MEAN (STD DEV) & MEAN (STD DEV) & MEAN (STD DEV) & MEAN (STD DEV) & MEAN (STD DEV) & MEAN (STD DEV) & MEAN (STD DEV) & MEAN (STD DEV) \\
    \midrule
    F1    & 9.03E-14 (1.55E-13) & \textbf{4.74E-14 (8.27E-14) =} & 6.63E-14 (9.21E-14) = & \textbf{4.18E-15 (6.53E-15)} & 5.29E-15 (6.93E-15) = & \textbf{4.18E-15 (6.53E-15) =} & 1.25E-12 (5.53E-12) & \textbf{1.89E-14 (2.95E-14) =} & 3.30E-13 (1.15E-12) = \\
    F2    & 3.04E-09 (9.79E-09) & 1.20E-09 (3.10E-09) = & \textbf{4.91E-10 (9.05E-10) =} & 3.49E+09 (2.49E+10) & \textbf{3.49E+00 (2.49E+01) =} & 8.80E+09 (6.29E+10) = & \textbf{3.89E+19 (2.77E+20)} & 1.34E+23 (9.56E+23) = & 4.02E+21 (2.87E+22) = \\
    F3    & 3.17E-13 (1.31E-13) & 2.93E-13 (1.15E-13) = & \textbf{2.88E-13 (9.05E-14) =} & 1.68E+04 (3.00E+04) & \textbf{1.06E+04 (2.42E+04) =} & 2.83E+04 (4.66E+04) = & \textbf{9.05E+03 (2.56E+04)} & 9.73E+03 (2.02E+04) = & 1.62E+04 (3.64E+04) = \\
    F4    & 3.73E+01 (4.51E+01) & \textbf{3.72E+01 (4.68E+01) =} & 4.40E+01 (4.47E+01) = & 5.41E+01 (4.10E+01) & \textbf{4.95E+01 (4.57E+01) =} & 5.89E+01 (4.74E+01) = & \textbf{5.82E+01 (4.85E+01)} & 6.22E+01 (4.78E+01) = & 5.95E+01 (4.95E+01) = \\
    F5    & \textbf{4.28E+01 (6.90E+00)} & 9.13E+01 (2.23E+01) + & 4.31E+01 (6.56E+00) = & \textbf{4.75E+01 (9.09E+00)} & 6.70E+01 (1.70E+01) + & 9.55E+01 (1.30E+01) + & \textbf{5.13E+01 (1.10E+01)} & 7.92E+01 (2.33E+01) + & 6.08E+01 (9.59E+00) + \\
    F6    & 9.35E-06 (2.94E-05) & \textbf{5.31E-06 (7.32E-06) =} & 6.54E-06 (1.14E-05) = & 8.43E-09 (5.38E-08) & 1.22E-08 (6.12E-08) = & \textbf{2.82E-09 (1.14E-08) =} & \textbf{9.29E-13 (2.86E-12)} & 6.52E-12 (1.32E-11) + & 3.26E-12 (1.15E-11) = \\
    F7    & \textbf{9.21E+01 (6.69E+00)} & 1.15E+02 (2.03E+01) + & 9.26E+01 (5.30E+00) = & \textbf{1.44E+02 (9.46E+00)} & 1.48E+02 (1.12E+01) + & 1.51E+02 (9.85E+00) + & \textbf{1.10E+02 (8.74E+00)} & 1.17E+02 (1.07E+01) + & 1.13E+02 (8.99E+00) = \\
    F8    & \textbf{4.19E+01 (6.94E+00)} & 9.07E+01 (2.30E+01) + & 4.64E+01 (6.01E+00) + & \textbf{4.56E+01 (8.45E+00)} & 7.08E+01 (1.57E+01) + & 9.66E+01 (9.61E+00) + & \textbf{4.92E+01 (1.16E+01)} & 8.23E+01 (2.10E+01) + & 6.39E+01 (9.52E+00) + \\
    F9    & \textbf{5.31E-01 (6.34E-01)} & 1.14E+00 (1.10E+00) + & 6.66E-01 (9.67E-01) = & 3.02E-01 (3.55E-01) & 4.66E-01 (6.83E-01) = & \textbf{2.93E-01 (3.65E-01) =} & 8.98E-01 (1.17E+00) & 9.86E-01 (9.62E-01) = & \textbf{7.44E-01 (8.01E-01) =} \\
    F10   & \textbf{3.09E+03 (4.21E+02)} & 4.22E+03 (6.51E+02) + & 3.56E+03 (3.32E+02) + & \textbf{3.43E+03 (5.14E+02)} & 4.59E+03 (7.22E+02) + & 6.41E+03 (3.78E+02) + & \textbf{3.33E+03 (4.93E+02)} & 4.42E+03 (6.50E+02) + & 4.44E+03 (3.94E+02) + \\
    F11   & 9.98E+01 (2.93E+01) & 1.18E+02 (3.46E+01) + & \textbf{9.89E+01 (2.64E+01) =} & \textbf{5.53E+01 (1.88E+01)} & 6.56E+01 (1.67E+01) + & 7.03E+01 (1.15E+01) + & \textbf{8.74E+01 (3.98E+01)} & 9.34E+01 (3.36E+01) = & 9.14E+01 (2.96E+01) = \\
    F12   & 6.40E+03 (5.25E+03) & \textbf{5.01E+03 (2.70E+03) =} & 5.81E+03 (5.42E+03) = & 7.90E+03 (5.31E+03) & \textbf{7.26E+03 (6.64E+03) =} & 8.39E+03 (6.13E+03) = & \textbf{6.86E+03 (5.80E+03)} & 7.18E+03 (4.91E+03) = & 7.95E+03 (4.63E+03) = \\
    F13   & 3.58E+02 (2.26E+02) & \textbf{3.56E+02 (2.49E+02) =} & 4.00E+02 (3.00E+02) = & \textbf{1.48E+03 (9.30E+03)} & 3.07E+03 (1.20E+04) = & 4.94E+03 (1.97E+04) = & 5.76E+02 (1.45E+03) & 6.81E+02 (1.92E+03) = & \textbf{3.73E+02 (5.13E+02) =} \\
    F14   & \textbf{2.28E+02 (5.63E+01)} & 2.45E+02 (7.19E+01) = & 2.33E+02 (6.97E+01) = & \textbf{2.40E+03 (3.86E+03)} & 6.49E+03 (9.59E+03) = & 6.17E+04 (6.33E+04) + & \textbf{1.55E+02 (5.14E+01)} & 1.57E+02 (7.12E+01) = & 1.65E+03 (8.26E+03) = \\
    F15   & 2.95E+02 (1.30E+02) & \textbf{2.88E+02 (1.19E+02) =} & 3.19E+02 (1.21E+02) = & 6.29E+03 (9.13E+03) & \textbf{2.70E+03 (5.00E+03) -} & 8.88E+03 (1.21E+04) = & 2.15E+02 (1.18E+02) & \textbf{2.00E+02 (9.75E+01) =} & 2.26E+02 (1.38E+02) = \\
    F16   & \textbf{6.69E+02 (1.84E+02)} & 7.60E+02 (2.44E+02) = & 7.63E+02 (1.65E+02) + & \textbf{7.03E+02 (1.85E+02)} & 9.15E+02 (3.09E+02) + & 1.12E+03 (2.06E+02) + & \textbf{7.84E+02 (2.29E+02)} & 9.55E+02 (2.85E+02) + & 1.00E+03 (1.75E+02) + \\
    F17   & \textbf{4.95E+02 (1.50E+02)} & 6.24E+02 (2.13E+02) + & 5.18E+02 (1.39E+02) = & \textbf{4.66E+02 (1.56E+02)} & 6.82E+02 (2.26E+02) + & 8.56E+02 (1.28E+02) + & \textbf{4.93E+02 (1.56E+02)} & 8.06E+02 (2.44E+02) + & 6.67E+02 (1.50E+02) + \\
    F18   & \textbf{1.89E+02 (1.10E+02)} & 2.47E+02 (1.69E+02) = & 1.92E+02 (1.19E+02) = & \textbf{2.52E+04 (4.95E+04)} & 8.92E+04 (1.91E+05) = & 1.63E+05 (4.09E+05) = & \textbf{3.73E+03 (1.51E+04)} & 5.46E+04 (2.07E+05) = & 1.25E+05 (2.89E+05) = \\
    F19   & \textbf{1.39E+02 (4.51E+01)} & 1.42E+02 (4.82E+01) = & 1.45E+02 (4.67E+01) = & 4.06E+03 (4.71E+03) & \textbf{2.66E+03 (4.43E+03) =} & 5.17E+03 (6.53E+03) = & \textbf{8.46E+01 (3.94E+01)} & 1.48E+02 (2.99E+02) + & 8.99E+01 (3.79E+01) = \\
    F20   & \textbf{2.61E+02 (1.31E+02)} & 4.94E+02 (2.36E+02) + & 3.14E+02 (1.21E+02) = & \textbf{3.42E+02 (1.64E+02)} & 4.85E+02 (2.49E+02) + & 7.65E+02 (1.19E+02) + & \textbf{3.33E+02 (1.41E+02)} & 5.74E+02 (2.37E+02) + & 5.64E+02 (1.30E+02) + \\
    F21   & \textbf{2.44E+02 (7.37E+00)} & 2.88E+02 (2.63E+01) + & 2.46E+02 (5.71E+00) = & \textbf{2.46E+02 (9.49E+00)} & 2.72E+02 (1.47E+01) + & 2.98E+02 (1.21E+01) + & \textbf{2.56E+02 (8.99E+00)} & 2.79E+02 (2.50E+01) + & 2.65E+02 (1.07E+01) + \\
    F22   & 3.27E+03 (1.33E+03) & 4.46E+03 (1.30E+03) + & \textbf{2.77E+03 (1.98E+03) =} & \textbf{1.83E+03 (2.05E+03)} & 3.65E+03 (2.82E+03) + & 3.95E+03 (3.31E+03) + & \textbf{3.26E+03 (1.65E+03)} & 3.61E+03 (2.24E+03) = & 4.10E+03 (2.00E+03) + \\
    F23   & \textbf{4.65E+02 (1.11E+01)} & 5.20E+02 (2.71E+01) + & 4.70E+02 (9.32E+00) = & \textbf{4.70E+02 (1.11E+01)} & 4.97E+02 (1.77E+01) + & 5.19E+02 (1.07E+01) + & \textbf{4.77E+02 (1.21E+01)} & 5.01E+02 (2.63E+01) + & 4.91E+02 (1.35E+01) + \\
    F24   & \textbf{5.37E+02 (7.96E+00)} & 5.44E+02 (1.07E+01) + & 5.39E+02 (9.19E+00) = & \textbf{5.31E+02 (7.17E+00)} & 5.55E+02 (1.22E+01) + & 5.71E+02 (1.29E+01) + & \textbf{5.40E+02 (9.15E+00)} & 5.49E+02 (8.51E+00) + & 5.45E+02 (8.41E+00) + \\
    F25   & 5.34E+02 (2.65E+01) & 5.32E+02 (3.52E+01) = & \textbf{5.28E+02 (3.60E+01) =} & 5.28E+02 (3.03E+01) & \textbf{5.27E+02 (3.26E+01) =} & 5.31E+02 (3.01E+01) = & 5.12E+02 (3.60E+01) & 5.17E+02 (3.25E+01) = & \textbf{5.10E+02 (3.59E+01) =} \\
    F26   & \textbf{1.52E+03 (1.08E+02)} & 2.07E+03 (2.66E+02) + & 1.53E+03 (1.21E+02) = & \textbf{1.51E+03 (8.87E+01)} & 1.73E+03 (1.65E+02) + & 1.87E+03 (1.11E+02) + & \textbf{1.62E+03 (1.17E+02)} & 2.02E+03 (2.99E+02) + & 1.71E+03 (1.29E+02) + \\
    F27   & \textbf{5.46E+02 (2.17E+01)} & 5.66E+02 (3.60E+01) + & 5.51E+02 (2.69E+01) = & \textbf{5.27E+02 (1.19E+01)} & 5.33E+02 (1.27E+01) + & 5.35E+02 (1.85E+01) + & \textbf{5.54E+02 (2.82E+01)} & 5.64E+02 (3.75E+01) = & \textbf{5.54E+02 (3.20E+01) =} \\
    F28   & \textbf{4.90E+02 (2.16E+01)} & 4.92E+02 (2.13E+01) = & 4.93E+02 (1.85E+01) = & 4.89E+02 (2.37E+01) & 4.93E+02 (2.22E+01) = & \textbf{4.87E+02 (2.41E+01) =} & 4.94E+02 (2.11E+01) & 4.93E+02 (2.24E+01) = & \textbf{4.87E+02 (2.35E+01) =} \\
    F29   & \textbf{4.19E+02 (8.69E+01)} & 7.25E+02 (2.29E+02) + & 4.70E+02 (8.30E+01) + & \textbf{3.70E+02 (3.57E+01)} & 4.83E+02 (1.05E+02) + & 5.55E+02 (5.74E+01) + & \textbf{4.01E+02 (7.37E+01)} & 5.47E+02 (1.53E+02) + & 5.21E+02 (8.81E+01) + \\
    F30   & 6.29E+05 (5.28E+04) & \textbf{6.28E+05 (5.61E+04) =} & 6.45E+05 (7.31E+04) = & 6.60E+05 (7.23E+04) & \textbf{6.50E+05 (7.55E+04) =} & 6.85E+05 (1.08E+05) = & 6.57E+05 (7.15E+04) & 6.56E+05 (6.49E+04) = & \textbf{6.49E+05 (6.30E+04) =} \\
    \midrule
    +/=/- &       & 15/15/0 & 4/26/0 &       & 15/14/1 & 16/14/0 &       & 14/16/0 & 12/18/0 \\
    \bottomrule
    \end{tabular}%
  \end{adjustwidth}
  \label{tab:advanced_d50}%
  The symbols "+/=/-" show the statistical results of the Wilcoxon signed-rank test with $\alpha = 0.05$ significance level. "+" represents that ACM variant is significantly superior than the corresponding variant. "=" represents that the performance difference between ACM variant and the corresponding variant is not statistically significant. And, "-" represents that ACM variant is significantly inferior than the corresponding variant.
\end{table*}%

\begin{figure*}[htbp]
 \centering
 \subfigure[$F_{10}$]{
  \includegraphics[scale=0.25]{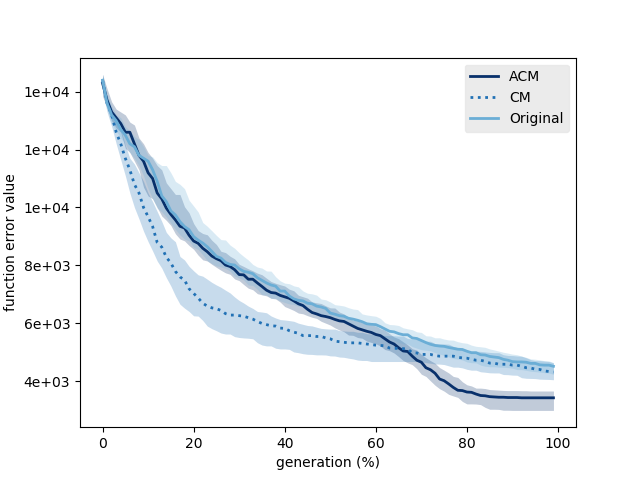}
   }
 \subfigure[$F_{16}$]{
  \includegraphics[scale=0.25]{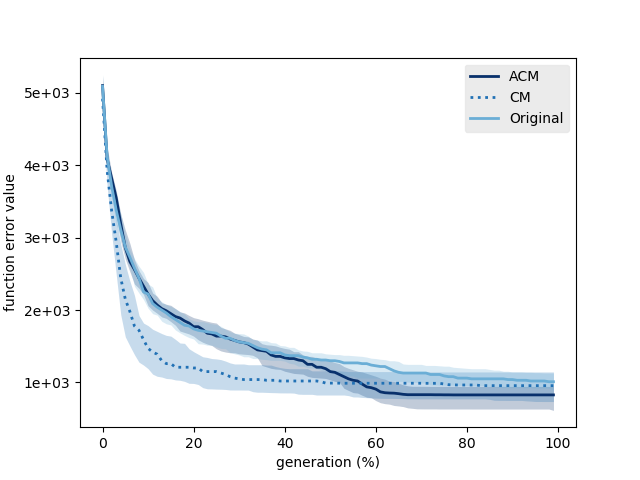}
   }
 \subfigure[$F_{20}$]{
  \includegraphics[scale=0.25]{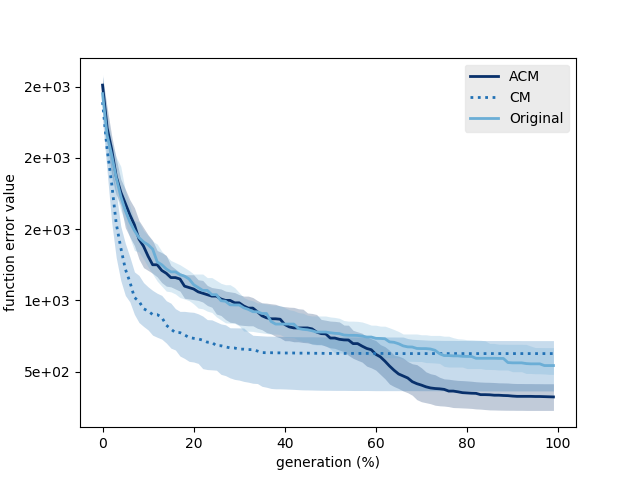}
   }
 \subfigure[$F_{26}$]{
  \includegraphics[scale=0.25]{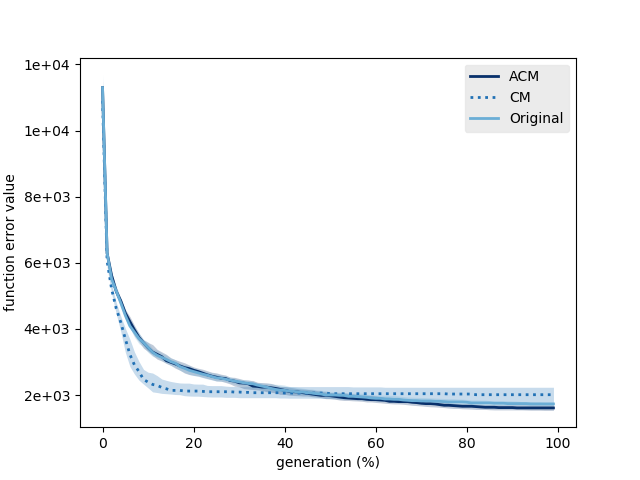}
   }
 \caption[]{Median and interquartile ranges (25th and 75th) of function error values with EDEV variants ($D = 50$)}
 \label{fig:convergenceGraphs_EDEV_d50}
\end{figure*}

Experiments and comparisons were carried out with six conventional and six advanced DE variants to demonstrate the effectiveness of the proposed Cauchy mutation.

\subsection{Conventional DE variants}
We applied the proposed Cauchy mutation to six conventional DE variants, namely DE/rand/1/bin, DE/best/1/bin, DE/current-to-best/1/bin, DE/current-to-rand/1, DE/rand/2/bin, and DE/current-to-best/2/bin. For all of the test algorithms, the scaling factor $F$, the crossover rate $CR$, and the population size $NP$ are initialized to 0.5, 0.5, and 100, respectively. Additionally, $FT_{init} = 100$ and $FT_{fin} = 5$ are used for the proposed Cauchy mutation (ACM), and $FT = 5$ is used for the previous Cauchy mutation (CM) as same as in \cite{ali2011improving}.

Table \ref{tab:conventional_d30} presents the averages and the standard deviations of the FEVs obtained by independently running each algorithm 51 times at 30 dimensions. The summary of the experimental results is as follows.

\begin{itemize}
\item DE/rand/1/bin and DE/rand/2/bin: These mutation strategies are designed for higher exploration. ACM-DE/rand/1/bin finds 17/4 statistically better/worse solutions compared to CM-DE/rand/1/bin and 24/2 compared to DE/rand/1/bin. Also, ACM-DE/rand/2/bin finds 14/8 statistically better/worse solutions compared to CM-DE/rand/2/bin and 28/0 compared to DE/rand/2/bin.
\item DE/best/1/bin, DE/current-to-best/1/bin, and DE/current-to-best/2/bin: These mutation strategies are designed for higher exploitation. 
ACM-DE/best/1/bin finds 15/1 statistically better/worse solutions compared to CM-DE/best/1/bin and 6/1 compared to DE/best/1/bin.
Also, ACM-DE/current-to-best/1/bin finds 18/4 statistically better/worse solutions compared to CM-DE/current-to-best/1/bin and 19/0 compared to DE/current-to-best/1/bin, and ACM-DE/current-to-best/2/bin finds 20/4 statistically better/worse solutions compared to CM-DE/current-to-best/2/bin and 25/2 compared to DE/current-to-best/2/bin.
\item DE/current-to-rand/1: This mutation strategy is designed for rotation invariant. ACM-DE/current-to-rand/1 finds 21/3 statistically better/worse solutions compared to CM-DE/current-to-rand/1 and 22/0 compared to DE/current-to-rand/1.
\end{itemize}

Although CM-DE variants are more greedy than ACM-DE variants, the performance differences between them on the unimodal functions ($F_{1}$-$F_{3}$) are insignificant. On the contrary, ACM-DE variants yield significantly better performance than CM-DE variants on the multimodal functions ($F_{4}$-$F_{30}$). These observations are supported by the convergence graphs of DE/rand/1/bin variants from Fig. \ref{fig:convergenceGraphs_rand1bin_d30}. We selected DE/rand/1/bin because it is the most widely used one among the six conventional DE variants. As shown in the figures, the convergence speed of CM-DE/rand/1/bin is faster than ACM-DE/rand/1/bin, but it gets trapped in a local optimum during the search process. On the other hand, the convergence speed of ACM-DE/rand/1/bin is more robust than CM-DE/rand/1/bin. We can see a similar tendency at 50 dimensions from Table \ref{tab:conventional_d50} and Fig. \ref{fig:convergenceGraphs_rand1bin_d50}.

As a result, the proposed Cauchy mutation can enhance the performance of conventional DE variants that have different characteristics such as higher exploration, higher exploitation, and rotation invariant, especially for multimodal functions.

\subsection{Advanced DE variants}
Although conventional DE variants serve as the basis for many advanced DE variants, further investigation with advanced DE variants may be intriguing. We applied the proposed Cauchy mutation to six advanced DE variants, namely SaDE \cite{qin2008differential}, EPSDE \cite{mallipeddi2011differential}, CoDE \cite{wang2011differential}, SHADE \cite{tanabe2013success}, MPEDE \cite{wu2016differential}, and EDEV \cite{wu2018ensemble}. For all of the test algorithms, the control parameters are initialized to the recommended values by their authors. Additionally, $FT_{init} = 100$ and $FT_{fin} = 5$ are used for the proposed Cauchy mutation, and $FT = 5$ is used for the previous Cauchy mutation as same as in \cite{ali2011improving}.

Table \ref{tab:advanced_d30} presents the averages and the standard deviations of the FEVs obtained by independently running each algorithm 51 times at 30 dimensions. The summary of the experimental results is as follows.

\begin{itemize}
\item SaDE \cite{qin2008differential} and EPSDE \cite{mallipeddi2011differential}: These are the self-adaptive DE variants. ACM-SaDE finds 17/4 statistically better/worse solutions compared to CM-SaDE and 14/3 compared to SaDE. Also, ACM-EPSDE finds 14/3 statistically better/worse solutions compared to CM-EPSDE and 11/0 compared to EPSDE.
\item CoDE \cite{wang2011differential}: This DE variant uses three composite mutation strategies. ACM-CoDE finds 16/8 statistically better/worse solutions compared to CM-CoDE and 23/0 compared to CoDE.
\item SHADE \cite{tanabe2013success}: This DE variant uses an advanced mutation strategy. ACM-SHADE finds 15/0 statistically better/worse solutions compared to CM-SHADE and 8/1 compared to SHADE.
\item MPEDE \cite{wu2016differential} and EDEV \cite{wu2018ensemble}: These are the multi-population-based DE variants. ACM-MPEDE finds 15/0 statistically better/worse solutions compared to CM-MPEDE and 18/0 compared to MPEDE. Also, ACM-EDEV finds 15/1 statistically better/worse solutions compared to CM-EDEV and 12/0 compared to EDEV.
\end{itemize}

Although CM-DE variants are more greedy than ACM-DE variants, the performance differences between them on the unimodal functions ($F_{1}$-$F_{3}$) are insignificant. On the contrary, ACM-DE variants yield significantly better performance than CM-DE variants on the multimodal functions ($F_{4}$-$F_{30}$). These observations are supported by the convergence graphs of EDEV variants from Fig. \ref{fig:convergenceGraphs_EDEV_d30}. We selected EDEV because it is the most recently proposed one among the six advanced DE variants. Additionally, EDEV is a multi-population-based DE variant that uses three DE variants that have different characteristics. As shown in the figures, the convergence speed of CM-EDEV is faster than ACM-EDEV, but it gets trapped in a local optimum during the search process. On the other hand, the convergence speed of ACM-EDEV is more robust than CM-EDEV. We can see a similar tendency at 50 dimensions from Table \ref{tab:advanced_d50} and Fig. \ref{fig:convergenceGraphs_EDEV_d50}.

As a result, the proposed Cauchy mutation can enhance the performance of advanced DE variants, especially for multimodal functions.


\section{Analysis of ACM-DE}
\label{sec:Analysis}
We have insisted that the performance of ACM-DE is due to the sigmoid based parameter control and the $p$-best individual based Cauchy mutation. The advantages of using the $p$-best individual information over using the best individual information have been verified successfully \cite{zhang2009jade, tanabe2013success, tanabe2014improving}. Therefore, we performed a series of experiments to verify the contribution of the sigmoid based parameter control only.

\subsection{Sigmoid Based Parameter Control}
We carried out experiments with four parameter controls for the failure threshold as follows.

\begin{enumerate}
\item SFTD: decreases $FT$ as a sigmoid function of $g$.
\item SFTI: increases $FT$ as a sigmoid function of $g$.
\item LFTD: decreases $FT$ as a linear function of $g$.
\item LFTI: increases $FT$ as a linear function of $g$.
\end{enumerate}

There are two failure threshold decrease approaches (SFTD and LFTD) and two failure threshold increase approaches (SFTI and LFTI). SFTD and LFTD assign a high failure threshold at the beginning of the search process and gradually reduces it over generations, while SFTI and LFTI do the opposite. We used DE/rand/1/bin as a conventional DE variant and EDEV as an advanced DE variant. For ACM-DE/rand/1/bin, the scaling factor $F$, the crossover rate $CR$, and the population size $NP$ are initialized to 0.5, 0.5, and 100, respectively, and for ACM-EDEV, the control parameters are initialized to the recommended values by their authors. Tables \ref{tab:sigmoid_rand1bin} and \ref{tab:sigmoid_EDEV} present the averages and the standard deviations of the FEVs obtained by independently running DE/rand/1/bin variants and EDEV variants 51 times at 30 dimensions. As we can see from the tables, both DE/rand/1/bin and EDEV work best with SFTD. SFTD and LFTD yield significantly better performance than SFTI and LFTI because SFTD and LFTD can establish a balance between exploration and exploitation, while SFTI and LFTI cannot. Also, SFTD yields slightly better performance than LFTD because SFTD has higher exploration in the first half of the search process and higher exploitation in the second half of the search process.

\begin{table}[htbp]
  \tiny
  \centering
  \caption{DE/rand/1/bin with failure threshold decrease and increase approaches ($D = 30$)}
    \begin{adjustwidth}{0.0cm}{}
    \begin{tabular}{ccccc}
    \toprule
          & \multicolumn{4}{c}{DE/rand/1/bin} \\
          & SFTD  & SFTI  & LFTD  & LFTI \\
          & MEAN (STD DEV) & MEAN (STD DEV) & MEAN (STD DEV) & MEAN (STD DEV) \\
    \midrule
    F1    & 2.71E+02 (1.44E+02) & \textbf{5.87E+00 (4.15E+00) -} & 2.06E+02 (8.35E+01) - & 1.93E+01 (1.08E+01) - \\
    F2    & 1.73E+07 (9.97E+07) & \textbf{9.66E+05 (4.27E+06) -} & 4.05E+09 (1.17E+10) + & 7.84E+08 (1.99E+09) + \\
    F3    & 4.76E+03 (1.44E+03) & \textbf{4.51E+03 (2.52E+03) =} & 1.16E+04 (3.36E+03) + & 1.18E+04 (3.13E+03) + \\
    F4    & 8.52E+01 (2.36E-01) & \textbf{8.07E+01 (2.28E+01) =} & 8.52E+01 (2.28E-01) = & 8.34E+01 (2.37E+01) = \\
    F5    & 4.27E+01 (1.14E+01) & 6.33E+01 (1.76E+01) + & \textbf{3.99E+01 (1.20E+01) =} & 5.97E+01 (1.61E+01) + \\
    F6    & 3.95E-12 (3.52E-12) & \textbf{1.65E-13 (6.12E-14) -} & 1.68E-12 (1.16E-12) - & 1.94E-13 (6.11E-14) - \\
    F7    & \textbf{6.68E+01 (1.05E+01)} & 9.19E+01 (1.67E+01) + & 6.79E+01 (1.08E+01) = & 8.72E+01 (1.49E+01) + \\
    F8    & \textbf{3.79E+01 (1.21E+01)} & 6.59E+01 (1.86E+01) + & 3.82E+01 (1.16E+01) = & 6.18E+01 (1.71E+01) + \\
    F9    & 4.47E-15 (2.23E-14) & 2.61E+00 (1.26E+01) = & \textbf{2.24E-15 (1.60E-14) =} & \textbf{2.24E-15 (1.60E-14) =} \\
    F10   & \textbf{2.32E+03 (6.53E+02)} & 2.90E+03 (5.48E+02) + & 2.33E+03 (5.95E+02) = & 2.85E+03 (5.26E+02) + \\
    F11   & 2.66E+01 (2.47E+01) & 6.65E+01 (3.89E+01) + & \textbf{2.51E+01 (2.24E+01) =} & 5.10E+01 (3.40E+01) + \\
    F12   & 1.22E+05 (7.00E+04) & \textbf{7.79E+04 (3.48E+04) -} & 1.65E+05 (8.15E+04) + & 9.33E+04 (4.54E+04) - \\
    F13   & 1.29E+04 (5.15E+03) & \textbf{1.16E+04 (1.25E+04) =} & 1.38E+04 (5.26E+03) = & 1.18E+04 (1.29E+04) = \\
    F14   & \textbf{9.87E+01 (9.72E+00)} & 2.86E+02 (5.35E+02) + & 1.01E+02 (8.52E+00) = & 1.25E+02 (6.54E+01) + \\
    F15   & \textbf{1.93E+02 (3.55E+01)} & 7.08E+02 (8.10E+02) + & 2.52E+02 (4.18E+01) + & 1.12E+03 (2.38E+03) + \\
    F16   & \textbf{5.34E+02 (2.11E+02)} & 8.73E+02 (3.06E+02) + & 5.59E+02 (2.36E+02) = & 9.22E+02 (2.98E+02) + \\
    F17   & 7.07E+01 (6.99E+01) & 3.25E+02 (1.60E+02) + & \textbf{6.90E+01 (8.02E+01) =} & 2.87E+02 (1.65E+02) + \\
    F18   & 1.12E+05 (1.06E+05) & 2.14E+05 (1.79E+05) + & \textbf{9.98E+04 (9.78E+04) =} & 1.65E+05 (1.43E+05) + \\
    F19   & \textbf{8.64E+01 (1.70E+01)} & 1.23E+03 (2.89E+03) + & 9.36E+01 (1.87E+01) = & 7.12E+02 (1.46E+03) + \\
    F20   & \textbf{8.32E+01 (9.78E+01)} & 3.74E+02 (1.52E+02) + & 8.67E+01 (9.01E+01) = & 3.59E+02 (1.87E+02) + \\
    F21   & \textbf{2.40E+02 (1.04E+01)} & 2.64E+02 (1.54E+01) + & 2.43E+02 (1.15E+01) = & 2.67E+02 (1.66E+01) + \\
    F22   & \textbf{1.00E+02 (0.00E+00)} & 2.36E+02 (6.85E+02) = & \textbf{1.00E+02 (0.00E+00) =} & 2.31E+02 (6.61E+02) = \\
    F23   & \textbf{3.85E+02 (1.33E+01)} & 4.15E+02 (1.82E+01) + & 3.88E+02 (1.16E+01) = & 4.08E+02 (1.89E+01) + \\
    F24   & \textbf{4.62E+02 (1.11E+01)} & 4.86E+02 (1.93E+01) + & 4.63E+02 (1.06E+01) = & 4.78E+02 (1.46E+01) + \\
    F25   & \textbf{3.87E+02 (0.00E+00)} & \textbf{3.87E+02 (0.00E+00) =} & \textbf{3.87E+02 (0.00E+00) =} & \textbf{3.87E+02 (7.84E-01) =} \\
    F26   & \textbf{1.36E+03 (1.31E+02)} & 1.78E+03 (2.00E+02) + & 1.37E+03 (1.48E+02) = & 1.68E+03 (2.35E+02) + \\
    F27   & \textbf{5.00E+02 (6.24E+00)} & 5.18E+02 (9.17E+00) + & 5.01E+02 (5.14E+00) = & 5.11E+02 (7.86E+00) + \\
    F28   & \textbf{3.63E+02 (4.46E+01)} & 3.73E+02 (6.41E+01) = & 3.80E+02 (4.35E+01) + & \textbf{3.63E+02 (5.52E+01) =} \\
    F29   & 5.03E+02 (9.24E+01) & 7.78E+02 (1.66E+02) + & \textbf{4.93E+02 (1.06E+02) =} & 6.85E+02 (1.84E+02) + \\
    F30   & 1.05E+04 (2.42E+03) & \textbf{8.60E+03 (3.46E+03) -} & 1.43E+04 (2.79E+03) + & 9.26E+03 (3.12E+03) - \\
    \midrule
    +/=/- &       & 18/7/5 & 6/22/2 & 20/6/4 \\
    \bottomrule
    \end{tabular}%
  \end{adjustwidth}
  \label{tab:sigmoid_rand1bin}%
  The symbols "+/=/-" show the statistical results of the Wilcoxon signed-rank test with $\alpha = 0.05$ significance level. "+" represents that SFTD variant is significantly superior than the corresponding variant. "=" represents that the performance difference between SFTD variant and the corresponding variant is not statistically significant. And, "-" represents that SFTD variant is significantly inferior than the corresponding variant.
\end{table}%

\begin{table}[htbp]
  \tiny
  \centering
  \caption{EDEV with failure threshold decrease and increase approaches ($D = 30$)}
    \begin{adjustwidth}{0.0cm}{}
    \begin{tabular}{ccccc}
    \toprule
          & \multicolumn{4}{c}{EDEV} \\
          & SFTD  & SFTI  & LFTD  & LFTI \\
          & MEAN (STD DEV) & MEAN (STD DEV) & MEAN (STD DEV) & MEAN (STD DEV) \\
    \midrule
    F1    & 5.57E-16 (2.78E-15) & 8.35E-16 (3.37E-15) = & \textbf{0.00E+00 (0.00E+00) =} & 8.35E-16 (3.37E-15) = \\
    F2    & 7.69E+07 (5.16E+08) & \textbf{1.26E-06 (3.59E-06) =} & 1.43E+10 (8.13E+10) = & 2.70E+10 (1.15E+11) = \\
    F3    & 3.81E+02 (1.57E+03) & \textbf{2.94E+02 (1.91E+03) =} & 1.60E+03 (5.30E+03) = & 1.27E+03 (5.65E+03) = \\
    F4    & 5.14E+01 (2.06E+01) & \textbf{4.83E+01 (2.40E+01) =} & 5.49E+01 (1.57E+01) = & 5.19E+01 (2.08E+01) = \\
    F5    & 2.53E+01 (5.26E+00) & 5.86E+01 (1.72E+01) + & \textbf{2.40E+01 (5.55E+00) =} & 4.65E+01 (1.45E+01) + \\
    F6    & 1.11E-11 (5.05E-11) & \textbf{4.34E-13 (5.60E-13) -} & 4.30E-12 (1.27E-11) = & 2.02E-12 (6.21E-12) - \\
    F7    & \textbf{5.70E+01 (5.11E+00)} & 7.55E+01 (1.27E+01) + & 5.95E+01 (4.62E+00) + & 6.01E+01 (8.63E+00) = \\
    F8    & \textbf{2.47E+01 (6.09E+00)} & 5.74E+01 (1.64E+01) + & 2.66E+01 (6.32E+00) = & 4.91E+01 (1.70E+01) + \\
    F9    & 1.07E-02 (6.45E-02) & 1.87E-01 (8.92E-01) = & \textbf{1.75E-03 (1.25E-02) =} & 3.51E-03 (1.75E-02) = \\
    F10   & 1.62E+03 (3.82E+02) & 2.80E+03 (5.90E+02) + & \textbf{1.52E+03 (4.61E+02) =} & 2.49E+03 (4.60E+02) + \\
    F11   & 1.94E+01 (2.19E+01) & 4.38E+01 (2.84E+01) + & \textbf{1.54E+01 (1.54E+01) =} & 2.58E+01 (2.36E+01) = \\
    F12   & \textbf{1.32E+03 (6.03E+02)} & 1.96E+03 (2.49E+03) = & 2.21E+03 (3.79E+03) = & 1.72E+03 (1.63E+03) = \\
    F13   & 6.17E+01 (5.47E+01) & 6.98E+01 (5.76E+01) = & \textbf{6.04E+01 (8.58E+01) =} & 6.40E+01 (5.33E+01) = \\
    F14   & 3.64E+01 (1.56E+01) & 3.76E+01 (9.77E+00) = & 3.65E+01 (1.66E+01) = & \textbf{3.52E+01 (1.12E+01) =} \\
    F15   & 2.88E+01 (2.26E+01) & \textbf{2.51E+01 (1.83E+01) =} & 2.72E+01 (1.74E+01) = & 2.79E+01 (1.88E+01) = \\
    F16   & \textbf{3.08E+02 (1.25E+02)} & 8.64E+02 (2.53E+02) + & 3.33E+02 (1.53E+02) = & 8.54E+02 (3.13E+02) + \\
    F17   & \textbf{4.13E+01 (2.93E+01)} & 2.89E+02 (1.77E+02) + & 4.57E+01 (2.09E+01) + & 2.05E+02 (1.27E+02) + \\
    F18   & 7.44E+02 (3.81E+03) & \textbf{1.56E+02 (4.26E+02) =} & 6.78E+02 (3.17E+03) = & 8.66E+02 (4.04E+03) = \\
    F19   & 1.88E+01 (1.64E+01) & \textbf{1.48E+01 (4.10E+00) =} & 1.61E+01 (8.53E+00) = & 2.12E+01 (1.91E+01) = \\
    F20   & \textbf{3.66E+01 (4.67E+01)} & 2.75E+02 (1.45E+02) + & 6.17E+01 (5.35E+01) + & 2.10E+02 (1.48E+02) + \\
    F21   & 2.28E+02 (6.25E+00) & 2.61E+02 (2.82E+01) + & \textbf{2.27E+02 (5.76E+00) =} & 2.54E+02 (1.75E+01) + \\
    F22   & \textbf{1.00E+02 (0.00E+00)} & \textbf{1.00E+02 (2.80E-01) =} & \textbf{1.00E+02 (0.00E+00) =} & \textbf{1.00E+02 (0.00E+00) =} \\
    F23   & \textbf{3.71E+02 (7.99E+00)} & 4.07E+02 (1.82E+01) + & 3.72E+02 (7.29E+00) = & 3.98E+02 (1.41E+01) + \\
    F24   & 4.42E+02 (5.47E+00) & 4.78E+02 (1.92E+01) + & \textbf{4.41E+02 (5.97E+00) =} & 4.74E+02 (1.57E+01) + \\
    F25   & \textbf{3.87E+02 (0.00E+00)} & \textbf{3.87E+02 (5.60E-01) =} & \textbf{3.87E+02 (0.00E+00) =} & \textbf{3.87E+02 (0.00E+00) =} \\
    F26   & \textbf{1.18E+03 (1.49E+02)} & 1.67E+03 (2.06E+02) + & \textbf{1.18E+03 (8.62E+01) =} & 1.50E+03 (1.71E+02) + \\
    F27   & \textbf{5.00E+02 (8.39E+00)} & 5.11E+02 (9.11E+00) + & \textbf{5.00E+02 (7.55E+00) =} & 5.07E+02 (8.12E+00) + \\
    F28   & 3.29E+02 (5.14E+01) & \textbf{3.28E+02 (5.33E+01) =} & 3.34E+02 (5.09E+01) = & 3.30E+02 (5.03E+01) = \\
    F29   & \textbf{4.31E+02 (3.01E+01)} & 6.53E+02 (1.50E+02) + & 4.32E+02 (3.56E+01) = & 5.63E+02 (1.22E+02) + \\
    F30   & \textbf{2.12E+03 (1.90E+02)} & 2.13E+03 (1.48E+02) = & 2.46E+03 (1.60E+03) = & 2.16E+03 (1.50E+02) + \\
    \midrule
    +/=/- &       & 14/15/1 & 3/27/0 & 13/16/1 \\
    \bottomrule
    \end{tabular}%
  \end{adjustwidth}
  \label{tab:sigmoid_EDEV}%
  The symbols "+/=/-" show the statistical results of the Wilcoxon signed-rank test with $\alpha = 0.05$ significance level. "+" represents that SFTD variant is significantly superior than the corresponding variant. "=" represents that the performance difference between SFTD variant and the corresponding variant is not statistically significant. And, "-" represents that SFTD variant is significantly inferior than the corresponding variant.
\end{table}%

\subsection{Failure Threshold}
ACM-DE introduces two control parameters $FT_{init}$ and $FT_{fin}$ that determine a new failure threshold at each generation. Therefore, further investigation to find appropriate control parameters may be intriguing. We carried out experiments with seven initial failure thresholds $FT_{init} \in \{30, 50, 80, 100, 130, 150, 180\}$ and one final failure threshold $FT_{fin} = 5$. We used DE/rand/1/bin as a conventional DE variant and EDEV as an advanced DE variant. For ACM-DE/rand/1/bin, the scaling factor $F$, the crossover rate $CR$, and the population size $NP$ are initialized to 0.5, 0.5, and 100, respectively, and for ACM-EDEV, the control parameters are initialized to the recommended values by their authors. Tables \ref{tab:failure_threshold_rand1bin} and \ref{tab:failure_threshold_EDEV} present the averages and the standard deviations of the FEVs obtained by independently running DE/rand/1/bin variants and EDEV variants 51 times at 30 dimensions. As we can see from the tables, both ACM-DE/rand/1/bin and ACM-EDEV work best with $FT_{init} \in [80, 130]$. As a result, the setting of $FT_{init} = 100$ and $FT_{fin} = 5$ is considered as a standard setting of ACM-DE.

\begin{table*}[htp]
  \tiny
  \centering
  \caption{ACM-DE/rand/1/bin with different failure threshold settings ($D = 30$)}
    \begin{tabular}{cccccccc}
    \toprule
          & \multicolumn{7}{c}{DE/rand/1/bin} \\
          & $FT_{init}=100$ & $FT_{init}=30$ & $FT_{init}=50$ & $FT_{init}=80$ & $FT_{init}=130$ & $FT_{init}=150$ & $FT_{init}=180$ \\
          & MEAN (STD DEV) & MEAN (STD DEV) & MEAN (STD DEV) & MEAN (STD DEV) & MEAN (STD DEV) & MEAN (STD DEV) & MEAN (STD DEV) \\
    \midrule
    F1    & 2.93E+02 (1.28E+02) & 2.84E+02 (1.47E+02) = & 3.25E+02 (1.50E+02) = & 2.92E+02 (1.46E+02) = & 3.18E+02 (1.45E+02) = & 2.89E+02 (1.25E+02) = & \textbf{2.71E+02 (1.20E+02) =} \\
    F2    & 3.63E+06 (1.10E+07) & \textbf{1.54E+04 (8.83E+04) -} & 2.39E+04 (5.91E+04) - & 6.84E+05 (3.22E+06) - & 3.39E+08 (1.90E+09) + & 2.39E+08 (1.03E+09) + & 3.17E+08 (9.33E+08) + \\
    F3    & 4.85E+03 (1.87E+03) & \textbf{1.56E+03 (7.59E+02) -} & 2.68E+03 (1.25E+03) - & 4.04E+03 (1.55E+03) - & 6.39E+03 (1.91E+03) + & 6.85E+03 (2.25E+03) + & 6.42E+03 (1.94E+03) + \\
    F4    & 8.52E+01 (2.51E-01) & \textbf{8.51E+01 (5.18E-01) =} & 8.52E+01 (1.88E-01) = & 8.53E+01 (2.15E-01) = & 8.52E+01 (2.18E-01) = & 8.52E+01 (2.41E-01) = & 8.53E+01 (2.44E-01) = \\
    F5    & 3.83E+01 (1.19E+01) & 4.17E+01 (1.30E+01) = & 4.17E+01 (1.24E+01) = & 3.97E+01 (1.06E+01) = & \textbf{3.70E+01 (1.11E+01) =} & 3.86E+01 (1.15E+01) = & 3.92E+01 (1.09E+01) = \\
    F6    & 2.86E-12 (2.45E-12) & 9.08E-12 (6.28E-12) + & 7.39E-12 (5.89E-12) + & 4.74E-12 (3.15E-12) + & 3.26E-12 (2.39E-12) = & 3.49E-12 (4.10E-12) = & \textbf{2.77E-12 (2.64E-12) =} \\
    F7    & 6.59E+01 (1.01E+01) & 7.14E+01 (1.20E+01) + & 6.96E+01 (1.31E+01) = & 6.58E+01 (9.68E+00) = & 6.75E+01 (8.83E+00) = & 6.84E+01 (1.21E+01) = & \textbf{6.56E+01 (9.52E+00) =} \\
    F8    & 4.02E+01 (1.17E+01) & 4.44E+01 (1.17E+01) = & 4.35E+01 (1.26E+01) = & 4.06E+01 (1.14E+01) = & \textbf{3.72E+01 (1.26E+01) =} & 3.90E+01 (1.04E+01) = & 3.79E+01 (1.09E+01) = \\
    F9    & \textbf{4.47E-15 (2.23E-14)} & 8.94E-15 (3.10E-14) = & 1.34E-14 (3.71E-14) = & \textbf{4.47E-15 (2.23E-14) =} & 6.71E-15 (2.71E-14) = & 6.71E-15 (2.71E-14) = & 1.12E-14 (3.42E-14) = \\
    F10   & 2.29E+03 (6.10E+02) & 2.43E+03 (5.25E+02) = & 2.35E+03 (5.63E+02) = & 2.26E+03 (5.49E+02) = & 2.36E+03 (5.96E+02) = & 2.19E+03 (5.51E+02) = & \textbf{2.09E+03 (6.04E+02) =} \\
    F11   & 2.82E+01 (2.26E+01) & 3.18E+01 (2.72E+01) = & 2.63E+01 (2.32E+01) = & 2.59E+01 (2.38E+01) = & 2.55E+01 (2.33E+01) = & 2.40E+01 (2.08E+01) = & \textbf{2.10E+01 (1.78E+01) -} \\
    F12   & 1.28E+05 (7.34E+04) & \textbf{7.55E+04 (3.79E+04) -} & 9.06E+04 (4.24E+04) - & 1.16E+05 (5.38E+04) = & 1.32E+05 (7.31E+04) = & 1.66E+05 (8.49E+04) + & 1.37E+05 (7.89E+04) = \\
    F13   & 1.30E+04 (5.61E+03) & \textbf{7.61E+03 (5.56E+03) -} & 1.02E+04 (5.44E+03) - & 1.28E+04 (4.86E+03) = & 1.40E+04 (5.16E+03) = & 1.47E+04 (5.45E+03) = & 1.61E+04 (5.44E+03) + \\
    F14   & 9.82E+01 (8.74E+00) & \textbf{9.09E+01 (8.71E+00) -} & 9.20E+01 (8.82E+00) - & 9.58E+01 (8.04E+00) = & 1.04E+02 (8.66E+00) + & 1.05E+02 (9.29E+00) + & 1.05E+02 (9.77E+00) + \\
    F15   & 2.08E+02 (3.89E+01) & \textbf{1.46E+02 (2.90E+01) -} & 1.67E+02 (2.89E+01) - & 1.95E+02 (3.75E+01) = & 2.14E+02 (3.84E+01) = & 2.08E+02 (2.98E+01) = & 2.18E+02 (4.29E+01) = \\
    F16   & 5.60E+02 (2.18E+02) & 6.96E+02 (2.74E+02) + & 6.46E+02 (2.70E+02) + & 5.68E+02 (2.07E+02) = & 5.42E+02 (2.61E+02) = & 5.20E+02 (2.27E+02) = & \textbf{5.10E+02 (2.15E+02) =} \\
    F17   & 6.63E+01 (7.65E+01) & 1.53E+02 (1.09E+02) + & 1.23E+02 (8.89E+01) + & 8.64E+01 (8.79E+01) = & 6.12E+01 (6.61E+01) = & 4.39E+01 (3.04E+01) = & \textbf{3.90E+01 (2.16E+01) =} \\
    F18   & 9.51E+04 (9.09E+04) & 1.55E+05 (1.30E+05) + & 1.32E+05 (1.18E+05) + & 9.73E+04 (6.86E+04) = & \textbf{8.15E+04 (6.30E+04) =} & 1.07E+05 (8.27E+04) = & 8.83E+04 (6.34E+04) = \\
    F19   & 9.00E+01 (2.01E+01) & \textbf{6.86E+01 (1.72E+01) -} & 7.65E+01 (1.22E+01) - & 9.23E+01 (1.88E+01) = & 9.62E+01 (1.79E+01) = & 9.54E+01 (2.14E+01) = & 9.81E+01 (2.25E+01) = \\
    F20   & 8.23E+01 (8.07E+01) & 1.99E+02 (1.45E+02) + & 1.65E+02 (1.13E+02) + & 1.27E+02 (1.11E+02) + & 8.84E+01 (8.59E+01) = & \textbf{4.69E+01 (6.42E+01) =} & 5.26E+01 (6.68E+01) - \\
    F21   & 2.42E+02 (1.25E+01) & 2.49E+02 (1.42E+01) + & 2.44E+02 (1.41E+01) = & 2.42E+02 (1.08E+01) = & 2.42E+02 (1.21E+01) = & 2.41E+02 (1.18E+01) = & \textbf{2.40E+02 (9.82E+00) =} \\
    F22   & \textbf{1.00E+02 (0.00E+00)} & \textbf{1.00E+02 (0.00E+00) =} & \textbf{1.00E+02 (0.00E+00) =} & \textbf{1.00E+02 (0.00E+00) =} & \textbf{1.00E+02 (0.00E+00) =} & \textbf{1.00E+02 (0.00E+00) =} & \textbf{1.00E+02 (0.00E+00) =} \\
    F23   & 3.87E+02 (1.10E+01) & 3.92E+02 (1.25E+01) = & 3.86E+02 (1.26E+01) = & 3.85E+02 (1.20E+01) = & 3.86E+02 (1.13E+01) = & 3.85E+02 (1.19E+01) = & \textbf{3.84E+02 (1.06E+01) =} \\
    F24   & 4.60E+02 (9.70E+00) & 4.65E+02 (1.36E+01) + & 4.63E+02 (1.14E+01) = & 4.63E+02 (1.18E+01) = & \textbf{4.58E+02 (1.17E+01) =} & 4.62E+02 (1.05E+01) = & 4.60E+02 (1.03E+01) = \\
    F25   & \textbf{3.87E+02 (0.00E+00)} & \textbf{3.87E+02 (0.00E+00) =} & \textbf{3.87E+02 (0.00E+00) =} & \textbf{3.87E+02 (0.00E+00) =} & \textbf{3.87E+02 (0.00E+00) =} & \textbf{3.87E+02 (0.00E+00) =} & \textbf{3.87E+02 (0.00E+00) =} \\
    F26   & 1.33E+03 (1.22E+02) & 1.44E+03 (1.64E+02) + & 1.38E+03 (1.45E+02) = & 1.34E+03 (1.53E+02) = & 1.35E+03 (1.30E+02) = & 1.36E+03 (1.42E+02) = & \textbf{1.32E+03 (1.20E+02) =} \\
    F27   & 5.00E+02 (5.93E+00) & 5.04E+02 (5.14E+00) + & 5.02E+02 (4.01E+00) = & 5.01E+02 (5.94E+00) = & 5.00E+02 (4.83E+00) = & 5.01E+02 (5.60E+00) = & \textbf{4.98E+02 (4.55E+00) -} \\
    F28   & 3.70E+02 (4.56E+01) & \textbf{3.48E+02 (5.34E+01) =} & 3.56E+02 (5.48E+01) = & 3.61E+02 (4.04E+01) = & 3.72E+02 (4.28E+01) = & 3.60E+02 (4.59E+01) = & 3.77E+02 (3.40E+01) = \\
    F29   & 4.82E+02 (8.60E+01) & 5.07E+02 (1.10E+02) = & 5.22E+02 (1.16E+02) + & 4.72E+02 (8.00E+01) = & 4.60E+02 (5.32E+01) = & \textbf{4.48E+02 (4.14E+01) =} & \textbf{4.48E+02 (2.85E+01) =} \\
    F30   & 1.03E+04 (2.07E+03) & \textbf{7.11E+03 (1.56E+03) -} & 9.00E+03 (2.18E+03) - & 1.01E+04 (2.03E+03) = & 1.07E+04 (2.32E+03) = & 1.10E+04 (2.14E+03) = & 1.20E+04 (2.56E+03) + \\
    \midrule
    +/=/- &       & 10/12/8 & 6/16/8 & 2/26/2 & 3/27/0 & 4/26/0 & 5/22/3 \\
    \bottomrule
    \end{tabular}%
  \label{tab:failure_threshold_rand1bin}%
  \\The symbols "+/=/-" show the statistical results of the Wilcoxon signed-rank test with $\alpha = 0.05$ significance level. "+" represents that $FT_{init}=100$ variant is significantly superior than the corresponding variant. "=" represents that the performance difference between $FT_{init}=100$ variant and the corresponding variant is not statistically significant. And, "-" represents that $FT_{init}=100$ variant is significantly inferior than the corresponding variant.
\end{table*}%

\begin{table*}[htbp]
  \tiny
  \centering
  \caption{ACM-EDEV with different failure threshold settings ($D = 30$)}
    \begin{tabular}{cccccccc}
    \toprule
          & \multicolumn{7}{c}{EDEV} \\
          & $FT_{init}=100$ & $FT_{init}=30$ & $FT_{init}=50$ & $FT_{init}=80$ & $FT_{init}=130$ & $FT_{init}=150$ & $FT_{init}=180$ \\
          & MEAN (STD DEV) & MEAN (STD DEV) & MEAN (STD DEV) & MEAN (STD DEV) & MEAN (STD DEV) & MEAN (STD DEV) & MEAN (STD DEV) \\
    \midrule
    F1    & 2.78E-14 (1.95E-13) & 2.78E-16 (1.99E-15) = & \textbf{0.00E+00 (0.00E+00) =} & 2.78E-16 (1.99E-15) = & \textbf{0.00E+00 (0.00E+00) =} & \textbf{0.00E+00 (0.00E+00) =} & 1.39E-15 (4.26E-15) = \\
    F2    & 2.10E+08 (1.50E+09) & 2.15E+02 (1.53E+03) = & \textbf{1.17E+02 (8.27E+02) =} & 4.38E+04 (2.65E+05) = & 2.41E+07 (1.24E+08) = & 2.79E+07 (1.99E+08) = & 1.60E+11 (1.14E+12) = \\
    F3    & 2.58E+02 (1.12E+03) & \textbf{2.25E+01 (1.10E+02) =} & 4.26E+02 (1.77E+03) = & 4.08E+02 (1.63E+03) = & 1.18E+03 (3.64E+03) = & 1.23E+03 (4.46E+03) = & 9.94E+02 (3.04E+03) = \\
    F4    & \textbf{4.81E+01 (2.37E+01)} & 4.91E+01 (2.26E+01) = & 5.72E+01 (1.19E+01) = & 5.29E+01 (1.96E+01) = & 5.59E+01 (1.42E+01) = & 5.25E+01 (1.94E+01) = & 4.99E+01 (2.13E+01) = \\
    F5    & 2.58E+01 (5.60E+00) & 2.78E+01 (8.44E+00) = & 2.62E+01 (6.11E+00) = & \textbf{2.44E+01 (6.16E+00) =} & 2.48E+01 (5.23E+00) = & 2.51E+01 (5.21E+00) = & 2.47E+01 (5.34E+00) = \\
    F6    & 2.29E-11 (1.26E-10) & 9.39E-11 (6.17E-10) = & 8.20E-12 (3.02E-11) = & 2.25E-11 (6.39E-11) = & 7.26E-12 (2.53E-11) = & \textbf{4.41E-12 (1.12E-11) =} & 8.10E-12 (2.04E-11) = \\
    F7    & 5.75E+01 (5.37E+00) & \textbf{5.20E+01 (5.20E+00) -} & 5.37E+01 (5.50E+00) - & 5.56E+01 (5.83E+00) = & 5.81E+01 (5.20E+00) = & 5.72E+01 (4.68E+00) = & 5.84E+01 (5.21E+00) = \\
    F8    & 2.50E+01 (4.86E+00) & 2.90E+01 (8.27E+00) + & 2.60E+01 (6.71E+00) = & \textbf{2.38E+01 (4.52E+00) =} & 2.60E+01 (5.54E+00) = & 2.40E+01 (5.52E+00) = & 2.49E+01 (5.56E+00) = \\
    F9    & 1.42E-02 (6.63E-02) & 1.07E-02 (6.45E-02) = & 8.90E-03 (6.36E-02) = & 3.51E-03 (1.75E-02) = & 8.90E-03 (6.36E-02) = & \textbf{0.00E+00 (0.00E+00) =} & 7.02E-03 (3.02E-02) = \\
    F10   & \textbf{1.66E+03 (3.69E+02)} & 1.77E+03 (4.88E+02) = & 1.70E+03 (4.81E+02) = & 1.67E+03 (4.87E+02) = & 1.67E+03 (3.76E+02) = & 1.70E+03 (3.29E+02) = & 1.67E+03 (2.97E+02) = \\
    F11   & 1.78E+01 (1.89E+01) & 2.28E+01 (2.22E+01) = & 2.13E+01 (2.28E+01) = & 1.57E+01 (1.70E+01) = & \textbf{1.53E+01 (1.49E+01) =} & 1.95E+01 (2.40E+01) = & 1.77E+01 (1.81E+01) = \\
    F12   & 1.99E+03 (2.85E+03) & 1.29E+03 (8.13E+02) = & \textbf{1.18E+03 (3.97E+02) -} & 1.60E+03 (1.42E+03) = & 1.21E+03 (4.76E+02) - & 1.78E+03 (2.47E+03) = & 1.77E+03 (1.60E+03) = \\
    F13   & \textbf{4.40E+01 (2.15E+01)} & 5.54E+01 (3.54E+01) + & 6.41E+01 (6.27E+01) = & 6.04E+01 (4.12E+01) + & 4.95E+01 (3.76E+01) = & 7.95E+01 (9.22E+01) = & 6.35E+01 (5.92E+01) = \\
    F14   & 3.31E+01 (1.70E+01) & \textbf{3.17E+01 (1.45E+01) =} & 3.18E+01 (1.47E+01) = & 3.41E+01 (1.12E+01) = & 3.21E+01 (1.28E+01) = & 3.44E+01 (1.72E+01) = & 3.46E+01 (1.45E+01) = \\
    F15   & 3.71E+01 (3.95E+01) & 2.67E+01 (2.48E+01) = & 3.29E+01 (4.23E+01) = & 2.83E+01 (2.45E+01) = & 2.83E+01 (2.36E+01) = & \textbf{2.64E+01 (2.11E+01) =} & 2.74E+01 (2.13E+01) = \\
    F16   & 3.78E+02 (1.45E+02) & 4.19E+02 (1.99E+02) = & 3.39E+02 (1.39E+02) = & \textbf{2.90E+02 (1.15E+02) -} & 3.39E+02 (1.54E+02) = & 2.92E+02 (1.46E+02) - & 3.32E+02 (1.42E+02) = \\
    F17   & \textbf{3.57E+01 (1.83E+01)} & 6.62E+01 (7.72E+01) = & 4.70E+01 (4.27E+01) = & 3.68E+01 (2.88E+01) = & 3.85E+01 (1.56E+01) = & 4.49E+01 (1.45E+01) + & 4.34E+01 (2.02E+01) = \\
    F18   & 1.10E+03 (3.97E+03) & \textbf{1.78E+02 (6.99E+02) =} & 6.05E+02 (2.46E+03) = & 9.56E+02 (4.28E+03) = & 1.50E+03 (5.24E+03) = & 3.90E+03 (1.58E+04) = & 2.12E+02 (9.47E+02) = \\
    F19   & 1.73E+01 (1.17E+01) & 1.83E+01 (2.13E+01) = & 2.01E+01 (2.16E+01) = & 1.78E+01 (1.40E+01) = & 1.70E+01 (1.16E+01) = & \textbf{1.69E+01 (1.12E+01) =} & 2.03E+01 (2.21E+01) = \\
    F20   & 6.16E+01 (6.01E+01) & 9.65E+01 (8.45E+01) = & 4.12E+01 (6.05E+01) - & 5.23E+01 (5.79E+01) = & 4.81E+01 (5.36E+01) = & 5.36E+01 (5.51E+01) = & \textbf{4.07E+01 (4.54E+01) =} \\
    F21   & 2.28E+02 (6.41E+00) & 2.30E+02 (8.03E+00) = & 2.28E+02 (6.07E+00) = & 2.28E+02 (5.98E+00) = & \textbf{2.26E+02 (5.92E+00) =} & \textbf{2.26E+02 (5.68E+00) =} & 2.27E+02 (6.73E+00) = \\
    F22   & \textbf{1.00E+02 (0.00E+00)} & \textbf{1.00E+02 (0.00E+00) =} & \textbf{1.00E+02 (0.00E+00) =} & \textbf{1.00E+02 (0.00E+00) =} & \textbf{1.00E+02 (0.00E+00) =} & \textbf{1.00E+02 (0.00E+00) =} & \textbf{1.00E+02 (0.00E+00) =} \\
    F23   & 3.73E+02 (7.12E+00) & 3.78E+02 (8.68E+00) + & 3.72E+02 (7.20E+00) = & 3.75E+02 (6.80E+00) = & 3.72E+02 (5.18E+00) = & 3.72E+02 (7.43E+00) = & \textbf{3.71E+02 (6.51E+00) =} \\
    F24   & \textbf{4.41E+02 (5.58E+00)} & 4.48E+02 (6.59E+00) + & 4.44E+02 (7.01E+00) + & 4.43E+02 (6.15E+00) + & 4.42E+02 (5.36E+00) = & \textbf{4.41E+02 (5.42E+00) =} & 4.42E+02 (5.96E+00) = \\
    F25   & \textbf{3.87E+02 (0.00E+00)} & \textbf{3.87E+02 (0.00E+00) =} & \textbf{3.87E+02 (0.00E+00) =} & \textbf{3.87E+02 (0.00E+00) =} & \textbf{3.87E+02 (0.00E+00) =} & \textbf{3.87E+02 (0.00E+00) =} & \textbf{3.87E+02 (0.00E+00) =} \\
    F26   & 1.17E+03 (1.43E+02) & 1.22E+03 (9.78E+01) = & 1.21E+03 (8.43E+01) = & 1.20E+03 (8.44E+01) = & \textbf{1.16E+03 (1.42E+02) =} & 1.20E+03 (6.95E+01) = & 1.23E+03 (7.29E+01) + \\
    F27   & 5.02E+02 (7.04E+00) & 5.04E+02 (8.28E+00) = & \textbf{4.99E+02 (8.42E+00) =} & \textbf{4.99E+02 (6.58E+00) =} & 5.01E+02 (6.71E+00) = & 5.00E+02 (7.53E+00) = & 5.00E+02 (6.44E+00) = \\
    F28   & \textbf{3.21E+02 (4.65E+01)} & 3.29E+02 (5.31E+01) = & 3.50E+02 (6.09E+01) = & 3.38E+02 (5.82E+01) = & 3.29E+02 (5.10E+01) = & 3.35E+02 (5.25E+01) = & 3.36E+02 (5.48E+01) = \\
    F29   & \textbf{4.20E+02 (3.67E+01)} & 4.38E+02 (2.64E+01) + & 4.26E+02 (3.34E+01) = & \textbf{4.20E+02 (4.13E+01) =} & 4.31E+02 (3.22E+01) = & 4.31E+02 (4.04E+01) + & 4.29E+02 (3.30E+01) = \\
    F30   & 2.41E+03 (1.32E+03) & \textbf{2.11E+03 (1.30E+02) =} & 2.12E+03 (1.69E+02) = & 2.16E+03 (1.73E+02) = & 2.19E+03 (5.75E+02) = & 2.25E+03 (1.16E+03) = & 2.25E+03 (8.39E+02) = \\
    \midrule
    +/=/- &       & 5/24/1 & 1/26/3 & 2/27/1 & 0/29/1 & 2/27/1 & 1/29/0 \\
    \bottomrule
    \end{tabular}%
  \label{tab:failure_threshold_EDEV}%
  \\The symbols "+/=/-" show the statistical results of the Wilcoxon signed-rank test with $\alpha = 0.05$ significance level. "+" represents that $FT_{init}=100$ variant is significantly superior than the corresponding variant. "=" represents that the performance difference between $FT_{init}=100$ variant and the corresponding variant is not statistically significant. And, "-" represents that $FT_{init}=100$ variant is significantly inferior than the corresponding variant.
\end{table*}%


\section{Conclusion}
\label{sec:Conclusion}
EAs need to establish a balance between exploration and exploitation to be successful \cite{eiben1998evolutionary, vcrepinvsek2013exploration}. The previous Cauchy mutation of MDE was proposed to increase the convergence speed of DE by using the best individual information and the Cauchy distribution, which can locate the consecutively failed individuals to new positions close to the best individual. Although the effectiveness of MDE has been demonstrated successfully on the classical benchmark problems, MDE lacks robustness and faces difficulty in optimizing complex problems because of its strong exploitation.

We have proposed a variant of MDE called advanced Cauchy mutation DE (ACM-DE). We employed a sigmoid based parameter control, which alters the failure threshold for performing the Cauchy mutation in a time-varying schedule. That is, ACM-DE assigns a high failure threshold at the beginning of the search process and gradually reduces it over generations. We also employed the $p$-best individual based Cauchy mutation to prevent premature convergence.

ACM-DE has been tested on a set of 30 different and difficult CEC 2017 benchmark problems. Experiments and comparisons were carried out with six conventional and six advanced DE variants to demonstrate the effectiveness of the proposed Cauchy mutation. The experiment results indicate that the DE variants assisted by the proposed Cauchy mutation have better or at least competitive performance in terms of accuracy and robustness, particularly for multimodal functions compared to the DE variants assisted by the previous Cauchy mutation as well as the original DE variants.

Possible directions for future work include 1) designing a mutation based on the multivariate Cauchy distribution; 2) extending the proposed Cauchy mutation for multi-objective optimization; 3) applying the proposed Cauchy mutation to other population-based metaheuristics.

\bibliographystyle{IEEEtran}
\bibliography{refs}

\EOD

\end{document}